\documentclass[preprint,3p,12pt]{elsarticle}

\usepackage{amsmath}
\usepackage{amsfonts}
\usepackage{amssymb}
\usepackage{amsthm}
\usepackage{amsopn}
\usepackage{dsfont}
\usepackage{xfrac}

\usepackage{graphicx}
\usepackage{epstopdf}
\usepackage{tikz}
\usepackage{pgfplots}
\usepgfplotslibrary{groupplots,dateplot}

\usepackage{booktabs}
\usepackage{multirow}
\usepackage{tabularray}
\usepackage{colortbl}
\usepackage{makecell}
\usepackage{rotating}

\usepackage[utf8]{inputenc}
\usepackage{lipsum}
\usepackage{enumitem}
\usepackage{comment}
\usepackage{adjustbox}
\usepackage{tcolorbox}
\tcbuselibrary{skins,breakable}
\usepackage{cancel}
\usepackage{float}

\usepackage[font=small,labelfont=bf]{caption}
\usepackage{subcaption}
\usepackage{hyperref}
\usepackage{cleveref}
\usepackage{xr}

\usepackage[normalem]{ulem}

\usepackage{algorithmic}

\usetikzlibrary{
  positioning,
  arrows,
  shapes,
  decorations.pathreplacing,
  calligraphy,
  patterns,
  shapes.arrows
}

\DeclareUnicodeCharacter{2212}{−}
\usepgfplotslibrary{groupplots,dateplot}
\usetikzlibrary{patterns,shapes.arrows}
\pgfplotsset{compat=newest}

\usetikzlibrary{positioning}

\ifpdf
  \DeclareGraphicsExtensions{.eps,.pdf,.png,.jpg}
\else
  \DeclareGraphicsExtensions{.eps}
\fi
\usepackage{comment}

\usepackage{tikz}
\usetikzlibrary{arrows,shapes}
\usetikzlibrary{decorations.pathreplacing,calligraphy}
\usetikzlibrary{patterns}
\usetikzlibrary{arrows,shapes}
\usetikzlibrary{decorations.pathreplacing,calligraphy}

\usepackage{amsopn}
\DeclareMathOperator{\diag}{diag}

\usepackage{xr}
\makeatletter
\newcommand*{\addFileDependency}[1]{
  \typeout{(#1)}
  \@addtofilelist{#1}
  \IfFileExists{#1}{}{\typeout{No file #1.}}
}
\makeatother

\newcommand{\nor}[1]{\left\| #1 \right\|} 
\newcommand{\snor}[1]{\left| #1 \right|} 
\newcommand{\LRp}[1]{\left( #1 \right)} 
\newcommand{\LRs}[1]{\left[ #1 \right]} 
\newcommand{\LRc}[1]{\left\{ #1 \right\}} 

\newcommand{\mc}[1]{\mathcal{#1}} 
\newcommand{\mb}[1]{\mathbf{#1}} 
\newcommand{\half}{\frac{1}{2}}
\newcommand{\halfv}[1]{\frac{#1}{2}}

\newcommand{\W}{W}

\newcommand{\F}{\mc{G}}

\newcommand{\bs}[1]{\boldsymbol{#1}}
\renewcommand{\P}{U}
\newcommand{\U}{\P}

\newcommand{\Psie}{\Psi_{\mathrm{e}}}
\newcommand{\Psid}{\Psi_{\mathrm{d}}}

\newcommand{\bb}{{\bf b}}

\newcommand{\B}{B}

\newcommand{\Y}{Y}

\newcommand{\yb}{\bs{y}}

\newcommand{\ybtest}{\yb^\text{test}}
\newcommand{\One}{\mathds{1}}

\newcommand{\R}{{\bs{\mathbb{R}}}}
\newcommand{\I}{I}
\newcommand{\Ib}{{\I}}

\newcommand{\G}{G}
\newcommand{\GB}{\G^{\B}}
\newcommand{\zb}{\bs{z}}
\newcommand{\p}{u}
\newcommand{\pb}{\bs{\p}}
\newcommand{\ub}{\pb}

\newcommand{\ubtest}{\ub^\text{test}}

\newcommand{\ybfull}{\bs{\omega}}
\newcommand{\ybfulltest}{\bs{\omega}^\text{test}}

\newcommand{\etab}{\bs{\eta}}

\usepackage{hyperref}
\hypersetup{
  colorlinks=true,
  linkcolor=blue,
  filecolor=magenta,
  urlcolor=red,
  pdftitle={Model-Constrained Deep Learning Approaches for Inverse Problems},
  pdfpagemode=FullScreen,
}

\newcommand{\figlab}[1]{\label{fig:#1}}
\newcommand{\eqnlab}[1]{\label{eq:#1}}

\newcommand{\tablab}[1]{\label{tab:#1}}

\newcommand{\seclab}[1]{\label{sect:#1}}

\newcommand{\eval}[2][\right]{\relax \ifx#1\right\relax \left.\fi#2#1\rvert}
\renewcommand{\epsilon}{\varepsilon}
\newcommand{\epsb}{\bs{\epsilon}}

\newcommand{\TNet}{\textcolor{black}{\texttt{TNet}}}

\newcommand{\purePOP}{\textcolor{black}{\texttt{nPOP}}}
\newcommand{\pureOPO}{\textcolor{black}{\texttt{nOPO}}}
\newcommand{\mcPOP}{\textcolor{black}{\texttt{mcPOP}}}
\newcommand{\mcOPO}{\textcolor{black}{\texttt{mcOPO}}}
\newcommand{\mcOPOfull}{\textcolor{black}{\texttt{mcOPO-Full}}}

\newcommand{\TNetAE}{\textcolor{black}{\texttt{TAEN}}}
\newcommand{\TNetAEfull}{\textcolor{black}{\texttt{TAEN-Full}}}

\newcommand{\du}{{n}}
\newcommand{\dw}{{p}}
\newcommand{\dy}{{m}}

\usepackage{booktabs}

\journal{CMAME}

\begin{document}

\begin{frontmatter}

\title{\texttt{TAEN}: A Model-Constrained Tikhonov Autoencoder Network for Forward and Inverse Problems}

\author[label1]{Hai Van Nguyen} 
\author[label1,label2]{Tan Bui Thanh} 
\author[label1,label2]{Clint Dawson} 

\affiliation[label1]{organization={Department of Aerospace Engineering and Engineering Mechanics, the University of Texas at Austin},
            city={Austin},
            postcode={78712}, 
            state={Texas},
            country={USA}}

\affiliation[label2]{organization={The Oden Institute for Computational Engineering and Sciences, the University of Texas at Austin},
            city={Austin},
            postcode={78712}, 
            state={Texas},
            country={USA}}






\begin{abstract}
    Efficient real-time solvers for forward and inverse problems are essential in engineering and science applications. Machine learning surrogate models have emerged as promising alternatives to traditional methods, offering substantially reduced computational time. Nevertheless, these models typically demand extensive training datasets to achieve robust generalization across diverse scenarios. While physics-based approaches can partially mitigate this data dependency and ensure physics-interpretable solutions, addressing scarce data regimes remains a challenge. Both purely data-driven and physics-based machine learning approaches demonstrate severe overfitting issues when trained with insufficient data. {\em We propose a novel model-constrained Tikhonov autoencoder neural network framework, called \TNetAE{}, capable of learning both forward and inverse surrogate models using a single arbitrary observational sample}. We develop comprehensive theoretical foundations including forward and inverse inference error bounds for the proposed approach for linear cases. For comparative analysis, we derive equivalent formulations for pure data-driven and model-constrained approach counterparts. At the heart of our approach is a data randomization strategy with theoretical justification, which functions as a generative mechanism for exploring the training data space, enabling effective training of both forward and inverse surrogate models even with a single observation, while regularizing the learning process. We validate our approach through extensive numerical experiments on two challenging inverse problems: 2D heat conductivity inversion and initial condition reconstruction for time-dependent 2D Navier--Stokes equations. Results demonstrate that \TNetAE{} achieves accuracy comparable to traditional Tikhonov solvers and numerical forward solvers for both inverse and forward problems, respectively, while delivering orders of magnitude computational speedups.
\end{abstract}


\end{frontmatter}



\section{Introduction}
\seclab{introduction}



Partial differential equations (PDEs) serve as the mathematical foundation for describing physical phenomena in science and engineering applications \cite{sommerfeld1949partial, guenther1996partial}. Generally, these equations lack analytical solutions except in limited special cases, necessitating numerical approaches such as finite difference \cite{perrone1975general, godunov1959finite, ozicsik2017finite}, finite element \cite{rao2010finite, reddy1993introduction}, and finite volume methods \cite{eymard2000finite, leveque2002finite}. These computational techniques, while essential, often impose significant computational burdens, particularly for high-dimensional problems.
In this context, PDE-constrained inverse problems represent pervasive mathematical methods in inferring knowledge from experimental and observational data by leveraging numerical simulations and models \cite{kaipio2006statistical, parker1994geophysical, vogel2002computational, tarantola2005inverse}. Among various methodologies, the Tikhonov regularization framework stands as a predominant approach for addressing general inverse problems \cite{ito2014inverse, zhdanov2002geophysical, nguyen2022dias, golub1999tikhonov}. This optimization-based methodology necessitates multiple evaluations of the forward model, which is typically governed by PDEs. Consequently, PDE-constrained inverse problems face two primary challenges: they not only demand substantial computational resources but also require extensive processing time to obtain solutions, making them impractical for applications requiring real-time responses.


Machine learning has demonstrated remarkable success across diverse domains, from computer vision and natural language processing \cite{krizhevsky2012imagenet, deng2009imagenet, vaswani2017attention} to physics-based applications including experimental design \cite{singh2017machine, chen2018using} and digital twins~\cite{moya2022digital, rathore2021role, austin2020architecting}. The autoencoder architecture has emerged as a versatile framework for applications in image processing tasks such as denoising and inpainting \cite{chen2017deep, gondara2016medical, he2022masked}, as well as in transformer-based natural language processing systems \cite{vaswani2017attention}. In science and engineering applications, machine learning approaches have demonstrated significant potential in developing computationally efficient surrogate models for both forward and inverse problems \cite{raissi2019physics, o2022derivative, o2024derivative, van2024model, nguyen2022model, li2020fourier, lu2021learning}.
Traditional data-driven learning methodologies, which construct surrogate models without incorporating physical principles 
\cite{wu2018inversionnet, arridge2019solving}, face several limitations. These include the requirement for extensive training datasets to achieve accurate forward or inverse surrogate models \cite{li2020nett, lunz2018adversarial, aggarwal2018modl, pakravan2021solving, wu2018inversionnet}, limited generalization capabilities, and generating non-physical solutions. To address these challenges, physics-based machine learning approaches have been developed, integrating governing physical principles into the learning framework \cite{raissi2019physics, ongie2020deep, nguyen2024tnet, fan2020solving}. However, as noted in \cite{nguyen2024tnet}, these physics-based approaches remain susceptible to overfitting when training data are scarce.
The acquisition of comprehensive training datasets presents significant challenges in many engineering applications due to prohibitive costs or practical limitations. For instance, in full wave-form inversion, data collection is constrained by the substantial expenses associated with sensor placement (such as multi-million dollar oil well drilling operations) or accessibility challenges in certain environments (like deep ocean floors). This underscores the critical need for developing machine learning methodologies capable of constructing physics-interpretable surrogate models from minimal training samples.



Physics-informed neural networks (PINNs)~\cite{raissi2019physics} employ neural networks to parametrize solutions for both PDEs \cite{raissi2019physics, wang2021learning, cai2021physics} and PDE-constrained inverse problems \cite{raissi2019physics, jagtap2020conservative, jagtap2022physics, chen2020physics}, using spatial or temporal variables as inputs. The neural network parameters are optimized through objective functions comprising multiple loss terms, including PDE constraints, boundary conditions, initial conditions, and observation data misfits.
An alternative approach, aligned with the concept of neural fields \cite{xie2022neural}, has emerged in inverse problem solving \cite{sun2023implicit, sitzmann2020implicit, fan2020solving, berg2017neural}. This methodology processes neural network inverted solutions through numerical forward simulations to generate synthetic data, which is then reconciled with observational data to optimize the neural representation of solutions. While these approaches offer the advantage of mesh-independent solutions---capable of producing results at any spatiotemporal point post-training—--they suffer from a significant limitation: the requirement to retrain the network for each new scenario, including variations in model parameters, boundary conditions, or observation sets.
As such, these methods do not have the ability to generalize to unseen inference cases, which are critical requirements for making physics-based machine learning surrogate models viable in real-time applications. Furthermore,
research has shown that PINNs demand extensive training periods and numerous architectural experiments to achieve adequate solution convergence \cite{cuomo2022scientific}. This limitation becomes particularly problematic in inverse problems, in which the PDE must be solved repeatedly throughout optimization iterations. Alternative approaches proposed in \cite{romano2017little, zhang2017learning} focus on learning regularizer functions for inverse problems from extensive training datasets. While these learned neural network regularizers enhance solution quality compared to traditional Tikhonov regularization, they essentially represent another computationally intensive method for solving PDEs or PDE-constrained inverse problems, making them unsuitable for real-time applications. 


Various methodologies have emerged for learning parameter-to-observation maps, or more general, numerical PDE simulations, focusing on predicting solution for unseen scenarios from historical snapshots of solutions. Neural ODE \cite{chen2018neural} and related approaches \cite{van2024model, 661124, nguyen2022model, Zhuang_2021} propose learning the right-hand side of semi-discretized equations, offering the distinctive advantage of predicting solutions at arbitrary temporal points through time integration schemes.
Alternative frameworks employ recurrent neural network architectures \cite{ku1995diagonal, funahashi1993approximation, mohajerin2019multistep} or transformer-based models \cite{geneva2022transformers, li2022transformer} to process sequences of historical solution snapshots for future state predictions. These methods, however, are usually constrained by the requirement of uniform temporal sampling for both input and predicted solutions.
Another promising direction involves learning dynamics in latent space, where solutions are projected into lower-dimensional representations using autoencoder-like architectures \cite{xu2020multi, wan2023evolve, gonzalez2018deep, jin2020deep, lee2020model} or graph pooling techniques \cite{barwey2023interpretable, pichi2024graph}. In these approaches, neural networks are employed to approximate the evolution of the system in the compressed latent space.
The operator learning approach \cite{li2020fourier, lu2021learning} potentially offers computationally efficient surrogate models by directly learning functional mappings between parameter fields and target solutions across the physical domain. While all these methodologies demonstrate considerable promise in capturing PDE system dynamics and generating future solution predictions, they generally require substantially large training data to develop accurate surrogate forward models or functional mapping operators.



A variety of machine learning approaches have been developed to create generalizable inverse solvers for PDE-constrained inverse problems. Several researchers \cite{jin2020physics, nguyen2024tnet, pakravan2021solving} have introduced methods that directly regularize neural network inverse surrogate models through differentiable PDE solvers. In these approaches, the predicted Parameters of Interest (PoI) from neural networks are processed through PDE solvers to regenerate synthetic observation data, which is expected to match training observation input data. These physics-based methods demonstrate significant improvements over pure data-driven approaches given equivalent training data volumes. However, the training data requirements for effective inverse mappings correlate with the non-linearity of the parameter-to-observables (PtO) map, making them inherently problem-specific.
Our previous research \cite{nguyen2024tnet} introduced the model-constrained \TNet{} approach for learning Tikhonov inverse solvers from a finite training dataset. Notably, with identical sample sizes, both pure machine learning and physics-based approaches face overfitting challenges compared to \TNet. Related work \cite{liu2018training, jin2020physics} proposes first training the forward map to learn the PDE solver, which then constrains the inverse map learning process with pre-trained PDE surrogate models, thereby reducing overall training costs.
Alternative approaches \cite{chung2024paired, boink2019learned} employ dual autoencoder architectures to compress both observation and PoI data spaces into lower-dimensional latent representations, followed by learning linear inverse mappings between these compressed spaces. Additionally, variational autoencoder architectures have been widely adopted for probabilistic inverse problem solving \cite{goh2019solving, almaeen2021variational}. Some efforts \cite{goh2019solving} focus on simultaneous learning of forward and inverse surrogate models. However, in this work, concurrent training of encoders and decoders for both forward and inverse mappings can complicate training convergence due to the increased dimensionality of the trainable parameter space, and the competition between the encoder and the decoder.

In scientific machine learning, enhancing surrogate model generalization remains a critical objective and a grand challenge. Several researchers \cite{pan2018long, yu2022gradient, drucker1992improving, ross2018improving, finlay2021scaleable} have proposed incorporating input-gradient information into the loss function to improve neural network solution accuracy and generalization capabilities. However, the computational cost of explicitly computing Jacobian matrices through back-propagation for optimization is substantial, particularly for high-dimensional input problems, due to the requirement for double back-propagation. While \cite{o2024derivative} demonstrates that exploiting low-rank properties of Jacobian matrices can reduce computational overhead, our previous work \cite{nguyen2022model, nguyen2024tnet} shows that comparable regularization effects can be achieved through data randomization techniques in forward surrogate model learning.
This randomization approach, which introduces controlled noise to neural network input data, has been shown to enhance long-term predictive stability \cite{sanchez2020learning, poggio1990networks,nguyen2022model}. Theoretical foundations established in \cite{reed1992regularization, bishop1995training, matsuoka1992noise} demonstrate that input data perturbation effectively functions as Tikhonov regularization, promoting input-space smoothness in neural network responses. Our recent work \cite{nguyen2024tnet,nguyen2022model} show that randomization induces regularization on the derivatives without actually computing them (which we try to avoid due to the aforementioned reasons). Our works further validate that data randomization not only improves the generalization capability of learned inverse surrogate models to unseen observation scenarios but also substantially reduces training data requirements.


This paper extends our previous research \cite{nguyen2024tnet} by introducing \TNetAE{}, a Tikhonov autoencoder model-constrained learning framework {\bf capable of addressing both forward and inverse problems using only a single arbitrary observational sample}.
For comparative analysis, we examinate various autoencoder architectures, including pure data-driven approaches \cite{wu2018inversionnet} and model-constrained methods \cite{pakravan2021solving, fan2020solving, jin2020physics}. Our theoretical analysis provides forward and inverse solution error estimates across all approaches for linear problems. These derivations, combined with numerical evidence, suggest that both pure data-driven and model-constrained approaches may yield suboptimal learning strategies for inverse problems due to potential training data bias.
In contrast, \TNetAE{} employs data randomization techniques with just one arbitrary observational sample to learn Tikhonov inverse solutions. For linear inverse problems, the approach successfully recovers Tikhonov-regularized solutions while maintaining exact compliance with governing equations at training points. Furthermore, given the pre-trained inverse map encoder and the same observation sample, the decoder accurately represents the forward map/parameter to observable (PtO) mapping for linear problems. We also provide theoretical justification for sequential encoder-decoder training rather than simultaneous optimization.
The effectiveness of our proposed approach is demonstrated through extensive numerical experiments on two challenging applications: 2D inverse heat conductivity inversion and initial condition reconstruction for time-dependent 2D Navier--Stokes equations.


There are several limitations to our approaches. The primary constraint stems from the requirement for a differentiable solver, which results in training and memory costs that scale with multiple forward solution computations. One potential solution, as demonstrated in \cite{jin2020physics}, is to learn surrogate models for PDEs before incorporating them into \TNetAE{} framework.
A second limitation arises from \TNetAE{}'s design principle: learn the Tikhonov inverse solver and the forward solver at the same time. While it offers improved interpretability and accuracy compared to purely data-driven approaches, its inverse performance is inherently bounded by the traditional Tikhonov method. Consequently, for scenarios with abundant data, alternative approaches such as \cite{lunz2018adversarial, aggarwal2018modl, li2020nett} may achieve superior accuracy relative to traditional Tikhonov methods, though these alternatives cannot provide real-time inverse solutions as \TNetAE{} does.
Finally, while \TNetAE{} offers an advantage over existing deep neural network approaches by deriving its regularization parameter directly from the Tikhonov framework—rather than lacking a principled parameter selection method—the challenge of determining optimal regularization parameters remains, just like any approach in finding optimal regularization parameter. This optimization task is highly problem-dependent and computationally extensive. 


The paper's organization is as follows. \Cref{sect:Intro_notation} establishes the fundamental framework of linear and nonlinear PDEs and PDE-constrained inverse problems, along with notational conventions. \Cref{sect:naive_autoencoder} examines data-driven autoencoder machine learning approaches, specifically \purePOP{} and \pureOPO{}. \Cref{sect:mc_autoencoder} explores model-constrained autoencoder machine learning approaches—\mcPOP{}, \mcOPO{}, and \mcOPOfull{}—which learn inverse and forward mappings while incorporating model constraints from the parameter-to-observable map of the discretized PDE. In \Cref{sect:TNet_autoencoder}, we introduce Tikhonov autoencoder neural networks, \TNetAE{} and its variant \TNetAEfull{},  designed to learn Tikhonov solver and PtO/forward maps while eliminating unnecessary training biases. \Cref{sect:simultaneous_training} provides a detailed analysis comparing sequential versus simultaneous training strategies for encoder and decoder networks. \Cref{sect:numerical_results} presents comprehensive numerical experiments validating our theoretical developments through applications to 2D inverse heat conductivity problems and inverse initial conditions for 2D time-dependent Navier-Stokes equations. The paper concludes in \Cref{sect:conclusions} with future research directions, while detailed proofs of theoretical results are provided in \Cref{sect:Appendix}.

\section{Methodology}
\subsection{Problem setting and notation}
\seclab{Intro_notation}
In this research, we explore various auto-encoder approaches for learning the following functions: the parameter-to-observable (PtO) map or forward map (transforming PoIs to PDE solutions), and inverse mappings (converting observables to PoIs). Our investigation encompasses na\"ive machine learning approaches (\Cref{sect:naive_autoencoder}), model-constrained machine learning techniques (\Cref{sect:mc_autoencoder}), and Tikhonov model-constrained machine learning strategies (\Cref{sect:TNet_autoencoder}). We deviate from traditional concurrent encoder-decoder training by implementing a sequential approach—first training the encoder network, then optimizing the decoder using the pre-trained encoder. The comparative merits of sequential versus simultaneous training are analyzed in \Cref{sect:simultaneous_training}.
Our notation follows standard conventions: matrices are denoted by capital letters, vectors by boldface normal letters, and scalars by lowercase Roman letters. We seek to learn the forward map $\F: \R^{\du} \mapsto \R^{\dw}$ governed by a PDE:
\begin{equation*}
\ybfull = \F \LRp{\ub},
\end{equation*}
or its corresponding parameter-to-observable (PtO) map:
\begin{equation*}
\yb = \B \circ \F \LRp{\ub} + \etab = \B \ybfull + \etab,
\label{eq:noisyObs}
\end{equation*}
where $\etab$ represents observation noise, $\ub \in \R^{\du}$ denotes the parameter of interest (PoI),\footnote{The parameter vector could be a discrete representation of some distributed parameter field.} $\ybfull \in \R^{\dw}$ represents the PDE solution, $\B: \R^{\dw} \mapsto \R^{\dy}$ defines the observation operator, and $\yb \in \R^{\dy}$ represents the observation. For inverse problems, we aim to learn the mapping:
\begin{equation*}
\ub = \LRp{\B \circ \F}^{-1} \LRp{\yb}.
\end{equation*}
Note that the inverse map is not well-defined in general. As we shall show, we learn regularized inverse maps instead.

For clarity in our derivations, we define the following notations: Training data matrices $\P \in \R^{\du \times n_t}$ and $\Y \in \R^{\dy \times n_t}$ contain $n_t$ PoI samples and $n_t$ observation samples, respectively. For linear problems, $\G$ represents the linear forward map, while $\GB = B \circ \G$ denotes its corresponding linear PtO map. Neural network components $\Psie$ and $\Psid$ represent encoder and decoder networks, respectively. In linear neural networks, $\LRc{\W_e, \bb_e}$ and $\LRc{\W_d, \bb_d}$ denote weight matrices and bias vectors for encoder and decoder networks, respectively. We define additional notations: $\bar{\bs{x}} = X \frac{\One}{n_t}$ computes the column mean of matrix $X$, $\bar{X} = X - \bar{\bs{x}} \One^T$ represents the centered-column matrix, $\One$ denotes a column vector of ones, while $\nor{\cdot}_{F}$ indicates the Frobenius norm and $\dagger$ denotes matrix pseudo-inverse.

\subsection{Na\"ive autoencoder learning approaches}
\seclab{naive_autoencoder}

In this section, we present two na\"ive data-driven autoencoder approaches that are commonly employed for learning PtO and inverse mappings in engineering applications. These approaches rely exclusively on training data pairs $\LRc{\P,\Y}$ without incorporating any physical models or domain knowledge into their loss functions \cite{goh2019solving,chung2024paired,wu2018inversionnet, arridge2019solving}. 

\subsubsection{Na\"ive autoencoder PtO-inverse learning approach \purePOP{}}
\seclab{naive_POP}

In this approach, we learn the encoder for PtO map and the decoder for inverse map. Given the training data set $\LRc{\P, \Y}$, the loss functions are defined as
\begin{equation}
    \tag{\purePOP}
    \eqnlab{purePOP}
    \begin{aligned}
        \Psie^* = \min_{\Psie} & \halfv{1} \nor{\Psie\LRp{\P} - \Y}_{F}^2, \\
        \Psid^* = \min_{\Psid} & \halfv{1}\nor{\Psid\LRp{\Psie^*\LRp{\P}} - \P}_{F}^2.
    \end{aligned}
\end{equation}
For analytical insights, we examine a noise-free linear inverse problem using linear neural networks. The linear encoder learns the PtO map as $Z = \Psie \LRp{\P} = \W_e U + \bb_e \One^T$, while the linear decoder learns the inverse map $\Psid\LRp{Z} = \W_d Z + \bb_d \One^T$. Substituting into \cref{eq:purePOP}, we obtain the following
\begin{equation}
        \begin{aligned}
            {\W_e^*, \bb_e^*} = & \min_{\W_e, \bb_e} \halfv{1} \nor{{\W_e \P + \bb_e \One^T} -\Y}_{F}^2, \\
            {\W_d^*, \bb_d^*} = & \min_{\W_d, \bb_d} \halfv{1}\nor{\W_d \LRp{\W_e^* \P + \bb_e^* \One^T} + \bb_d \One^T - \P}_{F}^2.
        \end{aligned}
        \label{eq:purePOP_linear}
    \end{equation}
Applying the first-order optimality conditions for \cref{eq:purePOP_linear} yields the following optimal solutions (see the complete derivation in \Cref{sect:derivation_purePOP}):
\begin{equation*}
    \eqnlab{purePOP_opt}
    \begin{aligned}
        \W_e^* & = \GB \bar{\P} \bar{\P}^\dagger,                                                                       & \bb_e^* & = \GB \LRp{ \Ib - \bar{\P} \bar{\P}^\dagger }\bar{\ub},                                         \\
        \W_d^* & = \bar{\P} \bar{\Y}^{\dagger} , & \bb_d^* & = \bar{\ub} - \bar{\P} \bar{\Y}^{\dagger} \bar{\yb}.\\
    \end{aligned}
\end{equation*}
It can be seen that if a sufficient number of training data samples are provided such that $\bar{\P}$ is a full row rank matrix, i.e. $\bar{\P} \bar{\P}^\dagger = \Ib$,  then the encoder is able to represent exactly the PtO map $\GB$, i.e., $\Psie^* \LRp{\ubtest} = \GB \ubtest = \ybtest$. In other words, the predicted PtO error estimation is zero, $\epsb_{\ybtest}^{\purePOP} = 0$. At the same time, 
the decoder is able to represent the right inverse of $\GB$. Indeed, given a test PoI sample $\ybtest$, we have 
\begin{equation*}
    \eqnlab{decoder_inverse_purePOP}
    \GB \Psid^* \LRp{\ybtest} = \GB \LRp{\W_d^* \ybtest + \bb_d^*} = \GB \bar{\P} \bar{\Y}^\dagger \ybtest  + \GB \bar{\ub} - \GB \bar{\P} \bar{\Y}^{\dagger} \bar{\yb} = \Ib \ybtest . 
\end{equation*}
It is important to note that the right inverse of $\GB$ is not unique. In our \purePOP{} approach, the decoder is concurrently satisfied with the following identity
\begin{equation*}
    \eqnlab{decoder_inverse_purePOP_2}
    \LRp{\Psid^* \LRp{\Y } - \P} {\Y}^T = 0. 
\end{equation*}
This identity indicates that the misfit of inverse solutions on training data is orthogonal to the column space of $\bar{\Y}$. Additionally, it also implies the data-dependent nature (or data-driven property) of the decoder. On the other hand, the decoder is not guaranteed to be the left inverse of $\GB$, i.e., the inverted solution error can be written as
\begin{equation}
    \eqnlab{inverse_purePOP_error}
    \epsb_{\ubtest}^{\purePOP} = \nor{\Psid^* \LRp{\ybtest} - \ubtest }_2^2 = \nor{ \LRp{\bar{\P} \bar{\Y}^\dagger \GB -\Ib} \LRp{\ubtest - \bar{\ub}}}_2^2.
\end{equation}
In the ideal case, we expect \purePOP{} to learn the unique left inverse of $\GB$, namely, $\LRp{\GB}^\dagger$, which is the best inverse map in the sense of solving least squares on the test sample. In that case, the error of inverse solutions reads
\begin{equation*}
    \nor{\LRp{\GB}^\dagger \ybtest - \ubtest}_2^2 = \nor{\LRp{\LRp{\GB}^\dagger \GB - \Ib} \ubtest}_2^2.
\end{equation*}
It is worth noting that, although we can achieve full row rank matrices $\bar{\P}, \bar{\Y}$ by data randomization technique \cite{MatrixFullRank}, the inverse solution error \Cref{eq:inverse_purePOP_error} can be very large. 


\subsubsection{Na\"ive autoencoder inverse-PtO learning approach \pureOPO{}}
\seclab{naive_OPO}

In contrast to the \purePOP{} approach, we propose the \pureOPO{} approach that learns inverse map with encoder and PtO map with decoder. The loss functions for \pureOPO{} are defined as
\begin{equation}
    \tag{\pureOPO}
    \eqnlab{pureOPO}
    \begin{aligned}
        \Psie^* = & \min \halfv{1} \nor{\Psie\LRp{\Y} - \P}_{F}^2, \\ 
        \Psid^* = & \min_{\Psid}  \halfv{1} \nor{\Psid\LRp{\Psie^*\LRp{\Y}} - \Y}_{F}^2.
    \end{aligned}
\end{equation}
For insights on \pureOPO{}, we analyze linear problems using a linear encoder for the inverse map $Z = \Psie \LRp{\Y} = \W_e Y + \bb_e \One^T$ and a linear decoder for learning PtO map $\Psid\LRp{Z} = \W_d Z + \bb_d \One^T$. 
Substituting into \cref{eq:pureOPO}, we have

\begin{equation}
    \eqnlab{pureOPO_linear}
    \begin{aligned}
        {\W_e^*, \bb_e^*} = & \min_{\W_e, \bb_e} \halfv{1} \nor{\W_e \Y + \bb_e \One^T - \P}_{F}^2, \\
        {\W_d^*, \bb_d^*} = & \min_{\W_d, \bb_d} \halfv{1} \nor{\W_d \LRp{\W_e^* \Y + \bb_e^* \One^T} + \bb_d \One^T - \Y}_{F}^2.
    \end{aligned}
\end{equation}
By applying the first optimality condition, the optimal solutions for \cref{eq:pureOPO_linear} (see \cref{sect:derivation_pureOPO} for full derivations) are

\begin{equation*}
    \eqnlab{pureOPO_opt}
    \begin{aligned}
        \W_e^* & = \bar{\P} \bar{\Y}^{\dagger},                                                                       & \bb_e^* & = \bar{\ub} - \bar{\P} \bar{\Y}^{\dagger} \bar{\yb},                                        \\
        \W_d^* & = \bar{\Y} \bar{Z}^{\dagger} , & \bb_d^* & = \bar{\yb} - \bar{\Y} \bar{Z}^{\dagger}\bar{\ub},
    \end{aligned}
\end{equation*}
where $\bar{Z} = \bar{\P} \bar{\Y}^{\dagger} \bar{\Y} $. It can be seen that 
the encoder is exactly the same as the decoder of \purePOP{} approach. Therefore, the inverse test error estimation is
\begin{equation*}
    \eqnlab{inverse_pureOPO_error}
    \epsb_{\ubtest}^{\pureOPO} = \epsb_{\ubtest}^{\purePOP} = \nor{ \LRp{\bar{\P} \bar{\Y}^\dagger \GB -\Ib} \LRp{\ubtest - \bar{\ub}}}_2^2.
\end{equation*}
For the decoder, we first rewrite the predicted PtO solution as
\begin{equation*}
    \begin{aligned}
        \yb^{\pureOPO} =  \W_d^* \ubtest + \bb_d^* \One^T & = \bar{\Y} \LRp{\bar{Z}^{\dagger} - \bar{\P}^\dagger + \bar{\P}^\dagger} \ubtest + \bar{\yb} - \bar{\Y} \LRp{\bar{Z}^{\dagger} - \bar{\P}^\dagger + \bar{\P}^\dagger}\bar{\ub} \\
        & = \G \ubtest + \bar{\Y} \LRp{\bar{Z}^{\dagger} - \bar{\P}^\dagger} \LRp{\ubtest - \bar{\ub}}.
    \end{aligned}
\end{equation*}
It implies the the PtO map prediction error can be expressed as
\begin{equation*}
    \epsb_{\ybtest}^{\pureOPO} = \nor{\yb^{\pureOPO} - \ybtest}_2^2 = \nor{\bar{\Y} \LRp{\bar{Z}^{\dagger} - \bar{\P}^\dagger} \LRp{\ubtest - \bar{\ub}}}_2^2.
\end{equation*}
It can be seen that the decoder is not able to learn the PtO map $\GB$ exactly even if a large amount of training data is available. Indeed, there is always a deviation due to error induced by the mismatch in inverted solutions obtained by the encoder solutions $\bar{Z}$ and ground truth training data $\bar{\P}$. Note that this error is inevitable since the encoder is not able to reconstruct the true $\P$ due to the ill-posed nature of the inverse map.

\subsection{Model constrained autoencoder learning approaches}
\seclab{mc_autoencoder}

In this section, we present two model-constrained AutoEncoder approaches that incorporate the forward model $\F$ into the loss function. As we shall show, {\em introducing the physics/discretization via the forward map provides better inverse surrogate models than na\"ive AutoEncoder approaches, presented in \cref{sect:naive_autoencoder}. The model-constrained AutoEncoder approaches can also learn the PtO/forward maps with only one training data sample.}

\subsubsection{Model-constrained autoEncoder PtO-inverse learning approach \mcPOP{}}
\seclab{mc_POP}

We propose a model-constrained AutoEncoder learning approach, \mcPOP{}, in which the encoder learns the PtO map and the decoder learns the inverse map. As opposed to \purePOP,  \mcPOP  adds a model-constrained term that takes the predicted inverse solution from the decoder and  reproduces the observation via the PtO map, $\GB$. The loss functions for \mcPOP{} are expressed as
\begin{equation}
    \tag{\mcPOP}
    \begin{aligned}
        \Psie^* = \min_{\Psie} & \halfv{1} \nor{\Psie\LRp{\P} - \Y}_{F}^2 , \\
        \Psid^* = \min_{\Psid} & \halfv{1} \nor{\Psid\LRp{\Psie^*\LRp{\P}} - \P}_{F}^2 +
        \halfv{\lambda} \nor{\B \circ \F\LRp{\Psid\LRp{\Psie^*\LRp{\P}}} - \Y}_{F}^2.
    \end{aligned}
\end{equation}
Following the same procedure for linear problems using linear neural networks, the loss functions for \mcPOP{} reads

\begin{equation}
    \eqnlab{mcPOP_linear}
    \begin{aligned}
        {\W_e^*, \bb_e^*} = \min_{\W_e, \bb_e} & \halfv{1} \nor{\W_e \P + \bb_e \One^T - \Y}_{F}^2, \\
        {\W_d^*, \bb_d^*} = \min_{\W_d, \bb_d} & \halfv{1} \nor{\W_d Z  + \bb_d \One^T - \P}_{F}^2
                     + \halfv{\lambda} \nor{\GB\LRp{\W_d Z + \bb_d \One^T} - \Y}_{F}^2,
    \end{aligned}
\end{equation}
where $Z = \W_e^* \P + \bb_e^* \One^T$. By apply the first optimality condition, the optimal solutions for \cref{eq:mcPOP_linear} for linear problems (derived in \cref{sect:derivation_mcPOP}) are
\begin{equation*}
    \eqnlab{mcPOP_opt}
    \begin{aligned}
        \W_e^*   & = \GB \bar{\P} \bar{\P}^\dagger,  \quad \quad \quad \bb_e^* = \GB \LRp{\Ib - \bar{\P} \bar{\P}^\dagger }\bar{\ub},                                                                                              \\
        \W_d^*   & = (\Ib + \lambda {\GB}^T{\GB})^{-1} \left( \bar{\P} \bar{\Y}^{\dagger} + \lambda {\GB}^T \bar{\Y} \bar{\Y}^{\dagger} \right),                                                                  \\
        \bb_d^* & = (\Ib + \lambda {\GB}^T{\GB})^{-1} \left( \bar{\ub} + \lambda {\GB}^T \bar{\yb}  - \left(\bar{\P} \bar{\Y}^{\dagger} + \lambda {\GB}^T \bar{\Y} \bar{\Y}^{\dagger} \right) \bar{\yb}   \right). \\
    \end{aligned}
\end{equation*}
It can be seen that the encoder is identical to the encoder obtained from \purePOP{}. If a sufficient amount of data is available, i.e., $\bar{\P}$ is full row rank, the PtO map $\GB$ can be learned well by the encoder, and thus $\epsb_{\ybtest}^{\mcPOP} = 0$. On the other hand, the inverse solutions obtained by the decoder turn out the same as the solution of the Tikonov regularization approach. Specifically, given an observation sample $\ybtest$, the inverse solution $\ub^{\mcPOP}$ can be derived as
\begin{equation*}
    \ub^{\mcPOP} = \left(\Ib + \lambda {\GB}^T{\GB}\right)^{-1} \left( \bar{\ub} + \lambda {\GB}^T \bar{\yb}  + \left(\bar{\P} \bar{\Y}^{\dagger} + \lambda {\GB}^T \bar{\Y} \bar{\Y}^{\dagger} \right) \LRp{\ybtest - \bar{\yb}}  \right)
\end{equation*}
which is exactly the solution of the following Tikonov regularization problem
\begin{equation*}
    \min_{\ub} \half \nor{\ub - \ub^\text{mc}}_2^2 + \halfv{\lambda} \nor{\GB \ub - \ybtest}_2^2,
\end{equation*}
where 
\begin{equation*}
    \ub^\text{mc} = \bar{\ub} + \bar{\P} \bar{\Y}^{\dagger} \LRp{\ybtest - \yb} -  \lambda {\GB}^T \left(\Ib - \bar{\Y} \bar{\Y}^{\dagger} \right) \LRp{\ybtest - \yb}.
\end{equation*}
The corresponding error of predicted inverse solution $\ub^{\mcPOP}$ can be derived as
\begin{multline*}
    \eqnlab{inverse_mcPOP_error}
        \epsb_{\ubtest}^{\mcPOP}  = \nor{\ub^{\mcPOP} - \ub^\text{test}}_2^2 \\
        = \nor{\left(\Ib + \lambda {\GB}^T{\GB}\right)^{-1} \LRp{  \LRp{\bar{\P} \bar{\Y}^{\dagger} \GB - \Ib } \LRp{\ub^\text{test} - \bar{\ub}}  + \lambda {\GB}^T \LRp{ \Ib - \bar{\Y} \bar{\Y}^{\dagger} } \GB \LRp{\ub^\text{test} - \bar{\ub}}} }_2^2.
\end{multline*}
It can be seen that if a sufficient amount of training data is given such that $\bar{\Y} \bar{\Y}^\dagger = \Ib$, the error of predicted inverse solution $\ub^{\mcPOP}$ is lower by a factor of $\LRp{\Ib + \lambda {\GB}^T {\GB}}^{-1}$ compared to \purePOP{} and \pureOPO{}.  This observation reveals the key benefit of the model-constrained AutoEncoder approach in learning the inverse map compared to the pure data-driven counterparts. However, we emphasize that the error of predicted inverse solution $\ub^{\mcPOP}$ is still data-dependent in that the inverse error is proportional to the distance $\LRp{\ubtest - \bar{\ub}}$. In other words, it could have strongly bias to the mean of training data $\bar{\ub}$. This bias 
could lead to poor generalization of the inverse surrogate model.

\subsubsection{Model-constrained autoencoder inverse-PtO learning approach \mcPOP}
\seclab{mc_OPO}

Conversely, we consider a model-constrained AutoEncoder inverse-PtO learning approach, \mcOPO{}, in which the encoder learns the inverse map and the decoder learn the PtO map. Unlike \pureOPO{},  the encoder output is fed to the PtO model to reproduce observation data. As shown in  \mcPOP{} approach, this model-constrained strategy provides a better inverse solution surrogate model. Additionally,  we learn the decoder by minimizing the misfit between decoder outputs and PtO maps, whose inputs are any outputs from encoder networks during training. The loss functions for \mcOPO{} approach read

\begin{equation}
    \tag{\mcOPO}
    \begin{aligned}
        \Psie^* = \min_{\Psie} & \halfv{1} \nor{\Psie\LRp{\Y} - \U}_{F}^2 + \halfv{\lambda}\nor{\B \circ \F\LRp{\Psie\LRp{\Y}} - \Y}_{F}^2,\\ 
        \Psid^* = \min_{\Psid} & \halfv{1} \nor{\Psid\LRp{\Psie^*\LRp{\Y}} - \B \circ \F\LRp{\Psie^*\LRp{\Y}}}_{F}^2.
    \end{aligned}
\end{equation}
Similarly to previous analysis for other approaches, we apply to the linear problems with linear encoder and decoder neural networks, the loss functions for \mcOPO{} are given as

\begin{equation}
    \eqnlab{mcOPO_linear}
    \begin{aligned}
        {\W_e^*, \bb_e^*} = & \min_{\W_e, \bb_e} \halfv{1} \nor{\W_e \Y + \bb_e \One^T - \U}_{F}^2 + \halfv{\lambda}\nor{\GB \LRp{\W_e \Y + \bb_e \One^T} - \Y}_{F}^2, \\
        {\W_d^*, \bb_d^*} = & \min_{\W_d, \bb_d} \halfv{1} \nor{\W_d \LRp{\W_e^* \Y + \bb_e^* \One^T} +\bb_d \One^T - \GB\LRp{\W_e^* \Y + \bb_e^* \One^T}}_{F}^2. 
    \end{aligned}
\end{equation}
By applying the first optimality condition, the optimal solutions for \cref{eq:mcOPO_linear} as derived in \cref{sect:derivation_mcOPO} are
\begin{equation*}
    \eqnlab{mcOPO_opt}
    \begin{aligned}
        \W_e^*   & = (\Ib+ \lambda {\GB}^T{\GB})^{-1} \left(\bar{\P} \bar{\Y}^{\dagger} + \lambda {\GB}^T \bar{\Y} \bar{\Y}^{\dagger} \right),                                                      \\
        \bb_e^* & = (\Ib+ \lambda {\GB}^T{\GB})^{-1} \LRp{\bar{\ub} + \lambda {\GB}^T \bar{\yb} - \left(\bar{\P} \bar{\Y}^{\dagger} + \lambda {\GB}^T \bar{\Y} \bar{\Y}^{\dagger} \right) \bar{\yb}}, \\
        \W_d^*   & = {\GB} \bar{Z} \bar{Z}^{\dagger}, \quad \quad \quad  \bb_d^* = {\GB} \left( \Ib -  \bar{Z} \bar{Z}^{\dagger} \right) \bar{\zb},
    \end{aligned}
\end{equation*}
where
\begin{equation*}
    \eqnlab{bar_Z_III_opt}
    \begin{aligned}
        \bar{\zb} & = (\Ib+ \lambda {\GB}^T{\GB})^{-1} \left(\bar{\ub} + \lambda {\GB}^T \bar{\yb} \right), \\
        \bar{Z}   & = (\Ib+ \lambda {\GB}^T{\GB})^{-1} \left[ \bar{\P} \bar{\Y}^{\dagger} \bar{\Y} + \lambda {\GB}^T \bar{\Y} \right].
    \end{aligned}
\end{equation*}
It can be seen that the encoder learns exactly the same inverse map as the decoder in \mcPOP{} approach. Therefore, 
we have
$\epsb_{\ubtest}^{\mcOPO} = \epsb_{\ubtest}^{\mcPOP}$. On the other hand, for the learned decoder, we can see that as long as $\bar{Z}$, or equivalent $\bar{\Y}$, is full row rank, the PtO map $\GB$ can be learned well by the decoder. Unlike, the na\"ive data-driven approaches \purePOP{} where we need $\bar{\P}$ is full row rank or \pureOPO{} where it is impossible to construct the exact inverse map, \mcPOP{} can learn exactly the PtO map with only one observation sample, i.e., 
\begin{equation}
    \eqnlab{mcOPO_decoder_PtO}
    \Psid^* \LRp{\ubtest} = \GB \ubtest = \ybtest.
\end{equation}
Indeed, the condition of full row rank $\Y$ can be satisfied in two ways: a sufficient amount of training data is given, or columns of $\bar{\Y}$ are simply randomized observation samples from a single observation sample \cite{MatrixFullRank}. Consequently, we have a zero PtO error estimation, $\epsb_{\ybtest}^{\mcOPO} = 0$ (at least with high probability).

\subsubsection{Model-constrained autoencoder inverse-forward learning approach \mcOPOfull{}}
\seclab{mc_OPOfull}

Observed the advantages of \mcOPO{} approach on learning the PtO map by only one single training sample from \cref{eq:mcOPO_decoder_PtO}, 
we propose a model-constrained AutoEncoder inverse-forward learning approach, \mcOPOfull{}, in which the encoder learns the inverse map and the decoder learn the forward map (parameter to full solution of the PDE). The loss functions for \mcOPOfull{} are

\begin{equation}
    \tag{\mcOPOfull}
    \begin{aligned}
        \Psie^* = \min_{\Psie} & \halfv{1} \nor{\U - \Psie\LRp{\Y}}_{F}^2 + \halfv{\lambda}\nor{\Y - \B \circ \F\LRp{\Psie\LRp{\Y}}}_{F}^2, \\ 
        \Psid^* = \min_{\Psid} & \halfv{1} \nor{\F\LRp{\Psie^*\LRp{\Y}} -\Psid\LRp{\Psie^*\LRp{\Y}}}_{F}^2.
    \end{aligned}
\end{equation}
Following the similar procedure in \cref{sect:mc_OPO} for linear problems using linear neural networks, the optimal solutions for \mcOPOfull{} are
\begin{equation*}
    \eqnlab{mcOPOfull_opt}
    \begin{aligned}
        \W_e^*   & = (\Ib+ \lambda {\GB}^T{\GB})^{-1} \left(\bar{\P} \bar{\Y}^{\dagger} + \lambda {\GB}^T \bar{\Y} \bar{\Y}^{\dagger} \right),                                                      \\
        \bb_e^* & = (\Ib+ \lambda {\GB}^T{\GB})^{-1} ( \bar{\ub} + \lambda {\GB}^T \bar{\yb} - \left(\bar{\P} \bar{\Y}^{\dagger} + \lambda {\GB}^T \bar{\Y} \bar{\Y}^{\dagger} \right) \bar{\yb} ), \\
        \W_d^*   & = {G} \bar{Z} \bar{Z}^{\dagger}, \quad \quad \quad  \bb_d^* = {G} \left( \Ib -  \bar{Z} \bar{Z}^{\dagger} \right) \bar{\zb},
    \end{aligned}
\end{equation*}
where
\begin{equation*}
    \eqnlab{bar_Z_IV_opt}
    \begin{aligned}
        \bar{\zb} & = (\Ib+ \lambda {\GB}^T{\GB})^{-1} \left(\bar{\ub} + \lambda {\GB}^T \bar{\yb} \right),                        \\
        \bar{Z}   & = (\Ib+ \lambda {\GB}^T{\GB})^{-1} \left[ \bar{\P} \bar{\Y}^{\dagger} \bar{\Y} + \lambda {\GB}^T \bar{\Y} \right].
    \end{aligned}
\end{equation*}
It can be seen that the encoder is identical to the one obtained by \mcPOP{}, \mcOPO{}, thus the inverse error estimate is $\epsb_{\ubtest}^{\mcOPOfull} = \epsb_{\ubtest}^{\mcOPO} = \epsb_{\ubtest}^{\mcPOP}$. Meanwhile, the decoder is able to learn the PDEs linear solver $\G$ by only one single observation sample in a similar way that \mcOPO{} approach does, i.e., $\epsb_{\ybfulltest}^{\mcOPOfull} = 0$.

\subsection{Model-constrained Tikhonov autoencoder approaches} 
\seclab{TNet_autoencoder}
As discussed in the na\"ive and model-constrained approaches, \purePOP{}, \pureOPO{}, \mcPOP{}, \mcOPO{}, and \mcOPOfull{}, the performance of learned inverse surrogate models  depends on the training data samples. In our previous work \cite{nguyen2024tnet}, we proposed \texttt{TNet}, model-constrained Tikhonov neural network, which can learn the inverse map with much fewer training samples than na\"ive and model-constrained approaches. Furthermore, shown above for \mcOPO{} and \mcOPOfull{} with linear models, we can learn the exact  PtO and forward map with only one training observation sample. We now propose \TNetAE{} and \TNetAEfull{} approaches which, as will be shown, are able to learn the inverse and PtO/forward maps with only one training sample. Further,  we emphasize that \TNetAE{} and \TNetAEfull{} require only the ``label" part, namely $\yb$, of the training samples and the prior mean of the PoI, $\ub_0$, while no ground truth PoI is needed.\footnote{Tikhonov-type of learning method is thus suitable for most practical situations where we do not have training pairs $\LRp{\ub,\yb}$, but only the label $\yb$ parts.}

\subsubsection{Model-constrained Tikhonov autoencoder inverse-PtO learning approach \TNetAE{}}

The schematic of \TNetAE{} is presented in \cref{fig:AE_net}. A sequential learning strategy is applied to learn the encoder and decoder in two phases. In {\bf Phase 1}, at every epoch during training, we randomize the observational data with noise
$\epsb \sim \mc{N}\LRp{0, \epsilon^2 \LRs{\boldsymbol{\diag}\LRp{{\yb}}}^2}$ which is added to the observation data $\yb$ to generate randomized observation samples. The randomized data is then fed into the encoder network $\Psie$ to predict the inverse solution $\ub^*$. The predicted inverse $\ub^*$ is passed to the PtO map $\B \circ \F$ to predict the observation data $\B\ybfull^{*}$. We minimize the encoder loss $\mc{L}_e$ for the encoder network. In {\bf Phase 2}, we randomize observations and pass through the already-trained encoder network to produce inverse solutions $\ub^*$. Then, $\ub^*$ is treated as inputs to both the decoder network $\Psid$ to produce $\yb^*$ and PtO map to produce $B\boldsymbol{\omega}^*$. The decoder loss $\mc{L}_d$ is then minimized to find optimal decoder parameters. 
In other words, in {\bf Phase 2}, we can consider the encoder as the generative engine that generates physics samples $\ub^*$ to train the PtO surrogate model. The loss functions for \TNetAE{} from \cref{fig:AE_net} are given as

\def\layersep{1.5cm}
\def\nodeinlayersep{0.7cm}
\definecolor{utorange}{RGB}{255,130,0}

\begin{figure}[htb!]
        \centering
        \resizebox{1.\textwidth}{!}{
        \begin{tikzpicture}[
            node distance=\layersep,
            edge/.style={-stealth,shorten >=1pt, draw=black!50,thin},
            neuron/.style={circle,fill=black!25,minimum size=10pt,inner sep=0pt},
            operator/.style={rectangle,fill=green!,minimum height= \nodeinlayersep, minimum width= 0.8 * \layersep, inner sep=0pt, rounded corners},
            input noise/.style={neuron, draw = black, fill=gray!50,minimum size=16},
            input neuron/.style={neuron, draw = black, fill=green!10,minimum size=16},
            output neuron/.style={neuron, draw = black, fill=utorange!50,minimum size=16pt},
            hidden neuron/.style={neuron, draw = black, fill=utorange!10, minimum size=16pt},
            Forward map/.style={operator, draw = black, fill=utorange!100},
            annot/.style={text width=4em, text centered},
            every node/.style={scale=1.0},
            node1/.style={scale=2.0}
        ]
        \foreach \name / \y in {1,...,4}
            {\ifnum \y=3
                    \node (I-\name) at (-.8 * \layersep,-\nodeinlayersep * \y - \nodeinlayersep * 2.) {$\vdots$};
                \else
                    \ifnum \y=4
                        \node[input noise] (I-\name) at (-.8 * \layersep,-\nodeinlayersep * \y - \nodeinlayersep * 2.) {$\epsilon_{\dy}$};
                    \else
                        \node[input noise] (I-\name) at (-.8 * \layersep,-\nodeinlayersep *\y - \nodeinlayersep * 2. ) {$\epsilon_{\y}$};
                    \fi
                \fi}
        
        \foreach \name / \y in {1,...,4}
            {\ifnum \y=3
                    \node (I-\name) at (0,-\nodeinlayersep * \y - \nodeinlayersep * 2.) {$\vdots$};
                \else
                    \ifnum \y=4
                        \node[input neuron] (I-\name) at (0,-\nodeinlayersep * \y - \nodeinlayersep * 2.) {$y_{\dy}$};
                    \else
                        \node[input neuron] (I-\name) at (0,-\nodeinlayersep *\y - \nodeinlayersep * 2. ) {$y_{\y}$};
                    \fi
                \fi}

        \foreach \name / \y in {1,...,4}
            {\ifnum \y=3
                    \node (O-\name) at (4*\layersep,-\nodeinlayersep *\y - \nodeinlayersep * 2.) {$\vdots$};
                \else
                    \ifnum \y=4
                        \node[input neuron] (O-\name) at (4*\layersep,-\nodeinlayersep *\y - \nodeinlayersep * 2.) {$u_{\du}^*$};
                    \else
                        \node[input neuron] (O-\name) at (4*\layersep,-\nodeinlayersep *\y - \nodeinlayersep * 2.) {$u_{\y}^*$};
                    \fi
                \fi}

        \node[Forward map] (FM) at (5.5*\layersep,-\nodeinlayersep * 7.) {$\F$};

        \foreach \name / \y in {1,...,4}
            {\ifnum \y=3
                    \node (obs_p-\name) at (8*\layersep,-\nodeinlayersep *\y - \nodeinlayersep * 4.5) {$\vdots$};
                \else
                    \ifnum \y=4
                        \node[input neuron] (obs_p-\name) at (8*\layersep,-\nodeinlayersep *\y - \nodeinlayersep * 4.5) {$\omega_{\dw}^{*}$};
                    \else
                        \node[input neuron] (obs_p-\name) at (8*\layersep,-\nodeinlayersep *\y - \nodeinlayersep * 4.5) {$\omega_{\y}^{*}$};
                    \fi
                \fi}

        \foreach \name / \y in {1,...,4}
            {\ifnum \y=3
                    \node (obs_p-\name) at (8*\layersep,-\nodeinlayersep *\y + \nodeinlayersep * -.5) {$\vdots$};
                \else
                    \ifnum \y=4
                        \node[input neuron] (obs_p2-\name) at (8*\layersep,-\nodeinlayersep *\y + \nodeinlayersep * -.5 ) {$y_{\dy}^{*}$};
                    \else
                        \node[input neuron] (obs_p2-\name) at (8*\layersep,-\nodeinlayersep *\y + \nodeinlayersep * -.5) {$y_{\y}^{*}$};
                    \fi
                \fi}

        \newcommand \Nhidden{3}
        \foreach \N in {1,...,\Nhidden} {
                \foreach \y in {1,...,5} { 
                        \ifnum \y=4
                            \node at (\N*\layersep,-\y*\nodeinlayersep-1.5*\nodeinlayersep) {$\vdots$};
                        \else
                            \node[hidden neuron] (H\N-\y) at (\N*\layersep,-\y*\nodeinlayersep-1.5*\nodeinlayersep) {$\sigma$};
                        \fi
                    }
            }

        \foreach \N in {1,...,\Nhidden} {
                \foreach \y in {1,...,5} { 
                        \ifnum \y=4
                            \node at (\N*\layersep + 4*\layersep,-\y*\nodeinlayersep - 0*\nodeinlayersep) {$\vdots$};
                        \else
                            \node[hidden neuron] (D\N-\y) at (\N*\layersep + 4*\layersep ,-\y*\nodeinlayersep + 0*\nodeinlayersep) {$\sigma$};
                        \fi
                    }
            }

        \foreach \source in {1,2,4}
        \foreach \dest in {1,...,3,5} 
        \draw[edge] (I-\source) -- (H1-\dest);

        \foreach [remember=\N as \lastN (initially 1)] \N in {2,...,\Nhidden}
        \foreach \source in {1,...,3,5} 
        \foreach \dest in {1,...,3,5} 
        \draw[edge] (H\lastN-\source) -- (H\N-\dest);

        \foreach \source in {1,...,3,5} 
        \foreach \dest in {1,2,4}
        \draw[edge] (H\Nhidden-\source) -- (O-\dest);

        \foreach \source in {1,2,4}
        \draw[edge] (O-\source) -- (5.1*\layersep,-\nodeinlayersep * 7.);

        \foreach \source in {1,2,4}
        \draw[edge] (FM) -- (obs_p-\source);

        \foreach \source in {1,2,4}
        \foreach \dest in {1,...,3,5} 
        \draw[edge] (O-\source) -- (D1-\dest);

        \foreach [remember=\N as \lastN (initially 1)] \N in {2,...,\Nhidden}
        \foreach \source in {1,...,3,5} 
        \foreach \dest in {1,...,3,5} 
        \draw[edge] (D\lastN-\source) -- (D\N-\dest);

        \foreach \source in {1,...,3,5} 
        \foreach \dest in {1,2,4}
        \draw[edge] (D\Nhidden-\source) -- (obs_p2-\dest);

        \node[rectangle,draw = black,fill=orange!10,minimum height= \nodeinlayersep, minimum width= 0.8 * \layersep, rounded corners] (L_misfit) at (7.4*\layersep, 1.*\nodeinlayersep) {\textbf{Phase 2}: Loss $\mathcal{L}_d:= \half \nor{\yb^{*} - \B \ybfull^{*}}_2^2$};

        \draw [decorate, decoration = {calligraphic brace}, thin] (8.25*\layersep,-1.2*\nodeinlayersep) -- (8.25*\layersep,-4.8*\nodeinlayersep);

        \draw[edge,thin] (8.35*\layersep,-3.0*\nodeinlayersep) -- (8.9*\layersep, -3*\nodeinlayersep) -- (8.9*\layersep, .5*\nodeinlayersep);

        \draw [decorate, decoration = {calligraphic brace}, thin] (8.25*\layersep,-5.2*\nodeinlayersep) -- (8.25*\layersep,-8.8*\nodeinlayersep);
        \draw[edge,thin] (8.35*\layersep,-6.95*\nodeinlayersep) -- (9*\layersep, -6.95*\nodeinlayersep) -- (9*\layersep, .5*\nodeinlayersep);
        \draw[edge,thin] (8.35*\layersep,-7.05*\nodeinlayersep) -- (9*\layersep, -7.05*\nodeinlayersep) -- (9*\layersep, -10.*\nodeinlayersep);

        \draw [decorate, decoration = {calligraphic brace}, thin] (4.15*\layersep,-6.5*\nodeinlayersep) -- (3.85*\layersep,-6.5*\nodeinlayersep);
        \draw[edge,thin] (4*\layersep,-6.6 *\nodeinlayersep) -- (4*\layersep, -10 *\nodeinlayersep);

        \draw [decorate, decoration = {calligraphic brace}, thick] (.85*\layersep,-2.*\nodeinlayersep) -- (3.15*\layersep,-2.*\nodeinlayersep) node[pos=0.5,above=0.1cm,black]{Encoder $\Psie$};

        \draw [decorate, decoration = {calligraphic brace}, thick] (4.85*\layersep,-0.5*\nodeinlayersep) -- (7.15*\layersep,-.5*\nodeinlayersep) node[pos=0.5,above=0.1cm,black]{Decoder $\Psid$};

        \draw [decorate, decoration = {calligraphic brace, mirror}, thick] (5*\layersep,-9.*\nodeinlayersep) -- (8.25*\layersep,-9.*\nodeinlayersep) node[pos=0.5,above=0.cm,black]{Model-constrained};

        \node[rectangle,draw = black,fill=orange!10,minimum height= \nodeinlayersep, minimum width= 0.8 * \layersep, rounded corners] (L_misfit_2) at (6.5*\layersep, -10.5*\nodeinlayersep) {\textbf{Phase 1}: Loss $\mathcal{L}_e:= \halfv{1} \nor{\ub_0 - \ub^*}_2^2 + \halfv{\lambda} \nor{\yb - \B \ybfull^{*}}_2^2$};

        \node[rectangle,fill=gray!40,minimum height= \nodeinlayersep, minimum width= 1.6 * \layersep, rounded corners] (random_noise_engine) at (1.4*\layersep, -10.*\nodeinlayersep) {\begin{tabular}{c} Random noise \vspace*{.2cm} \\  $\epsb \sim \mc{N} \LRp{\mathbf{0}, \epsilon^2 \LRs{\boldsymbol{\diag}\LRp{{\yb}}}^2}$ \end{tabular}};
        \draw[edge,thin] (random_noise_engine) -- (-.8*\layersep, -10. *\nodeinlayersep) -- (-.8*\layersep, -6.6 *\nodeinlayersep);

        \draw[line width=3pt] (-0.55*\layersep, -4.6*\nodeinlayersep) -- (-0.25*\layersep, -4.6*\nodeinlayersep);

        \draw[line width=3pt] (-0.4*\layersep, -4.3*\nodeinlayersep) -- (-0.4*\layersep, -4.9*\nodeinlayersep);

    \end{tikzpicture}
        }
    \caption{
        The schematic of \TNetAE{} approach. A sequential learning strategy is applied to learn the encoder and decoder in two phases. In {\bf Phase 1}, at every epoch during training, we randomize the observation data with noise  $\epsb \sim \mc{N}\LRp{0, \epsilon^2 \LRs{\boldsymbol{\diag}\LRp{{\yb}}}^2}$ which is added to the observation data $\yb$ to generate randomized observation samples. The randomized data is then fed into the encoder network $\Psie$ to predict the inverse solution $\ub^*$. The predicted inverse $\ub^*$ is passed to the PtO map $\B \circ \F$ to predict the observation data $\B\ybfull^{*}$. We minimize the encoder loss $\mc{L}_e$ for the optimal encoder network. In {\bf Phase 2}, we randomize observations and pass through the pre-trained encoder network to produce inverse solutions $\ub^*$. Then, $\ub^*$ is treated as inputs to both the decoder network $\Psid$ to produce $\yb^*$ and PtO map to produce $B\boldsymbol{\omega}^*$. The decoder loss $\mc{L}_d$ is then minimized to find optimal decoder parameters. 
    }
    \figlab{AE_net}
\end{figure}
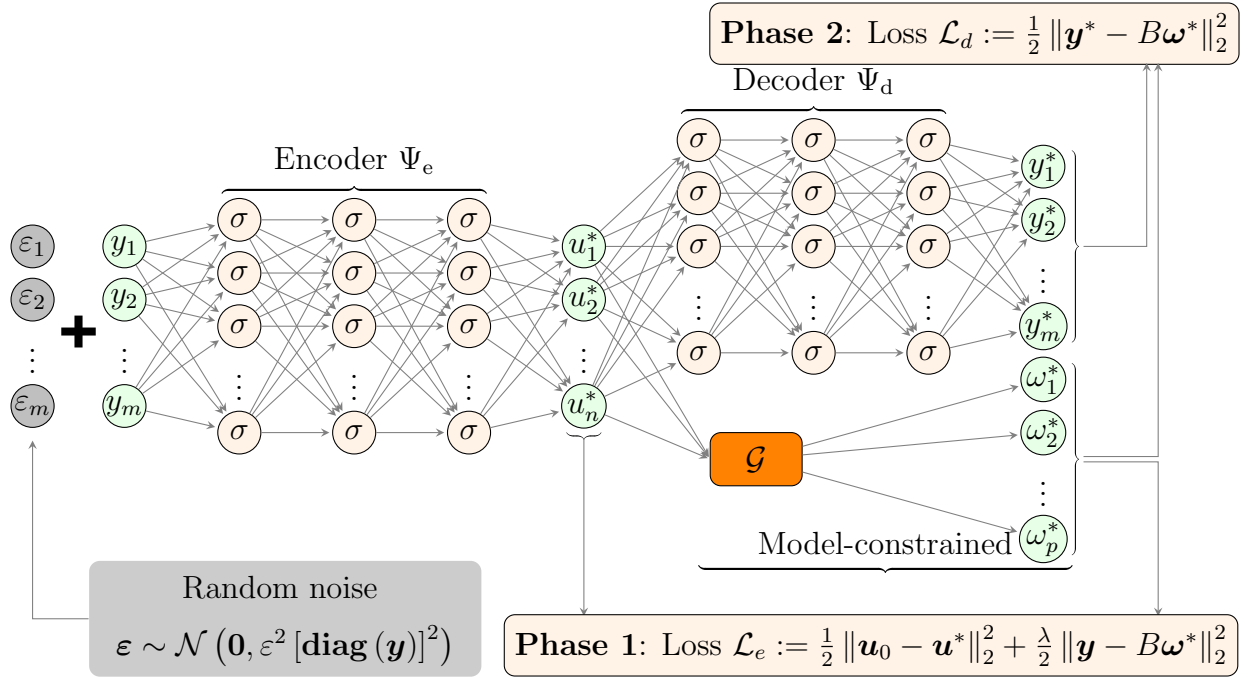

\begin{equation}
    \tag{\TNetAE}
    \begin{aligned}
        \Psie^* = \min_{\Psie} & \halfv{1} \nor{\Psie\LRp{\Y} - \ub_0 \One^T}_{F}^2
        +  \halfv{\lambda} \nor{\B \circ \F\LRp{\Psie\LRp{\Y}} - \Y}_{F}^2, \\ 
        \Psid^* = \min_{\Psid} & \halfv{1} \nor{\Psid\LRp{\Psie^*\LRp{\Y}} - \B \circ \F\LRp{\Psie^*\LRp{\Y}}}_{F}^2.
    \end{aligned}
    \label{eq:TNetAE}
\end{equation}
For linear problems and linear networks (see previous sections), the loss functions for \TNetAE{} becomes
\begin{equation}
    \eqnlab{TNetAElinear}
    \begin{aligned}
        {\W_e^*, \bb_e^*} = & \min_{\W_e, \bb_e} \halfv{1} \nor{\W_e \Y + \bb_e \One^T - \ub_0 \One^T}_{F}^2
        + \halfv{\lambda}\nor{\GB\LRp{\W_e \Y + \bb_e \One^T} - \Y}_{F}^2, \\
        {\W_d^*, \bb_d^*} = & \min_{\W_d, \bb_d} \halfv{1} \nor{\Psid\LRp{\W_e^* \Y + \bb_e^* \One^T} - \GB\LRp{\W_e^* \Y + \bb_e^* \One^T}}_{F}^2.
    \end{aligned}
\end{equation}
The main difference in \TNetAE{} compared to \mcOPO{} is now $\P = \ub_0 \One^T$. It is straightforward to achieve the optimal solutions for \cref{eq:TNetAElinear} (as derived in \cref{sect:derivation_TNetAE}): 
\begin{equation*}
    \eqnlab{TNetAE_opt}
    \begin{aligned}
        \W_e^*   & = (\Ib+ \lambda {\GB}^T{\GB})^{-1} \left(\lambda {\GB}^T \bar{\Y} \bar{\Y}^{\dagger} \right),                                                               \\
        \bb_e^* & = (\Ib+ \lambda {\GB}^T{\GB})^{-1} ( \ub_0 + \lambda {\GB}^T \bar{\yb} - \lambda {\GB}^T \bar{\Y} \bar{\Y}^{\dagger} \bar{\yb} ),                            \\
        \W_d^*   & = {\GB} \bar{Z} \bar{Z}^{\dagger}, \quad \quad \quad  \bb_d^* = {\GB} \left( \Ib -  \bar{Z} \bar{Z}^{\dagger} \right) \bar{\zb},
    \end{aligned}
\end{equation*}
where
\begin{equation*}
    \begin{aligned}
        \bar{\zb} & = (\Ib+ \lambda {\GB}^T{\GB})^{-1} \left(\ub_0 + \lambda {\GB}^T \bar{\yb} \right), \\
        \bar{Z}   & = (\Ib+ \lambda {\GB}^T{\GB})^{-1} \left(\lambda {\GB}^T \bar{\Y} \right).
    \end{aligned}
\end{equation*}
The \TNetAE{} inverse solution obtained by the encoder network for an observation test sample, $\ybtest$, is given by
\begin{equation*}
    \ub^{\TNetAE} = \LRp{ \Ib + \lambda {\GB}^T \GB }^{-1} \LRp{ \ub_0 + \lambda {\GB}^T \bar{\yb} + \lambda {\GB}^T \bar{\Y} \bar{\Y}^{\dagger} \LRp{\ybtest - \bar{\yb}}}
\end{equation*}
which is exactly the inverse solution of the following linear Tikhonov regularization problem
\begin{equation*}
    \min_{\ub}  \halfv{1} \nor{\ub - \ub_0}_{2}^2 + \halfv{\lambda} \nor{\GB \ub - \ybtest}_{2}^2.
\end{equation*}
Therefore, the error of inverse solutions obtained from the  \TNetAE{} encoder network  is
\begin{multline*}
    \epsb_{\ubtest}^{\TNetAE}  = \nor{\ub^{\TNetAE} - \ub^\text{test}}_2^2 \\ 
    = \nor{ \LRp{ \Ib + \lambda {\GB}^T \GB }^{-1} \LRp{ \LRp{\ub_0 - \ub^\text{test}} + \lambda {\GB}^T  \LRp{\Ib - \bar{\Y} \bar{\Y}^{\dagger}} \LRp{\ybtest - \bar{\yb}}} }_2^2.
\end{multline*}
It can be seen that as long as $\bar{\Y}$ is full row rank, which can be obtained with high probability by having large number of randomization observation samples \cite{MatrixFullRank}, the error can be written as (again with high probability)
\begin{equation*}
    \epsb_{\ubtest}^{\TNetAE} = \nor{ \LRp{ \Ib + \lambda {\GB}^T \GB }^{-1} { \LRp{\ub_0 - \ub^\text{test}}} }_2^2 \leq \nor{{\ub_0 - \ub^\text{test}}}_2^2,
\end{equation*}
independent of the single training sample.
On the other hand, similar to the decoder of \mcOPO{}, it can be shown that the \TNetAE{} decoder network is able to learn the PtO map exactly, $\epsb_{\ybtest}^{\TNetAE} = 0$, with only one observation training sample.

\subsubsection{Model-constrained Tikhonov autoencoder inverse-forward learning approach \TNetAEfull{}}

Inspired by advantages of \TNetAE{} and \mcOPOfull{}, we propose a new approach, \TNetAEfull{}, where the encoder learns the inverse Tikhonov regularization solver and the decoder learns the forward map (parameter to solution of PDEs). We construct the loss for \TNetAEfull{} as

\begin{equation}
    \tag{\TNetAEfull}
    \begin{aligned}
        \Psie^* = \min_{\Psie} & \halfv{1} \nor{\Psie\LRp{\Y} - \ub_0 \One^T}_{F}^2
        +  \halfv{\lambda} \nor{\B \circ \F\LRp{\Psie\LRp{\Y}} - \Y}_{F}^2, \\
        \Psid^* = \min_{\Psid} & \halfv{1} \nor{\Psid\LRp{\Psie\LRp{\Y}} - \F\LRp{\Psie\LRp{\Y}}}_{F}^2.
    \end{aligned}
\end{equation}
Analogous to other approaches, in the context of linear problems using linear autoencoder networks,
the optimal solutions for encoder and decoder networks are
\begin{equation*}
    \eqnlab{TNetAEfull_opt}
    \begin{aligned}
        \W_e^*   & = (\Ib+ \lambda {\GB}^T{\GB})^{-1} \left(\lambda {\GB}^T \bar{\Y} \bar{\Y}^{\dagger} \right),                                                                 \\
        \bb_e^* & = (\Ib+ \lambda {\GB}^T{\GB})^{-1} ( \ub_0 + \lambda {\GB}^T \bar{\yb} - \lambda {\GB}^T \bar{\Y} \bar{\Y}^{\dagger} \bar{\yb} ),                              \\
        \W_d^*   & = {G} \bar{Z} \bar{Z}^{\dagger}, \quad \quad \quad  \bb_d^* = {G} \left( \Ib -  \bar{Z} \bar{Z}^{\dagger} \right) \bar{\zb},
    \end{aligned}
\end{equation*}
where
\begin{equation*}
    \begin{aligned}
        \bar{\zb} & = (\Ib+ \lambda {\GB}^T{\GB})^{-1} \left(\ub_0 + \lambda {\GB}^T \bar{\yb} \right),  \\
        \bar{Z}   & = (\Ib+ \lambda {\GB}^T{\GB})^{-1} \left(\lambda {\GB}^T \bar{\Y} \right).
    \end{aligned}
\end{equation*}
It can be seen that the encoder network of \TNetAEfull{} is exactly the same as encoder network of \TNetAE{},  and thus $\epsb_{\ubtest}^{\TNetAEfull} = \epsb_{\ubtest}^{\TNetAE}$. Similar to the decoder of \mcOPOfull{}, the decoder network of \TNetAEfull{} is able to learn the forward map exactly with only one observation training sample, i.e., $\epsb_{\ybfulltest}^{\TNetAEfull} = 0$.

For clarity, we summarize the error estimations of the proposed approaches in \cref{tab:compare_approaches} and whether or not an approach can learn the PtO/forward and inverse map with only one training sample.

\begin{table}[htb!]
\centering
\caption{Summary of test error estimations obtained from various approaches for solving linear problems using linear encoder and decoder networks.}
\tablab{compare_approaches}
\resizebox{1.\textwidth}{!}{
\begin{tblr}{
  width = 1.05\linewidth,
  colspec = {X[c,m,1] X[c,m,1.4] X[c,m,2.2] X[c,m,1.4]},
  cell{2}{3} = {r=2}{},
  cell{4}{3} = {r=2}{},
  vlines,
  hline{1-2,4,6-7} = {-}{},
  hline{3,5} = {1-2,4}{},
}
Approaches    & {PtO/forward error \\ $\epsb_{\ybtest}$/ $\epsb_{\ybfulltest}$} & {Inverse error \\  $\epsb_{\ubtest}$}                                                           & {Ability to learn \\ with 1 sample} \\
{\purePOP} & 0                         & $\nor{ \LRp{\bar{\P} \bar{\Y}^\dagger \GB -\Ib} \LRp{\ubtest - \bar{\ub}}}_2^2$   & None                            \\
{\pureOPO} & inevitably \textgreater{} 0           &                                                                                   & None                            \\
{\mcPOP}        & 0                         & $\le \nor{ \LRp{\bar{\P} \bar{\Y}^\dagger \GB -\Ib} \LRp{\ubtest - \bar{\ub}}}_2^2$ & Only PtO                            \\
{\mcOPO(\textcolor{black}{\texttt{-Full}})}        & 0                         &                                                                                   & Only PtO/forward                            \\
{\TNetAE(\textcolor{black}{\texttt{-Full}})}        & 0                         & $\le \nor{{\ub^\text{test} - \ub_0}}_2^2$                                            & All                           
\end{tblr}
}
\end{table}

\subsection{Simultaneous training strategy for encoder and decoder networks}
\seclab{simultaneous_training}

In this section, we examine the drawbacks of simultaneous training compared to sequential training strategies. Although these drawbacks are common across all methodologies, we provide an analysis only for the \purePOP{} framework, as it provides a clear picture of the drawbacks. Specifically, we analyze a case where both the encoder and decoder neural networks of \purePOP{} undergo simultaneous training using the same training dataset $\LRp{\P, \Y}$. In this scenario, we introduce an additional hyperparameter $\beta$ to achieve an appropriate balance between the encoder and decoder loss functions. The loss function can be expressed as
\begin{equation}
    \eqnlab{purePOP_simultaneous}
    {\Psie^*, \Psid^*} = \min_{\Psie, \Psid} \halfv{1} \nor{\Psie\LRp{\P} - \Y}_{F}^2 + \halfv{\beta}\nor{\Psid\LRp{\Psie\LRp{\P}} - \P}_{F}^2,
\end{equation}
To gain insights into how different the simultaneous training strategy is compared to the sequential training counterpart, we analyze the same linear problems using linear encoder and decoder networks. 
By deriving the first optimality condition for \cref{eq:purePOP_simultaneous}, we can obtain the optimal weight and bias for the encoder and decoder networks as follows (see \cref{sect:derivation_purePOP_together} for full derivation):
\begin{equation*}
    \begin{aligned}
        & \W_e^*   =\left( \Ib + \beta {\W^*}_d^T \W_d^* \right)^{-1} \left( \GB \bar{\P} \bar{\P}^{\dagger} + \beta  {\W^*}_d^T \bar{\P} \bar{\P}^{\dagger} \right), && 
        & \bb_e^* =\bar{\yb} - \W_e^* \bar{\ub}, \\
        & \W_d^*  = \bar{\P} \LRp{\W_e^* \bar{\P}}^\dagger, &&
        & \bb_d^* = \bar{\ub} - \W_d^* \bar{\yb}, \\
        & \W_e^* \bar{\P} \bar{\P}^T = \W_e^* \W_d^* \W_e^* \bar{\P} \bar{\P}^T
    \end{aligned}
\end{equation*}
The first drawback emerges from the difficulty in deriving closed-form solutions for the optimal parameters $\W_e^*$, $\bb_e^*$, $\W_d^*$, and $\bb_d^*$ since $\W_e^*$ is a nonlinear function of $\W_d^*$. In other words, even with linear problems, we might need to use some iterative algorithm to obtain a solution.
Furthermore, even with sufficient training data, such that $\bar{\P} \bar{\P}^\dagger = \Ib$, the encoder network  becomes
\begin{equation*}
    \W_e^* = \LRp{\Ib + \beta {\W^*}_d^T \W_d^*}^{-1} \LRp{\GB + \beta {\W^*}_d^T}, \quad \bb_e^* = \bar{\yb} - \W_e \bar{\ub},
\end{equation*} 
and in this case, it is still not clear how to train the optimal encoder to recover the exact linear PtO map $\GB$. 
As a result, the decoder accuracy will be affected by the outputs from the inaccuracy of the encoder network. The second drawback of the simultaneous training strategy is that tuning the hyperparameter $\beta$ for the encoder loss and decoder loss is required and problem-dependent, in addition to having a larger training problem to solve.
In a special case of $\beta = 1$, one of the optimal solutions $\W_e^*$, $ \bb_e^*$, $ \W_d^*$, $ \bb_d^*$ for \cref{eq:purePOP_simultaneous} is indeed identical to the one obtained by the sequential training strategy in \cref{eq:purePOP_opt},
\begin{equation*}
    \eqnlab{purePOP_simulateneous_specialcase_opt}
    \W_e^* = \GB \bar{\P} \bar{\P}^\dagger, \quad \bb_e^* = \GB \LRp{ \Ib - \bar{\P} \bar{\P}^\dagger }\bar{\ub}, \quad  
        \W_d^* = \bar{\P} \bar{\Y}^{\dagger}, \quad  \bb_d^* = \bar{\ub} - \bar{\P} \bar{\Y}^{\dagger} \bar{\yb}.
\end{equation*}
Thus, the simultaneous training strategy (even with the identical linear setting), the encoder is not guaranteed to recover the exact PtO map. We also present a similar derivation for \pureOPO{} in \cref{sect:derivation_pureOPO_together} which shows two similar drawbacks of simultaneous training strategy.




\subsection{Some theoretical justification for the generalization capability with randomization for nonlinear problems}
\label{sect:nonlinearProbabilistic}
In this section, we provide theoretical justification for why randomization could facilitate accurate generalization even with one initial training sample. We shall focus on the sequential \cref{eq:TNetAE} approach in \cref{sect:TNet_autoencoder}, but our discussion is applicable to the other methods as well. {\em Recall from our previous work \cite[Theorem 4.3]{nguyen2024tnet} that randomization induces a regularization, with the noise variance as the regularization parameter, on the regularity of the encoder neural network $\Psie^*$ (the first subequation of \cref{eq:TNetAE}) and its difference with the true inverse operator up to second order derivatives. } We now present some further theoretical justification from a probabilistic angle. 

With randomization, the randomized data is distributed as  
\begin{equation*}
    \yb^{rand} \sim \mc{N}\LRp{\yb, \epsilon^2 \LRs{\boldsymbol{\diag}\LRp{{\yb}}}^2}, 
\end{equation*}
where $\yb$ is the data before randomization. We assume that the prior distribution of the PoI\footnote{For the simplicity of the presentation we use identity covariance here, but the result is also valid for arbitrary prior covariance matrix.} is a Gaussian:
\begin{equation*}
    \ub \sim \mc{N}\LRp{\ub_0, \I},
\end{equation*}
which is push-forwarded to the underlying distribution of the observational data $\yb^{obs}$ via the forward map $\F \LRp{\ub}$ as\footnote{It typically comes with some additive noise as in \cref{eq:noisyObs}, but for the clarity, we ignore it here.}
\begin{equation*}
    \yb^{obs} \sim  \B \circ\F_\#\mc{N}\LRp{\ub_0, \I}.
\end{equation*}

We conclude that {\em if the randomization is such that 
the true data distribution $\B \circ\F_\#\mc{N}\LRp{\ub_0, \I}$ is absolutely continuous with respect to the randomized data distribution $\mc{N}\LRp{\yb, \epsilon^2 \LRs{\boldsymbol{\diag}\LRp{{\yb}}}^2}$ \textemdash that is, the support of the former is contained in the support of the latter\textemdash randomization could explore unseen data/information that is not contained in the original data set (be it one training sample or more).} Together with the aforementioned regularization effect to ensure the encoder to closely matches the inverse map (the first term of the first subequation in \cref{eq:TNetAE}), and the push forward measure $\Psie^*{_\#}\mc{N}\LRp{\yb, \epsilon^2 \LRs{\boldsymbol{\diag}\LRp{{\yb}}}^2}$ to be similar to the prior distribution of PoI $\mc{N}\LRp{\ub_0, \I}$ so that the composition of the push forward measures $\B \circ\F_\#\Psie^*{_\#}\mc{N}\LRp{\yb, \epsilon^2 \LRs{\boldsymbol{\diag}\LRp{{\yb}}}^2}$ is close to the randomized data distribution $\mc{N}\LRp{\yb, \epsilon^2 \LRs{\boldsymbol{\diag}\LRp{{\yb}}}^2}$ (the second term of the first subequation in \cref{eq:TNetAE}). Here, the closeness is in the minimal Kullback-Leibler (KL) divergence sense. When that happens, the true data distribution $\B \circ\F_\#\mc{N}\LRp{\ub_0, \I}$ is absolutely continuous with respect to $\B \circ\F_\#\Psie^*{_\#}\mc{N}\LRp{\yb, \epsilon^2 \LRs{\boldsymbol{\diag}\LRp{{\yb}}}^2}$. In other words, the randomized data distribution, when push-forwarded through the encoder $\Psie^*$ and then the forward map, has minimal KL divergence change so as to cover the true data distribution. 

Next, the second optimization problem (the second subequation in  \cref{eq:TNetAE}) in the  \TNetAE{} 
approach is to ensure that the randomized data distribution, when push-forwarded through the encoder $\Psie^*$ and then the decoder $\Psid^*$, is indifferent (again in the minimal KL divergence) to $\B \circ\F_\#\Psie^*{_\#}\mc{N}\LRp{\yb, \epsilon^2 \LRs{\boldsymbol{\diag}\LRp{{\yb}}}^2}$. In other words, with appropriate conditions on the autoencoder and the forward map, it can be shown that the optimization problem in the second subequation in  \cref{eq:TNetAE} is equivalent to minimizing the KL divergence between $\Psid^*{_\#}\Psie^*{_\#}\mc{N}\LRp{\yb, \epsilon^2 \LRs{\boldsymbol{\diag}\LRp{{\yb}}}^2}$ and $\B \circ\F_\#\Psie^*{_\#}\mc{N}\LRp{\yb, \epsilon^2 \LRs{\boldsymbol{\diag}\LRp{{\yb}}}^2}$. As argued above, since the true data distribution $\B \circ\F_\#\mc{N}\LRp{\ub_0, \I}$ is absolutely continuous with the latter, so is it with the former. It follows that \TNetAE{} can generalize well when equipped with the right amount of random noise level. 

What remains is to estimate such a right random level for a particular problem under consideration. This is clearly problem-dependent as it depends on the forward map and the architecture of the autoencoder. Numerically (see \cref{fig:Heat_noise_level} and \cref{fig:2D_NS_noise_level}), we observe the generalization capability is robust to a wide range of noise level. This could be due to the compactness \cite{Bui-ThanhGhattas12f,Bui-ThanhGhattas12,Bui-ThanhGhattas12a} nature of the forward map.
Below we provide a theoretical insight on why too small or too large noise levels may not help improve generalization capability of \TNetAE{}.
Two observations/comments are in order.
\begin{enumerate}
    \item If the true data distribution $\B \circ\F_\#\mc{N}\LRp{\ub_0, \I}$ is not absolutely continuous with respect to the randomized data distribution $\mc{N}\LRp{\yb, \epsilon^2 \LRs{\boldsymbol{\diag}\LRp{{\yb}}}^2}$, which corresponds to the small noise levels, then the generalization capability of our approach is limited. 
    \item If the true data distribution $\B \circ\F_\#\mc{N}\LRp{\ub_0, \I}$ is absolutely continuous with respect to the randomized data distribution $\mc{N}\LRp{\yb, \epsilon^2 \LRs{\boldsymbol{\diag}\LRp{{\yb}}}^2}$, but the noise level is too high, then the generalization capability of our approach is also limited. 
\end{enumerate}

The poor generalization with the aforementioned two extreme scenarios will be numerically observed for both heat and Navier-Stokes equation in \cref{fig:Heat_noise_level} and \cref{fig:2D_NS_noise_level}, respectively. The reason for both cases is that the trained network does not see sufficient information covered in the true data distribution $\B \circ\F_\#\mc{N}\LRp{\ub_0, \I}$. 
{\em This is a direct consequence of the concentration of measure phenomena.} In particular, with high probability, the distance between the randomized data $\yb^{rand}$ and the observational data $\yb^{obs}$ scales like \cite{Vershynin_2018}
\begin{equation*}
    \nor{\yb^{rand}- \yb^{obs}} \approx \epsilon\sqrt{m}\nor{\yb}.
\end{equation*}
Thus, if $\epsilon$ is too small, the randomized data is concentrated around the initial data $\yb$. On the other hand, when $\epsilon$ is large, the randomized data is concentrated in the neighborhood of a hypersphere shell of radius approximately $\epsilon\sqrt{m}\nor{\yb}$ away from the initial data. In either case, the randomized data is not sufficiently rich for generalization purposes. 

We conclude this section with a remark that the above theoretical arguments can be made rigorous with additional assumptions on the forward map, the training procedure, and the autoencoder architecture. Since such a rigorous analysis does not provide additional insights, but simply an exercise,  we choose not to pursue it here.

\section{Numerical results}
\seclab{numerical_results}

We begin by outlining the training and testing configurations for our proposed approaches for solving forward and inverse problems, focusing on two PDEs: the heat equation in  \cref{sect:Heat_problem} and the Navier-Stokes equation in  \cref{sect:2D_NS}. \cref{tab:Summary_training_params} summarizes the specifications for neural network architectures, training settings, data sets, etc.

\begin{table}[htb!]
    \centering
    \caption{Summary of training parameters for learning forward and inverse problems in \cref{sect:Heat_problem} and \cref{sect:2D_NS}.}
    \begin{tabular}{|l l c |}
        \hline
        \multirow{5}{*}{Network}   & Architecture        & 1 layer with 5000 neurons         \\ \cline{2-3} \\ [-14pt]
                                   & Activation function & ReLU                              \\ \cline{2-3} \\ [-14pt]
                                   & Weight initializer  & $\mc{N}\LRp{0, 0.02}$             \\ \cline{2-3} \\ [-14pt]
                                   & Bias initializer    & $\bs{0}$                          \\ \cline{2-3} \\ [-14pt]
                                   & Random seed         & 100                               \\ \hline      \\ [-14pt]
        \multirow{3}{*}{Training}  & Optimizer           & \texttt{ADAM}                     \\ \cline{2-3} \\ [-14pt]
                                   & Learning rate       & $10^{-3}$                         \\ \cline{2-3} \\ [-14pt]
                                   & Batch size          & $n_t = 100$                             \\ \hline      \\ [-14pt]
        \multirow{4}{*}{Data}      & Train data          & 100 samples (drawn independently) \\ \cline{2-3} \\ [-14pt]
                                   & Test data           & 500 samples (drawn independently) \\ \cline{2-3} \\ [-14pt]
                                   & Train random seed   & 18                                \\ \cline{2-3} \\ [-14pt]
                                   & Test random seed    & 28                                \\ \hline      \\ [-14pt]
        \multirow{1}{*}{Precision} &                     & Double precision                  \\ \hline
    \end{tabular}
    \tablab{Summary_training_params}
\end{table}

{\bf Data generation.}
Our training dataset comprises 100 paired samples of PoIs and their corresponding clean observations $\yb^\text{clean}$, which are subsequently corrupted to $\ybtest$ according to \cref{eq:noise_corruption}. 
We consider two training cases: Case I - single-sample training; and Case II - 100-sample training. 
In case I, we replicate the randomly chosen sample to create a set of 100 identical samples. In case II, we use all different 100 training samples. 
For evaluation purposes, we utilize a test dataset of 500 sample pairs, corrupted with noise level $\delta$ as \cref{eq:noise_corruption}, to benchmark different approaches. The Tikhonov inverse solutions are computed using the standard BFGS optimization algorithm \cite{nocedal1999numerical} available in \texttt{Jax-Optax} \cite{jax2018github}. 

{\bf Two stages of adding noise.}
For each sample regardless of case I or case II, we corrupt noise at two stages in our experimental framework. Initially, we introduce a noise level $\delta$ to the clean synthetic data to simulate real-world observations:
\begin{equation}
\eqnlab{noise_corruption}
\ybtest = \yb^{\text{clean}} + \boldsymbol{\delta} \odot \yb^{\text{clean}} , \quad \boldsymbol{\delta} \sim  \mc{N}(0, \delta^2 \textbf{I}),
\end{equation}
where $\odot$ represents element-wise multiplication, $\yb^{\text{clean}}$ denotes the noise-free data generated by the forward solver, and $\textbf{I} \in \R^{\dy \times \dy}$ is the  $\dy \times \dy$ identity matrix. Subsequently, we perform the data randomization technique. Specifically, for each epoch during  training, for \mcOPO, \mcOPOfull, \TNetAE{} and \TNetAEfull{} methods, we apply an additional noise level $\epsilon$  on-the-fly to the simulated real-world data $\ybtest$:
\begin{equation}
\eqnlab{noise_training}
\tilde{\yb}^\text{rand} = \ybtest + \boldsymbol{\epsilon}, \quad \boldsymbol{\epsilon} \sim  \mc{N}(0, \epsilon^2 \LRs{\boldsymbol{\diag}{\LRp{\ybtest}}}^2),
\end{equation}
where $\LRs{\cdot}^2$ denotes the elementwise square operation.
In practice, we generate a randomized sample as follows, drawing standard Gaussian random samples $\zeta$ and then point-wise multiplication with $\ybtest$ to generate noise realizations, i.e.,  
\begin{equation*}
\tilde{\yb}^\text{rand} = \ybtest + \boldsymbol{\zeta} \odot \ybtest, \quad \boldsymbol{\zeta} \sim  \mc{N}(0, \epsilon^2 \textbf{I}).
\end{equation*}
Specifically, we draw 100 new noise realizations $\epsb$ per epoch and add them to 100 observation samples in both case I and case II. For the \purePOP{} and \mcPOP{} methods, we retain only the base noise level $\delta$, excluding the additional randomization noise level $\epsilon$, as these approaches utilize PoIs as input data.
We made this choice considering that introducing excessive noise to PoIs might lead to non-physical or implausible PoI samples. Additionally, for the \pureOPO{} approach, our experimental results show that it exhibits superior performance without the data randomization technique. Thus, we also do not apply $\epsilon$ noise level for the \pureOPO{} approach.

{\bf Accuracy metric.}
To evaluate the performance of the learned inverse and PtO/forward mappings for all approaches, we compare the overall accuracy over 500 independent test samples. For the inverse mapping, we employ two error metrics: one based on the average pointwise absolute error values 
\begin{equation}
    \eqnlab{AbsEj}
    \text{E}_{abs,j} = \frac{1}{500} \sum_{i=1}^{500} \snor{\p_{j}^{i,\text{pred}} - \p_{j}^{\text{i,true}}} \quad j = 1, \ldots, \du,
\end{equation}
and another using the average relative Euclidean norm error
\begin{equation}
    \eqnlab{Err}
    \text{E}_{rel} = \frac{1}{500} \sum_{i=1}^{500} \frac{\nor{\ub^{i,\text{pred}} - \ub^{i,\text{true}}}_2^2}{\nor{\ub^{i,\text{true}}}_2^2},
\end{equation}
where $i$ indexes the sample number, $j$ represents the component index of the discretized sample vector, and $\du$ is the number of spatial grid points. The superscripts ``pred"  and ``true" stand for the solution predicted by the neural network and the ground truth parameters of interest, respectively.
For the learned PtO mapping, we assess accuracy using the average relative error between predicted observations $\yb^{\text{pred}}$ and true observations $\yb^{\text{true}}$ following \cref{eq:Err}. Since \mcOPOfull{} and \TNetAEfull{} approaches are developed to learn the forward map predicting the complete solution $\ybfull$, we maintain consistency in comparison by applying the same accuracy metric on $\yb = \B \ybfull$ across all approaches.

{\bf Training settings.}
For both heat equation and Navier-Stokes equation, we use a shallow neural network having one hidden layer with 5000 ReLU neurons for both encoder and decoder neural networks. In our previous work \cite{nguyen2024tnet}, we verified that a dense feed-forward neural network architecture with multiple layers, in small data regimes, performs poorly due to the vanishing gradient and/or the bias-variance trade-off issues. We thus focus on neural networks with a single hidden layer and this is sufficient to demonstrate the proposed AutoEncoder frameworks. Regarding the optimization algorithm, the default \texttt{ADAM} \cite{kingma2014adam} optimizer in \texttt{JAX} \cite{jax2018github} is used.
To ensure fairness across all numerical experiments, we initialize the neural network parameters consistently: weights follow a standard Gaussian distribution while biases are set to zero vectors, all using an identical random seed. This ensures the same initializers across all test cases.

{\bf Sequential training protocol.} Our training strategy follows a sequential two-phase approach. Initially, we optimize the encoder network to maximize accuracy on a test set of 500 samples (for either PoIs or observations). Subsequently, using this pre-trained encoder, we optimize the decoder network to achieve optimal accuracy on the same test dataset size. Our experimental investigations have demonstrated that simultaneous training of both encoder and decoder networks does not yield superior accuracy compared to this sequential approach. This can be attributed to two key factors: first, the decoder's performance is primarily contingent on the quality of the encoder's output; and second, the simultaneous optimization of both networks can compromise encoder accuracy due to the presence of the decoder loss term. Furthermore, the expanded parameter space in simultaneous training increases the likelihood of the optimizer converging to local minima. A theoretical discussion of the simultaneous training strategy compared to the sequential training strategy is presented in \cref{sect:simultaneous_training}. Therefore,
we shall focus on the results obtained by the sequential training strategy.

\subsection{2D heat equation}
\seclab{Heat_problem}

We investigate the following heat equation:
\begin{align*}
    -\nabla \cdot \LRp{e^u \nabla \omega}    & = 20  \quad \text{in } \Omega = \LRp{0,1}^2 \\
    \omega                                   & = 0 \quad \text{ on } \Gamma^{\text{ext}}   \\
    \textbf{n} \cdot \LRp{e^u \nabla \omega} & = 0 \quad \text{ on } \Gamma^{\text{root}},
\end{align*}
where $u$ represents the (log) conductivity coefficient field (the parameter of interest PoI), $\omega$ denotes the temperature field, and $\textbf{n}$ is the unit outward normal vector along the Neumann boundary $\Gamma^{\text{root}}$. As illustrated in the left panel of \cref{fig:2D_Heat_Model}, we discretize the domain using a $16 \times 16$ grid, with 10 randomly distributed observation points sampling from the discretized field $\ybfull$.
We aim to achieve two primary goals: (1) learning an inverse mapping to directly reconstruct the conductivity coefficient field $\ub$ from 10 discrete observations $\yb = \B \ybfull$, and (2) learning a PtO map or forward map that predicts either the temperature observations $\yb$ or the temperature field $\ybfull$ given a conductivity coefficient field $\ub$.

\begin{figure}[htb!]
    \centering
    \begin{tabular*}{\textwidth}{c@{\hskip -0.01cm} c@{\hskip -0.01cm} c@{\hskip -0.01cm} }
        \centering
        \raisebox{-0.5\height}{\resizebox{0.33\textwidth}{!}{\input{Figs/2D_Poisson/domain_mesh_observables_modified.tex}}} &
        \raisebox{-0.5\height}{\includegraphics[width = 0.34\textwidth]{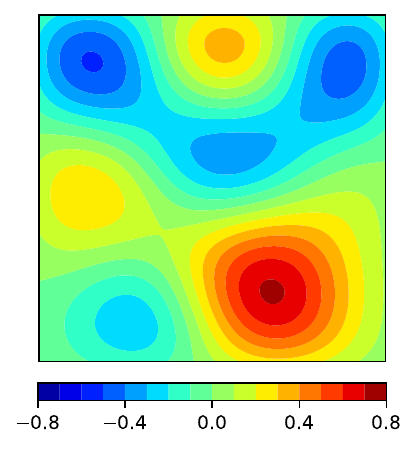}} &
        \raisebox{-0.5\height}{\includegraphics[width = 0.34\textwidth]{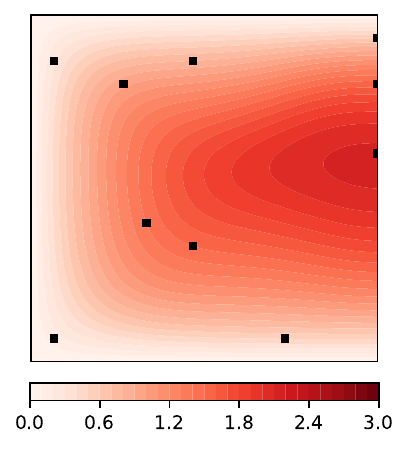}}
    \end{tabular*}
    \caption{{\bf 2D heat equation.} {\bf Left:} Domain, boundary conditions, $16 \times 16$ finite element discretization mesh, and $10$  random observation locations. {\bf Middle:} A sample of the PoI (the heat conductivity field). {\bf Right:} The corresponding state (temperature field), observations (temperatures) are taken at 10 observed points. This pair of PoI and observation sample is used for training in one training sample case.}
    \figlab{2D_Heat_Model}
\end{figure}

\begin{table}[htb!]
    \centering
    \caption{{\bf 2D heat equation.} The average relative error for inverse solutions and forward solutions (observations) over 500 test samples obtained by all approaches trained with $\LRc{1,100}$ training samples. 
    The model-constrained approaches are more accurate for both inverse (comparable to the Tikhonov\textemdash Tik\textemdash approach) and forward solution, and within the model-constrained approaches, \TNetAE{} and \TNetAEfull{} are the most accurate ones: in fact one training sample is sufficient for these two methods.}
    \begin{tabular}{c|rr|rr}
        \hline
                          & \multicolumn{2}{c|}{\textbf{1 training sample}} & \multicolumn{2}{c}{\textbf{100 training samples}}                                       \\
        \textbf{Approach} & \textbf{Inverse} (\%)                                & \textbf{Forward}                                  & \textbf{Inverse} (\%) & \textbf{Forward} \\
        \hline
        $\purePOP$        & 100.18                                          & 3.99$\times 10^{-1}$                                          & 80.48            & 5.30$\times 10^{-2}$         \\
        $\pureOPO$        & 107.55                                          & 2.90$\times 10^{-1}$                                          & 50.18            & 1.09$\times 10^{-1}$         \\
        $\mcPOP$          & 107.99                                          & 3.99$\times 10^{-1}$                                          & 87.60            & 5.30$\times 10^{-2}$         \\
        $\mcOPO$          & 108.28                                          & 2.73$\times 10^{-2}$                                          & 46.32            & 3.94$\times 10^{-4}$         \\
        $\mcOPOfull$      & 108.28                                          & 4.21$\times 10^{-2}$                                          & 46.32            & 4.56$\times 10^{-4}$         \\
        $\TNetAE$         & 45.23                                           & 1.57$\times 10^{-4}$                                          & 45.03            & 1.22$\times 10^{-4}$         \\
        $\TNetAEfull$     & 45.23                                           & 8.80$\times 10^{-4}$                                          & 45.03            & 2.12$\times 10^{-4}$         \\
        Tik               & 44.99                                           &                                                   & 44.99            &                  \\
        \hline
    \end{tabular}
    \tablab{2D_Heat_accuracy_table}
\end{table}

\begin{figure}[htbp]
    \centering
    \resizebox{.94\textwidth}{!}{
    \begin{tabular*}{\textwidth}{c@{\hskip -0.001cm} c@{\hskip -0.01cm} c@{\hskip -0.01cm} c@{\hskip -0.01cm} c@{\hskip -0.01cm}}
        \centering
        & \multicolumn{2}{c}{\textbf{1 training sample}}
        & \multicolumn{2}{c}{\textbf{100 training samples}}
        \\
        \rotatebox[origin=c]{90}{} &
        \raisebox{-0.5\height}{mean $ \quad $} &
        \raisebox{-0.5\height}{std $ \quad $} &
        \raisebox{-0.5\height}{mean $ \quad $}&
        \raisebox{-0.5\height}{std $ \quad $}
        \\
        \rotatebox[origin=c]{90}{$\purePOP$} &
        \raisebox{-0.5\height}{\includegraphics[width = 0.22\textwidth]{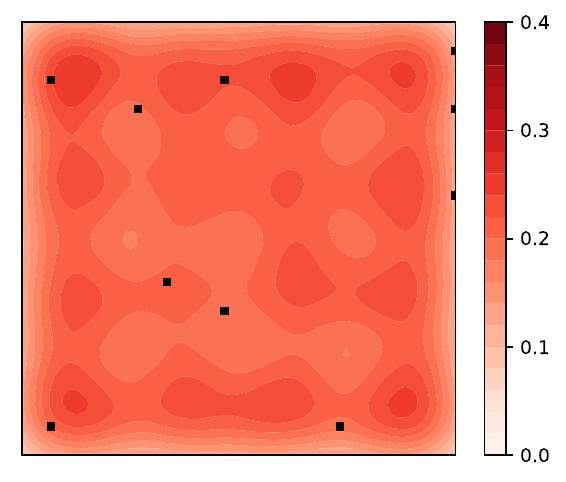}} &
        \raisebox{-0.5\height}{\includegraphics[width = 0.22\textwidth]{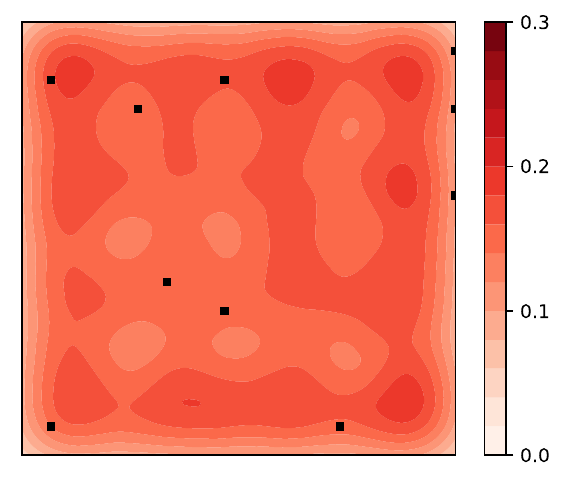}} &
        \raisebox{-0.5\height}{\includegraphics[width = 0.22\textwidth]{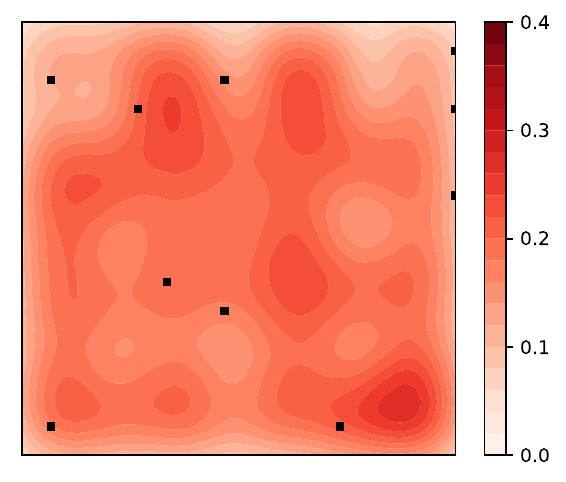}} &
        \raisebox{-0.5\height}{\includegraphics[width = 0.22\textwidth]{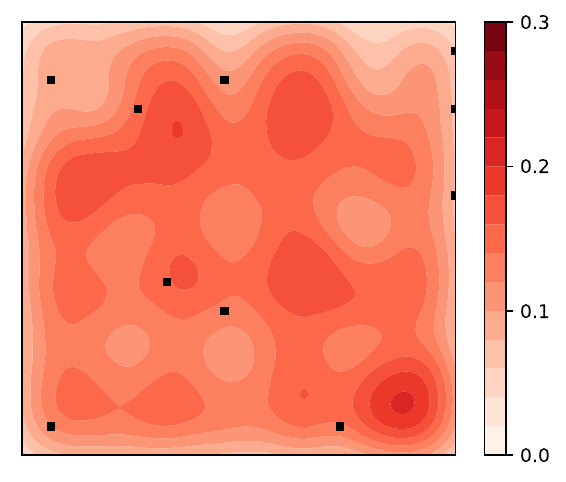}}
        \\
        \rotatebox[origin=c]{90}{$\pureOPO$} &
        \raisebox{-0.5\height}{\includegraphics[width = 0.22\textwidth]{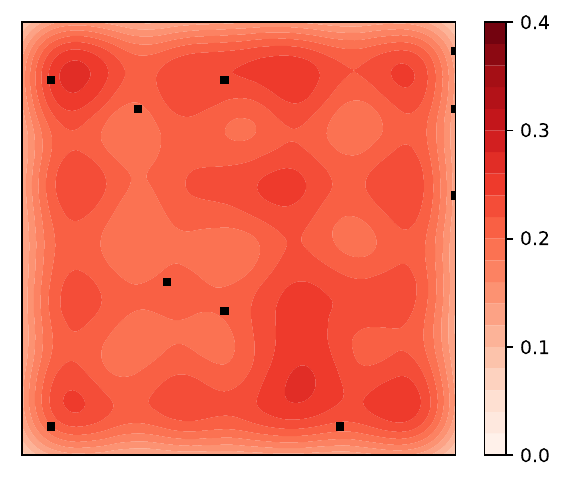}} &
        \raisebox{-0.5\height}{\includegraphics[width = 0.22\textwidth]{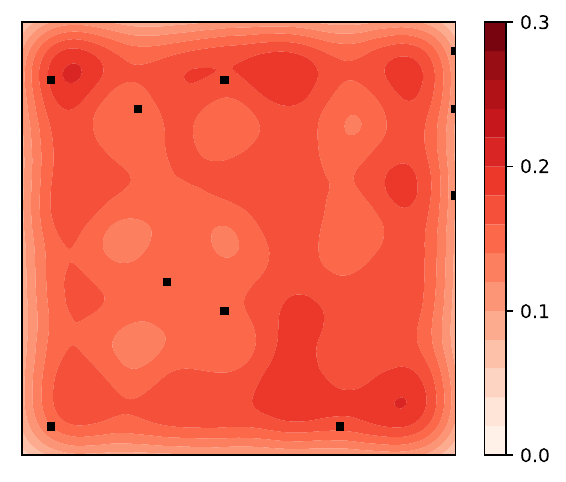}} &
        \raisebox{-0.5\height}{\includegraphics[width = 0.22\textwidth]{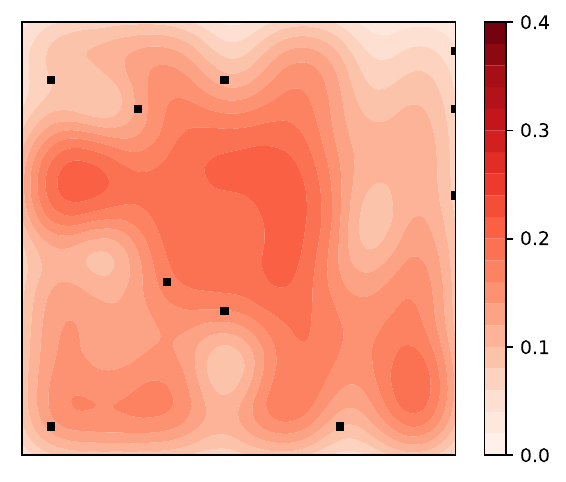}} &
        \raisebox{-0.5\height}{\includegraphics[width = 0.22\textwidth]{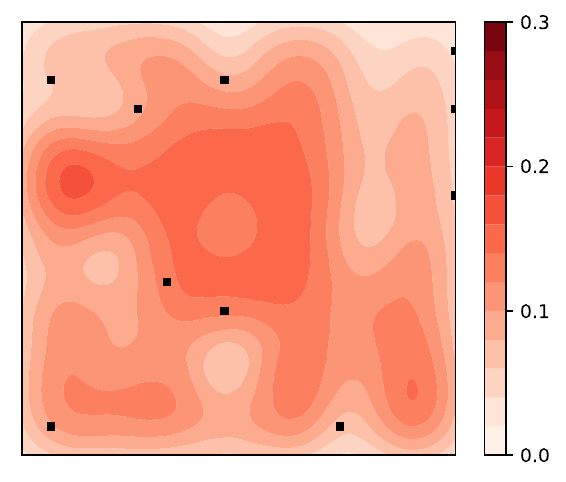}}
        \\
        \rotatebox[origin=c]{90}{$\mcPOP$} &
        \raisebox{-0.5\height}{\includegraphics[width = 0.22\textwidth]{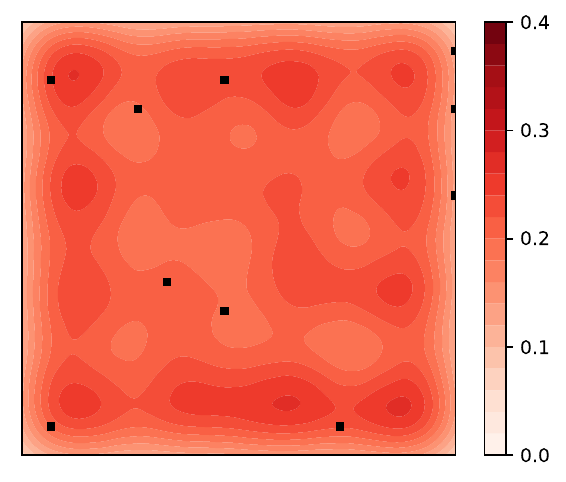}} &
        \raisebox{-0.5\height}{\includegraphics[width = 0.22\textwidth]{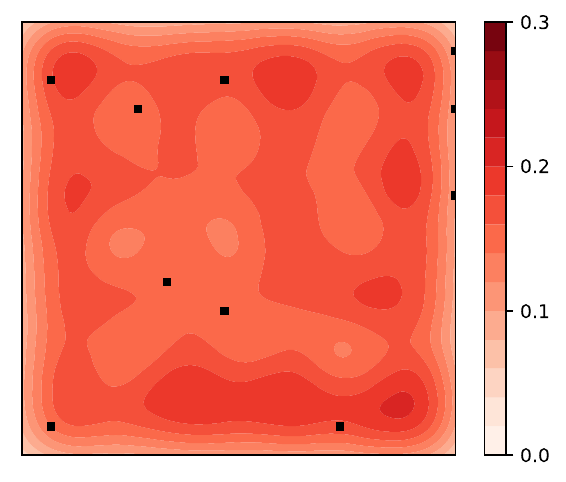}} &
        \raisebox{-0.5\height}{\includegraphics[width = 0.22\textwidth]{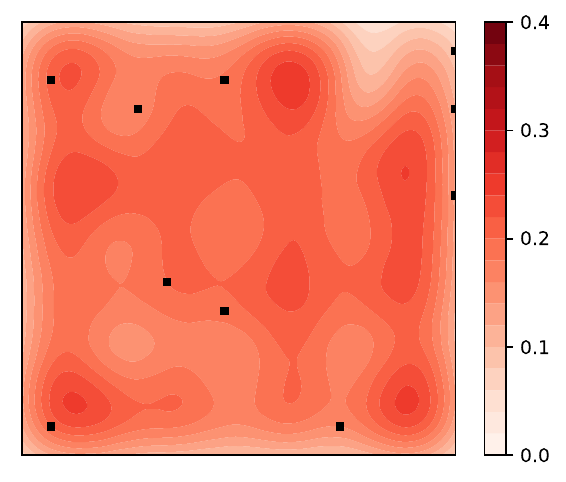}} &
        \raisebox{-0.5\height}{\includegraphics[width = 0.22\textwidth]{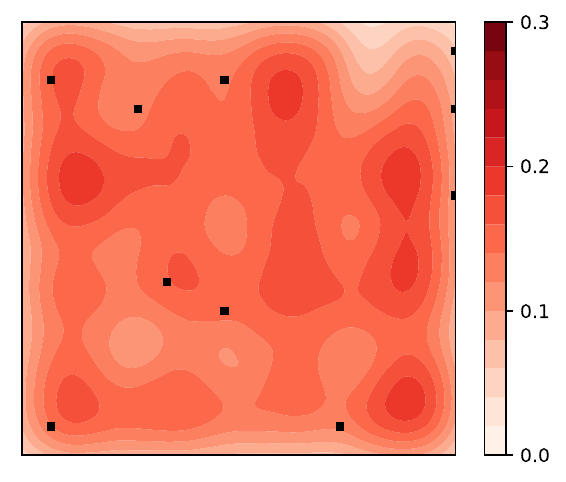}}
        \\
        \rotatebox[origin=c]{90}{$\begin{matrix} \mcOPO \\ \mcOPOfull \end{matrix}$} &
        \raisebox{-0.5\height}{\includegraphics[width = 0.22\textwidth]{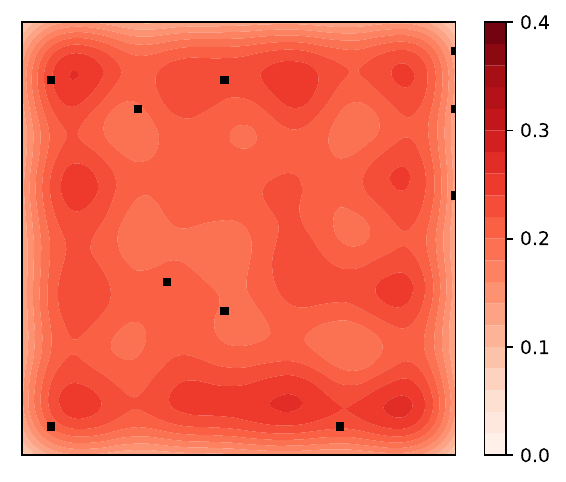}} &
        \raisebox{-0.5\height}{\includegraphics[width = 0.22\textwidth]{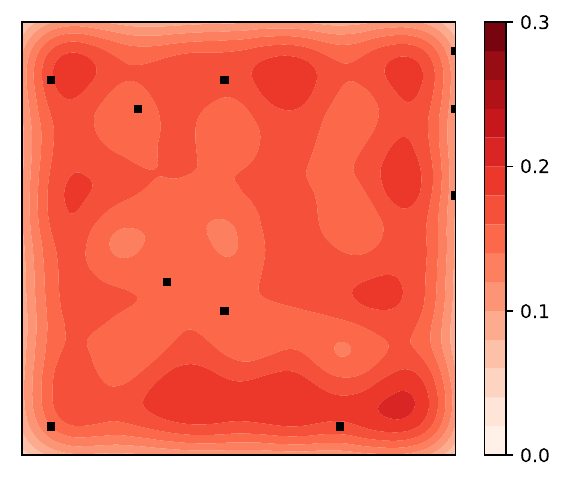}} &
        \raisebox{-0.5\height}{\includegraphics[width = 0.22\textwidth]{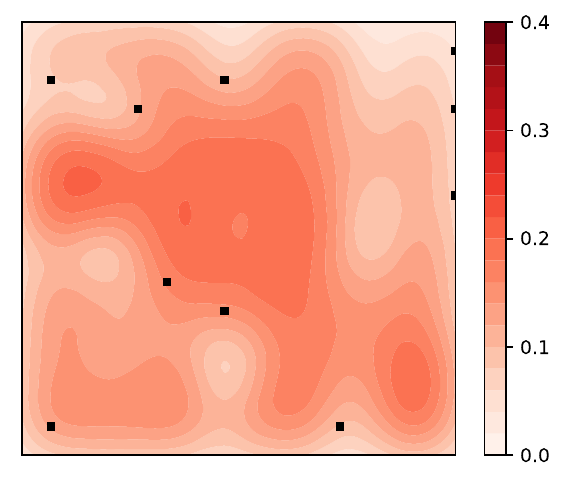}} &
        \raisebox{-0.5\height}{\includegraphics[width = 0.22\textwidth]{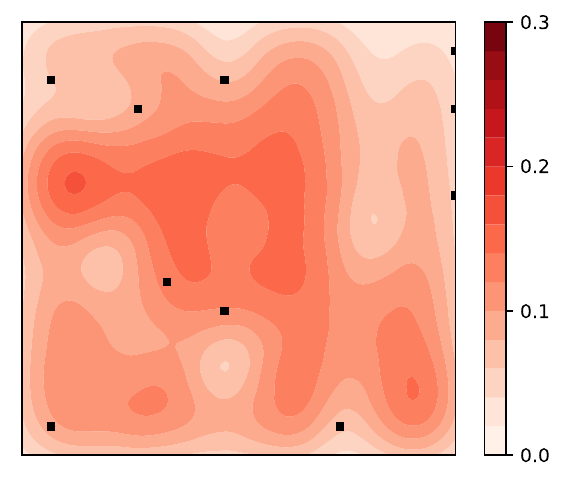}}
        \\
        \rotatebox[origin=c]{90}{$\begin{matrix} \TNetAE \\ \TNetAEfull \end{matrix}$} &
        \raisebox{-0.5\height}{\includegraphics[width = 0.22\textwidth]{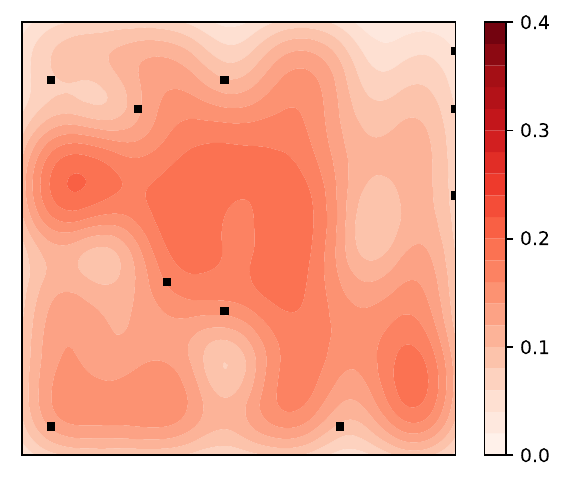}} &
        \raisebox{-0.5\height}{\includegraphics[width = 0.22\textwidth]{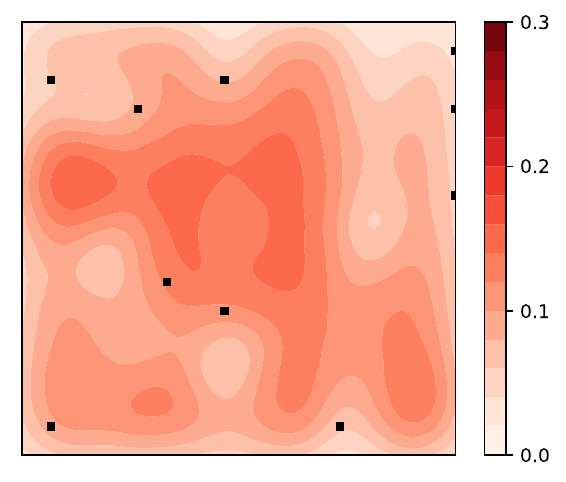}} &
        \raisebox{-0.5\height}{\includegraphics[width = 0.22\textwidth]{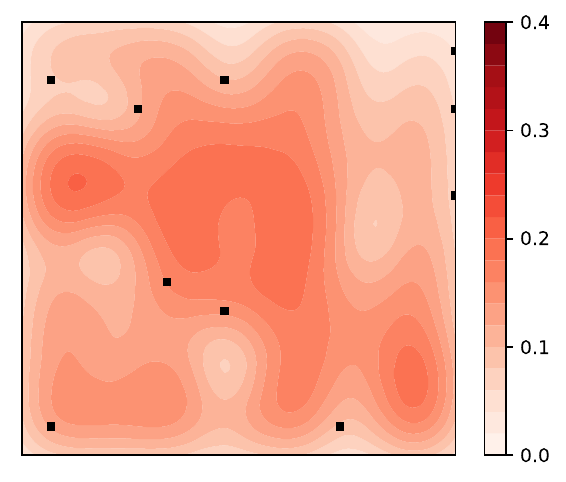}} &
        \raisebox{-0.5\height}{\includegraphics[width = 0.22\textwidth]{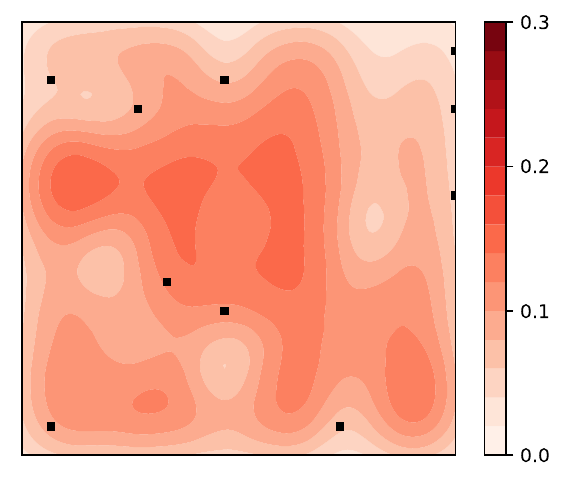}}
        \\
        \rotatebox[origin=c]{90}{Tik} &
        \raisebox{-0.5\height}{\includegraphics[width = 0.22\textwidth]{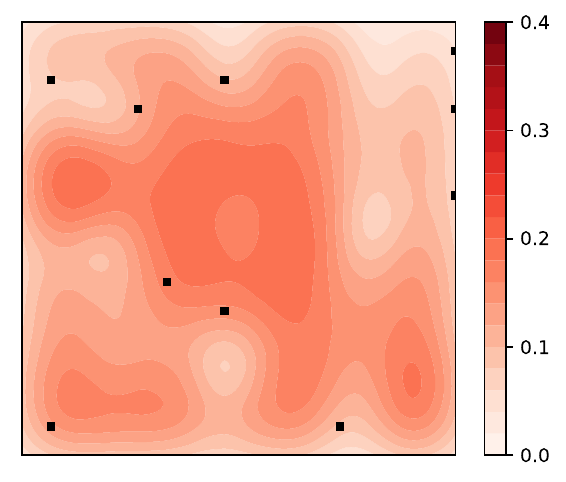}} &
        \raisebox{-0.5\height}{\includegraphics[width = 0.22\textwidth]{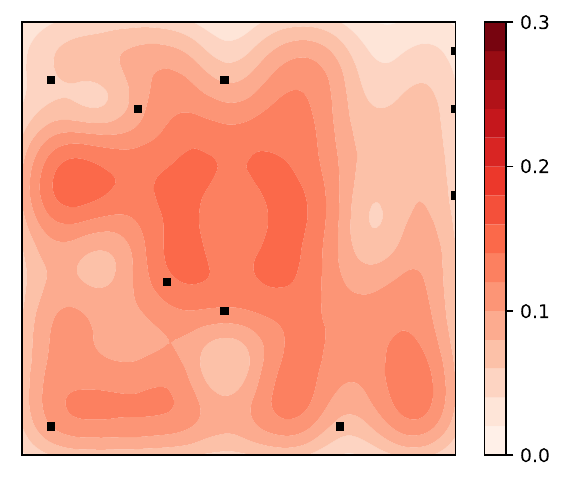}} &
        &
    \end{tabular*}
    }
    \caption{{\bf 2D heat equation.} Mean and standard deviation of absolute error for 500 test inverse solutions obtained from different approaches. Black points are observational locations. Note that \TNetAE{} and \TNetAEfull{} (and similarly for \purePOP{} and \mcPOP{} approaches) have the same encoder (that encodes the inverse solutions), their (identical) results are shown on the 5th row. Relatively to the Tikhonov approach (Tik), the model-constrained approaches are more accurate, and within the model-constrained approaches, \TNetAE{} and \TNetAEfull{} are the most accurate ones: in fact one training sample is sufficient for these two methods.}
    \label{fig:2D_Heat_accuracy_inverse}
\end{figure}

\begin{figure}[htb!]
    \centering
    \begin{tabular*}{\textwidth}{c@{\hskip -0.001cm}  c@{\hskip -0.01cm} c@{\hskip -0.01cm} c@{\hskip -0.01cm} c@{\hskip -0.01cm} c@{\hskip -0.01cm} c@{\hskip -0.01cm}}
        \centering
        \rotatebox[origin=c]{90}{} &
        \raisebox{-0.5\height}{$\begin{matrix} \purePOP{}/\mcPOP{} \end{matrix}$} &
        \raisebox{-0.5\height}{\pureOPO{}} &
        \raisebox{-0.5\height}{\mcOPO{}} &
        \raisebox{-0.5\height}{\mcOPOfull{}} &
        \raisebox{-0.5\height}{\TNetAE{}} &
        \raisebox{-0.5\height}{\TNetAEfull{}}
        \\
        \rotatebox[origin=c]{90}{1 sample} &
        \raisebox{-0.5\height}{\includegraphics[width = 0.16\textwidth]{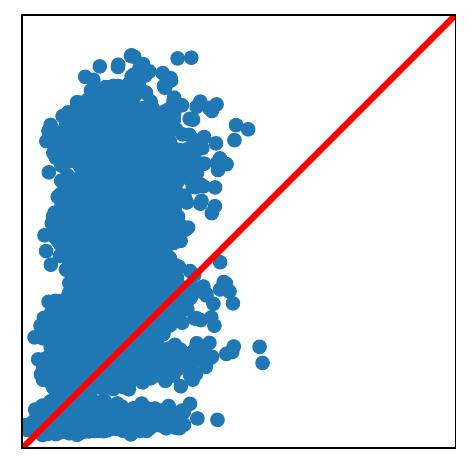}} &
        \raisebox{-0.5\height}{\includegraphics[width = 0.16\textwidth]{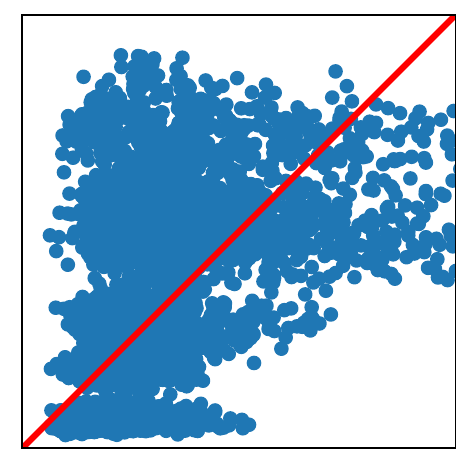}} &
        \raisebox{-0.5\height}{\includegraphics[width = 0.16\textwidth]{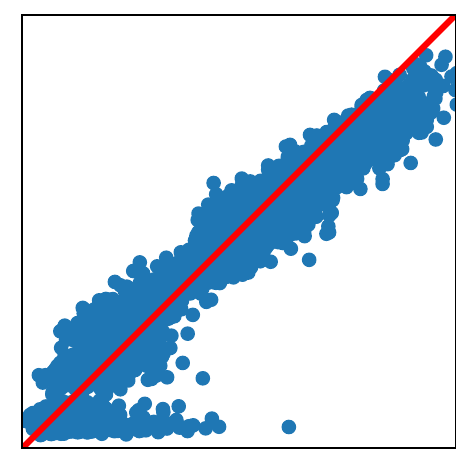}} &
        \raisebox{-0.5\height}{\includegraphics[width = 0.16\textwidth]{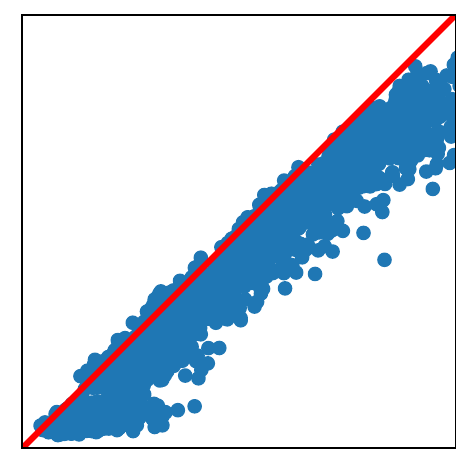}} &
        \raisebox{-0.5\height}{\includegraphics[width = 0.16\textwidth]{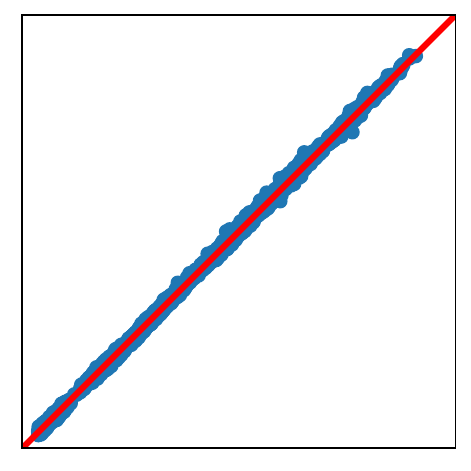}} &
        \raisebox{-0.5\height}{\includegraphics[width = 0.16\textwidth]{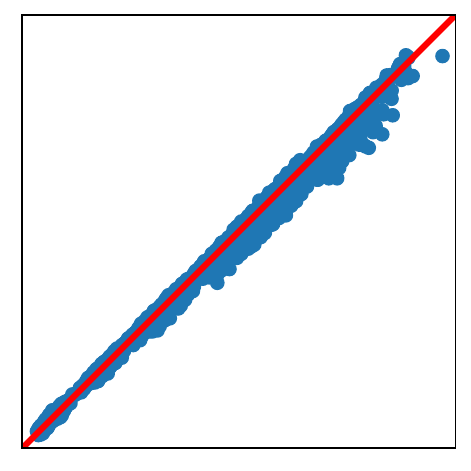}}
        \\
        \rotatebox[origin=c]{90}{100 samples} &
        \raisebox{-0.5\height}{\includegraphics[width = 0.16\textwidth]{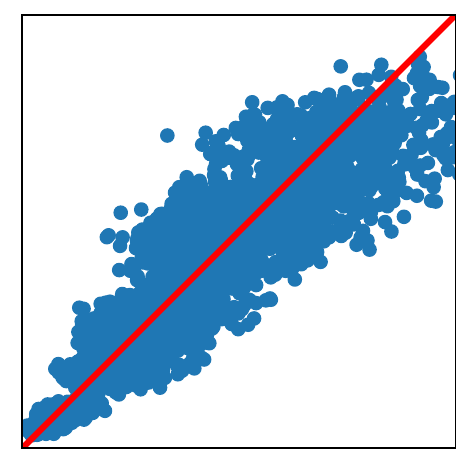}} &
        \raisebox{-0.5\height}{\includegraphics[width = 0.16\textwidth]{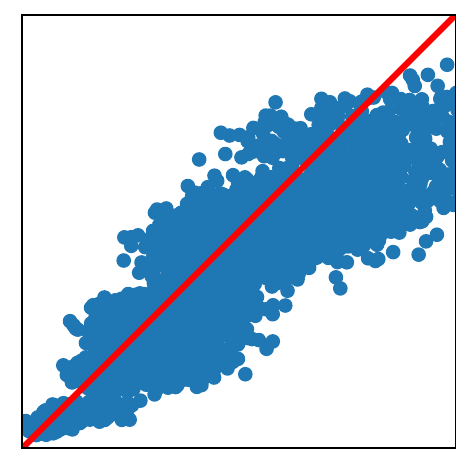}} &
        \raisebox{-0.5\height}{\includegraphics[width = 0.16\textwidth]{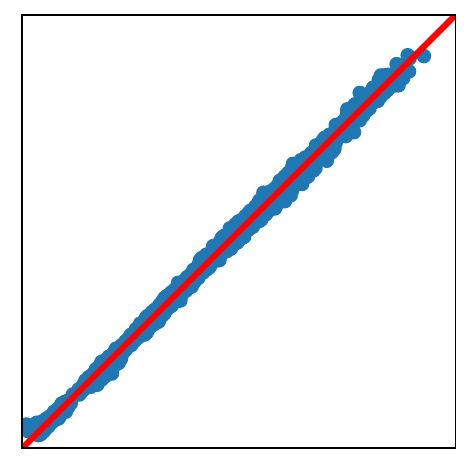}} &
        \raisebox{-0.5\height}{\includegraphics[width = 0.16\textwidth]{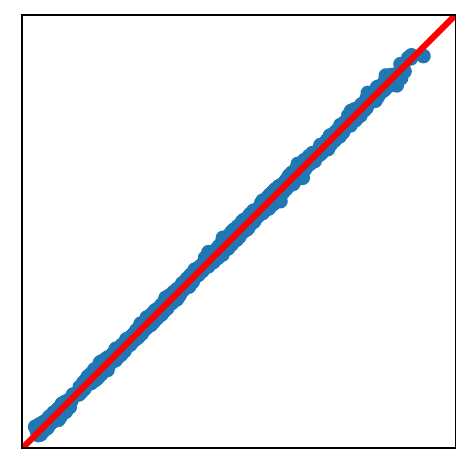}} &
        \raisebox{-0.5\height}{\includegraphics[width = 0.16\textwidth]{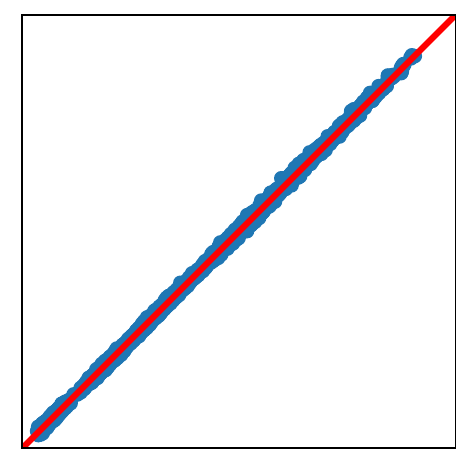}} &
        \raisebox{-0.5\height}{\includegraphics[width = 0.16\textwidth]{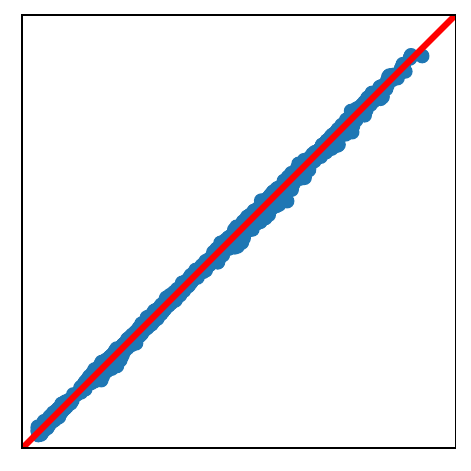}}
    \end{tabular*}
    \caption{{\bf 2D heat equation.} The comparison of 500 test predicted forward solution (at the observational locations) obtained from different approaches. In all plots plot, the x-axis is the magnitude of the true observation, and the y-axis is the magnitude of the predicted observation, both axises has range of $\LRs{0,3}$. The red line indicates the perfect matching between predictions and truth  observations. {\bf Top row:} Trained with $1$ training sample. {\bf Bottom row:} Trained with $100$ training samples.  As can be seen, model-constrained approaches are more accurate, and within the model-constrained approaches, \TNetAE{} and \TNetAEfull{} are the most accurate ones: in fact one training sample is sufficient for these two methods.}
    \label{fig:2D_Heat_accuracy_forward}
\end{figure}

\begin{figure}[htb!]
    \centering
    \resizebox{1.\textwidth}{!}{
    \begin{tabular*}{\textwidth}{c@{\hskip -0.01cm} c@{\hskip -0.01cm} c@{\hskip -0.01cm} c@{\hskip -0.01cm} c@{\hskip -0.01cm}}
        \centering
        & \multicolumn{2}{c}{\textbf{1 training sample}}
        & \multicolumn{2}{c}{\textbf{100 training samples}}
        \\
        & \raisebox{-0.5\height}{mean} &
        \raisebox{-0.5\height}{std} &
        \raisebox{-0.5\height}{mean}&
        \raisebox{-0.5\height}{std}
        \\
        \rotatebox[origin=c]{90}{$\mcOPOfull$} &
        \raisebox{-0.5\height}{\includegraphics[width = 0.24\textwidth]{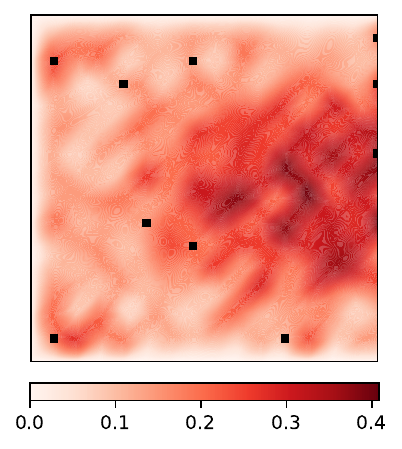}} &
        \raisebox{-0.5\height}{\includegraphics[width = 0.24\textwidth]{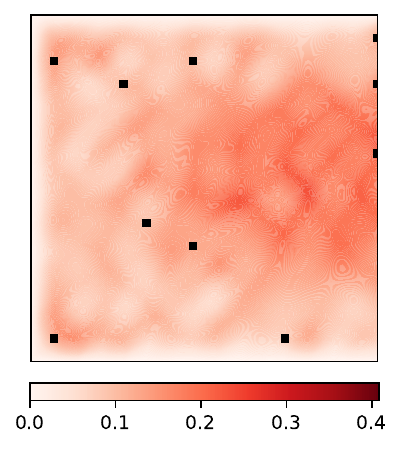}} &
        \raisebox{-0.5\height}{\includegraphics[width = 0.25\textwidth]{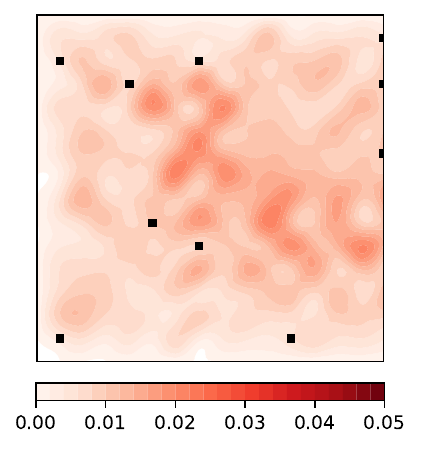}} &
        \raisebox{-0.5\height}{\includegraphics[width = 0.25\textwidth]{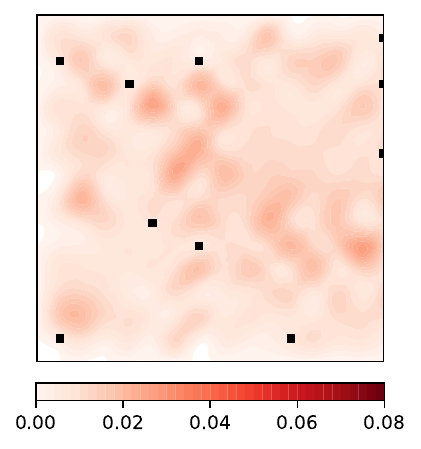}}
        \\
        \rotatebox[origin=c]{90}{$\TNetAEfull$} &
        \raisebox{-0.5\height}{\includegraphics[width = 0.25\textwidth]{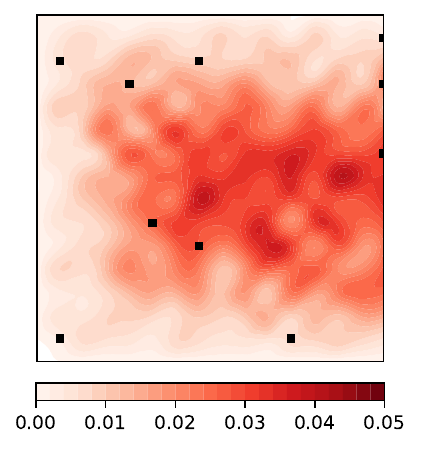}} &
        \raisebox{-0.5\height}{\includegraphics[width = 0.25\textwidth]{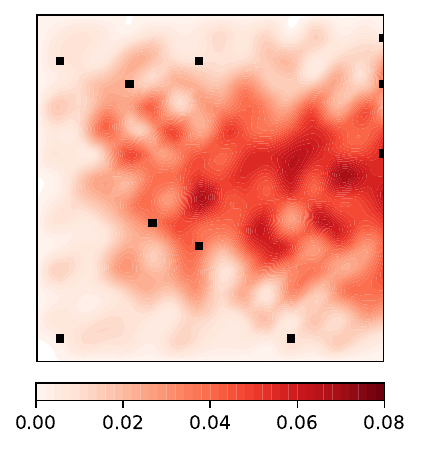}} &
        \raisebox{-0.5\height}{\includegraphics[width = 0.25\textwidth]{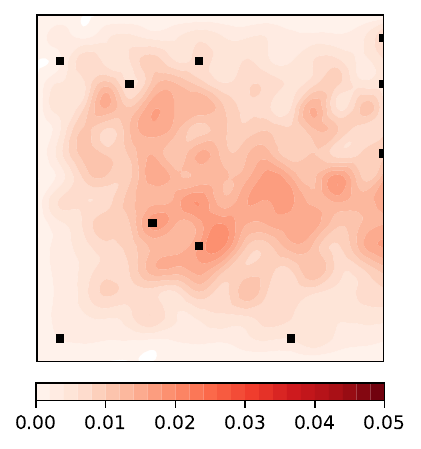}} &
        \raisebox{-0.5\height}{\includegraphics[width = 0.25\textwidth]{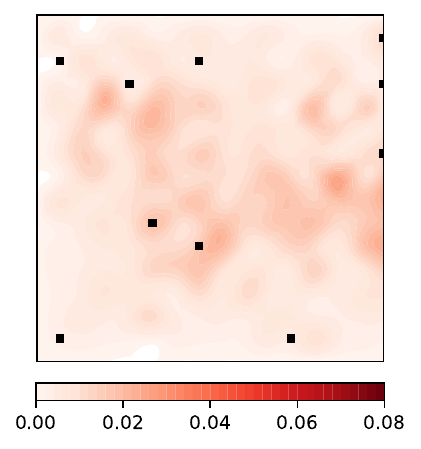}}
    \end{tabular*}
    }
    \caption{{\bf 2D heat equation.} Mean and standard deviation of absolute pointwise error for 500 full state test solutions obtained from $\TNetAEfull{}$ and \mcOPOfull{}. Black dots are the observational locations. The former is more accurate, especially for the case with one training sample in which it achieves two orders of magnitude smaller error.}
    \figlab{2D_Heat_accuracy_forward_full}
\end{figure}

\begin{figure}[htb!]
    \centering
    \begin{tabular*}{\textwidth}{c@{\hskip -0.01cm} c@{\hskip -0.01cm} c@{\hskip -0.01cm} c@{\hskip -0.01cm}}
        \centering
        \raisebox{-0.5\height}{$\ub_\text{Tik}$} &
        \raisebox{-0.5\height}{$\ub_{\TNetAEfull}$ } &
        \raisebox{-0.5\height}{$\ub_\text{True}$}&
        \raisebox{-0.5\height}{$\yb_{\TNetAEfull}$}
        \\
        \raisebox{-0.5\height}{\includegraphics[width = 0.25\textwidth]{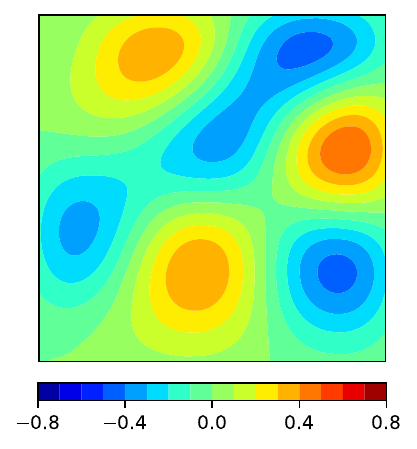}} &
        \raisebox{-0.5\height}{\includegraphics[width = 0.25\textwidth]{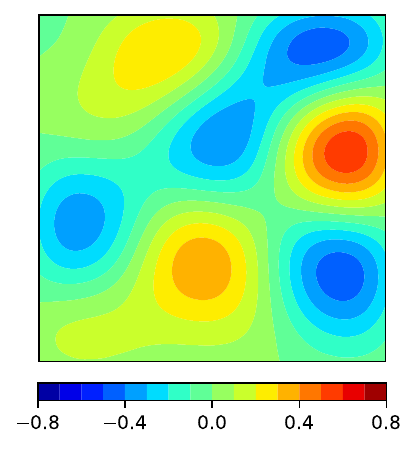}} &
        \raisebox{-0.5\height}{\includegraphics[width = 0.25\textwidth]{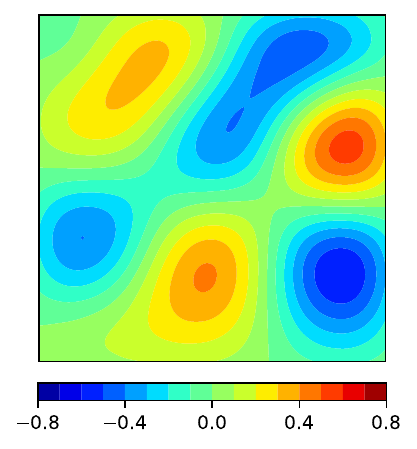}} &
        \raisebox{-0.5\height}{\includegraphics[width = 0.25\textwidth]{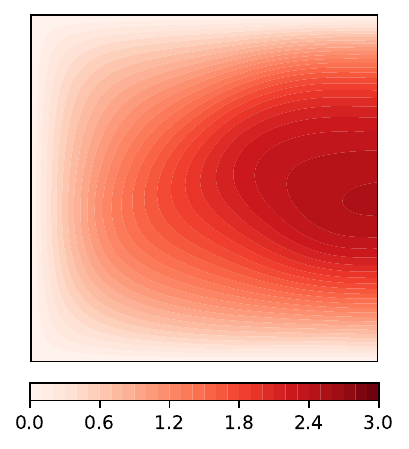}}
        \\
        \raisebox{-0.5\height}{$\snor{\ub_\text{Tik} - \ub_\text{True}}$} &
        \raisebox{-0.5\height}{$\snor{\ub_{\TNetAEfull} - \ub_\text{True}}$ } &
        \raisebox{-0.5\height}{$\yb_\text{True}$}&
        \raisebox{-0.5\height}{$\snor{\yb_{\TNetAEfull} - \yb_\text{True}}$}
        \\
        \raisebox{-0.5\height}{\includegraphics[width = 0.25\textwidth]{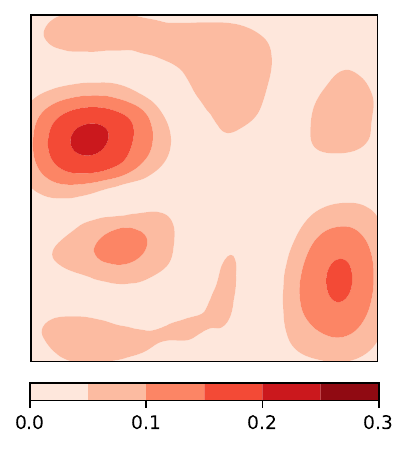}} &
        \raisebox{-0.5\height}{\includegraphics[width = 0.25\textwidth]{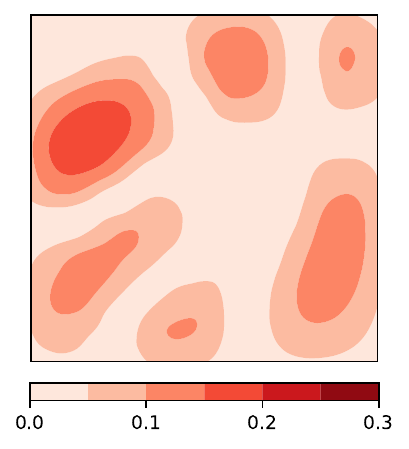}} &
        \raisebox{-0.5\height}{\includegraphics[width = 0.25\textwidth]{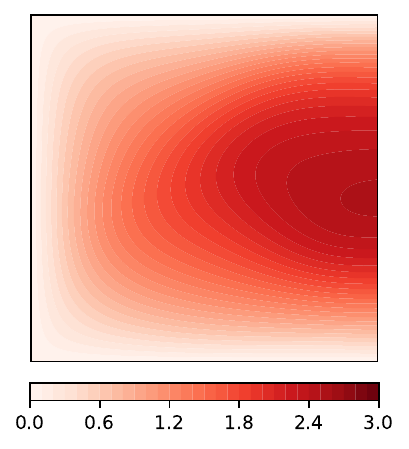}} &
        \raisebox{-0.5\height}{\includegraphics[width = 0.25\textwidth]{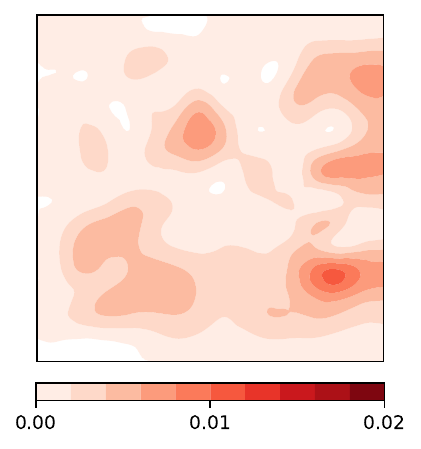}}
    \end{tabular*}
    \caption{{\bf 2D heat equation.} A (random) representative case of inverse and full forward solution obtained by $\TNetAEfull$ trained with 1 training sample coupled with data randomization of noise level $\sigma = 0.1$. $\TNetAEfull$ inverse solution is comparable to the Tikhonov (Tik) inverse counterpart, and both are consistent with the ground truth (True). $\TNetAEfull$ full forward solution is almost identical (in fact within 3 digits of accuracy) to the underlying true solution.}
    \figlab{2D_Heat_accuracy_TNetAEfull_framework}
\end{figure}

\begin{figure}[htb!]
    \centering
\begin{tikzpicture}[
    ]
    \definecolor{green01270}{RGB}{0,127,0}
    \definecolor{lightgray204}{RGB}{204,204,204}
    \definecolor{darkgray176}{RGB}{176,176,176}

    \begin{axis}[
            width=12cm,
            height=5cm,
            tick align=outside,
            tick pos=left,
            x grid style={darkgray176},
            xlabel={Noise level $\epsilon$},
            xmin=-0.02, xmax=0.42,
            xtick style={color=black},
            y grid style={darkgray176},
            ylabel={Relative error (\%)},
            ymin=0.4, ymax=0.8,
            ytick style={color=black},
            xtick={0,0.1,0.2,0.3,0.4},
            xticklabels={0,0.1,0.2,0.3,0.4},
            ytick={0.4,0.5,0.6,0.7,0.8},
            yticklabels={40, 50, 60, 70, 80},
            legend cell align={left},
            legend style={
                    fill opacity=0.8,
                    draw opacity=1,
                    text opacity=1,
                    at={(0.1,0.8)},
                    anchor=west,
                    draw=lightgray204
                },
        ]
        \addplot [thick, blue]
        table {%
                0 0.766974268854683
                0.02 0.523683383837577
                0.05 0.460874094482016
                0.1 0.448766294680066
                0.15 0.449361065883312
                0.2 0.453057512322396
                0.25 0.466427642188141
                0.3 0.506496804889437
                0.35 0.581920155123714
                0.4 0.7427047409309
            };
        \addlegendentry{$\TNetAEfull$ with one training sample}
        \addplot [thick, black]
        table {%
                0 0.449
                0.02 0.449
                0.05 0.449
                0.1 0.449
                0.15 0.449
                0.2 0.449
                0.25 0.449
                0.3 0.449
                0.35 0.449
                0.4 0.449
            };
        \addlegendentry{Tik}
        \addplot [draw=red, fill=red, mark=asterisk, only marks]
        table{%
                x  y
                0 0.766974268854683
                0.02 0.523683383837577
                0.05 0.460874094482016
                0.1 0.448766294680066
                0.15 0.449361065883312
                0.2 0.453057512322396
                0.25 0.466427642188141
                0.3 0.506496804889437
                0.35 0.581920155123714
                0.4 0.7427047409309
            };
    \end{axis}
\end{tikzpicture}
    \caption{{\bf 2D heat equation.} Relative error of inverse solution over 500 test samples with different noise levels.}
    \figlab{Heat_noise_level}
\end{figure}
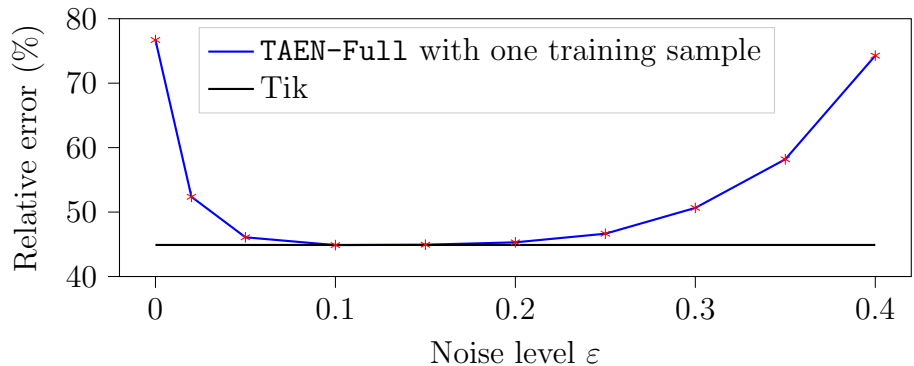

\begin{figure}[htb!]
    \centering
    \begin{tabular*}{\textwidth}{c@{\hskip -0.01cm} c@{\hskip -0.01cm}  c@{\hskip -0.01cm}}
        \centering
        & & \\
        \raisebox{-0.5\height}{\resizebox{0.33\textwidth}{0.33\textwidth}{
\begin{tikzpicture}

\begin{axis}[
unit vector ratio*=1 1 1,
width=5cm,
height=5cm,
tick pos=left,
xmin=-0.01, xmax=1.01,
ymin=-0.01, ymax=1.01,
xtick=\empty, 
ytick=\empty  
]
\addplot [thick, blue]
table {%
0 0
0 1
};
\addplot [thick, blue]
table {%
0 0
1 0
};
\addplot [thick, blue]
table {%
0 1
1 1
};
\addplot [thick, red]
table {%
1 0
1 1
};
\addplot [semithick, black, mark=square*, mark size=1.5, mark options={solid}, only marks]
table {%
0.0666666666666667 0.866666666666667
0.266666666666667 0.8
0.466666666666667 0.866666666666667
0.333333333333333 0.4
0.0666666666666667 0.0666666666666667
1 0.933333333333333
0.466666666666667 0.333333333333333
1 0.8
1 0.6
0.733333333333333 0.0666666666666667
};
\draw (axis cs:0.0866666666666667,0.856666666666667) node[
  scale=0.6,
  anchor=base west,
  text=black,
  rotate=0.0
]{1};
\draw (axis cs:0.286666666666667,0.79) node[
  scale=0.6,
  anchor=base west,
  text=black,
  rotate=0.0
]{2};
\draw (axis cs:0.486666666666667,0.856666666666667) node[
  scale=0.6,
  anchor=base west,
  text=black,
  rotate=0.0
]{3};
\draw (axis cs:0.353333333333333,0.39) node[
  scale=0.6,
  anchor=base west,
  text=black,
  rotate=0.0
]{4};
\draw (axis cs:0.0866666666666667,0.0566666666666667) node[
  scale=0.6,
  anchor=base west,
  text=black,
  rotate=0.0
]{5};
\draw (axis cs:0.89,0.923333333333333) node[
  scale=0.6,
  anchor=base west,
  text=black,
  rotate=0.0
]{6};
\draw (axis cs:0.486666666666667,0.323333333333333) node[
  scale=0.6,
  anchor=base west,
  text=black,
  rotate=0.0
]{7};
\draw (axis cs:0.89,0.79) node[
  scale=0.6,
  anchor=base west,
  text=black,
  rotate=0.0
]{8};
\draw (axis cs:0.89,0.59) node[
  scale=0.6,
  anchor=base west,
  text=black,
  rotate=0.0
]{9};
\draw (axis cs:0.753333333333333,0.0566666666666667) node[
  scale=0.6,
  anchor=base west,
  text=black,
  rotate=0.0
]{10};
\end{axis}

\end{tikzpicture}}}
        & \multicolumn{2}{c}{\raisebox{-0.5\height}{\resizebox{0.66\textwidth}{!}{
\begin{tikzpicture}

\definecolor{darkgray176}{RGB}{176,176,176}
\definecolor{green01270}{RGB}{0,127,0}
\definecolor{lightgray204}{RGB}{204,204,204}

\begin{axis}[
unit vector ratio*=1 2 1,
width=15cm,
height=15cm,
legend cell align={left},
legend style={
  fill opacity=0.8,
  draw opacity=1,
  text opacity=1,
  at={(0.03,0.97)},
  anchor=north west,
  draw=lightgray204
},
tick align=outside,
tick pos=left,
x grid style={darkgray176},
xlabel={Observation Index},
xmin=0.55, xmax=10.45,
xtick style={color=black},
y grid style={darkgray176},
ylabel={Observation Magnitude},
ymin=0, ymax=2.5,
ytick style={color=black},
ytick = {0, 0.5, 1, 1.5, 2, 2.5},
]
\addplot [draw=red, fill=red, forget plot, mark=o, only marks]
table{%
x  y
1 0.265873566050922
2 0.97471234899911
3 0.807413147829283
4 1.40992411704074
5 0.147688492910551
6 0.594184481432237
7 1.48878647473187
8 1.56088548744196
9 2.12965687513923
10 0.503570883597474
};
\addplot [draw=red, fill=red, forget plot, mark=o, only marks]
table{%
x  y
1 0.211510186135435
2 0.810047063763382
3 0.943274204240663
4 1.39014993097203
5 0.154965715059198
6 0.548822917628627
7 1.55856088362134
8 1.43673985706092
9 2.30259265984566
10 0.590496246160572
};
\addplot [draw=red, fill=red, forget plot, mark=o, only marks]
table{%
x  y
1 0.238916859741404
2 0.813516599110498
3 0.665259888407642
4 1.30561030767723
5 0.125761488655086
6 0.533639166743974
7 1.5933802603813
8 1.37648553844452
9 2.06391723123073
10 0.571620801726384
};
\addplot [draw=red, fill=red, forget plot, mark=o, only marks]
table{%
x  y
1 0.26997284521999
2 0.970270632262377
3 0.836690810917981
4 1.35756382352214
5 0.104792212168439
6 0.457478130276483
7 1.45303126398551
8 1.18681702623109
9 1.95696737601785
10 0.486429704540052
};
\addplot [draw=red, fill=red, forget plot, mark=o, only marks]
table{%
x  y
1 0.192997100336719
2 0.816623329211214
3 0.972114716473459
4 1.63700094201829
5 0.168595695976298
6 0.482310303597305
7 1.82196982667767
8 1.33052026775705
9 2.32469506848055
10 0.593371508849173
};
\addplot [draw=red, fill=red, forget plot, mark=o, only marks]
table{%
x  y
1 0.238090144814072
2 0.925814607257964
3 0.875885418069347
4 1.62302980140666
5 0.14362048793716
6 0.443754641448407
7 1.68796660031215
8 1.15133745599664
9 1.99847458772659
10 0.541131506770507
};
\addplot [draw=red, fill=red, forget plot, mark=o, only marks]
table{%
x  y
1 0.190827835793034
2 0.700410976081476
3 0.755382007951036
4 1.46423379242097
5 0.138729731078721
6 0.466812500542599
7 1.65240751842401
8 1.18094993670679
9 1.97688560529447
10 0.573982224747467
};
\addplot [draw=red, fill=red, forget plot, mark=o, only marks]
table{%
x  y
1 0.309863151104472
2 1.03487856304763
3 0.905906205465422
4 1.72260615961142
5 0.145678926811326
6 0.673521533389162
7 1.85372460203904
8 1.7098865888851
9 2.32952808919299
10 0.57859913005946
};
\addplot [draw=red, fill=red, forget plot, mark=o, only marks]
table{%
x  y
1 0.247578011215592
2 1.09254749668905
3 1.09223163493808
4 1.88391311665625
5 0.157711689629619
6 0.615756784159446
7 2.00344547375776
8 1.57605266807168
9 2.25071420068184
10 0.557734138306532
};
\addplot [draw=red, fill=red, forget plot, mark=o, only marks]
table{%
x  y
1 0.203601891715918
2 0.750142885185601
3 0.690494074762876
4 1.32776418535731
5 0.136359104385484
6 0.556454298295732
7 1.58539973010561
8 1.40933836055985
9 2.04124182567043
10 0.597252065070879
};
\addplot [draw=red, fill=red, mark=o, only marks]
table{%
x  y
1 0.203601891715918
2 0.750142885185601
3 0.690494074762876
4 1.32776418535731
5 0.136359104385484
6 0.556454298295732
7 1.58539973010561
8 1.40933836055985
9 2.04124182567043
10 0.597252065070879
};
\addlegendentry{Samples}
\path [draw=green01270, ultra thick]
(axis cs:1,0.2154041487395)
--(axis cs:1,0.277237579249784);

\path [draw=green01270, ultra thick]
(axis cs:2,0.811329796906926)
--(axis cs:2,1.02834367627276);

\path [draw=green01270, ultra thick]
(axis cs:3,0.792855765419241)
--(axis cs:3,1.00413828716208);

\path [draw=green01270, ultra thick]
(axis cs:4,1.38208322368079)
--(axis cs:4,1.64414328338489);

\path [draw=green01270, ultra thick]
(axis cs:5,0.126709216389365)
--(axis cs:5,0.164631425959144);

\path [draw=green01270, ultra thick]
(axis cs:6,0.503097911109278)
--(axis cs:6,0.648171650882951);

\path [draw=green01270, ultra thick]
(axis cs:7,1.53162379925694)
--(axis cs:7,1.81218664236401);

\path [draw=green01270, ultra thick]
(axis cs:8,1.27807199464406)
--(axis cs:8,1.6537617418004);

\path [draw=green01270, ultra thick]
(axis cs:9,1.98794673846897)
--(axis cs:9,2.36653130006972);

\path [draw=green01270, ultra thick]
(axis cs:10,0.492433165289959)
--(axis cs:10,0.615510931577083);

\addplot [ultra thick, green01270, mark=-, mark size=5, mark options={solid}, only marks]
table {%
1 0.2154041487395
2 0.811329796906926
3 0.792855765419241
4 1.38208322368079
5 0.126709216389365
6 0.503097911109278
7 1.53162379925694
8 1.27807199464406
9 1.98794673846897
10 0.492433165289959
};
\addlegendentry{Standard deviation}
\addplot [semithick, blue, mark=x, mark size=5, mark options={solid}, only marks]
table {%
1 0.246320863994642
2 0.919836736589845
3 0.89849702629066
4 1.51311325353284
5 0.145670321174254
6 0.575634780996114
7 1.67190522081047
8 1.46591686822223
9 2.17723901926934
10 0.553972048433521
};
\addlegendentry{Mean}
\addplot [ultra thick, green01270, mark=-, mark size=5, mark options={solid}, only marks]
table {%
1 0.277237579249784
2 1.02834367627276
3 1.00413828716208
4 1.64414328338489
5 0.164631425959144
6 0.648171650882951
7 1.81218664236401
8 1.6537617418004
9 2.36653130006972
10 0.615510931577083
};
\end{axis}

\end{tikzpicture}}}}
    \end{tabular*}
    \caption{{\bf 2D heat equation.} {\bf Left}: Index of $10$ observational locations. {\bf Right}: Mean and standard deviation of observation magnitudes of 10000 true observation samples at the observational locations. The magnitudes of the predicted solutions of 10 different observation samples for single-sample training cases.}
    \figlab{2D_Heat_random_traning_samples}
\end{figure}
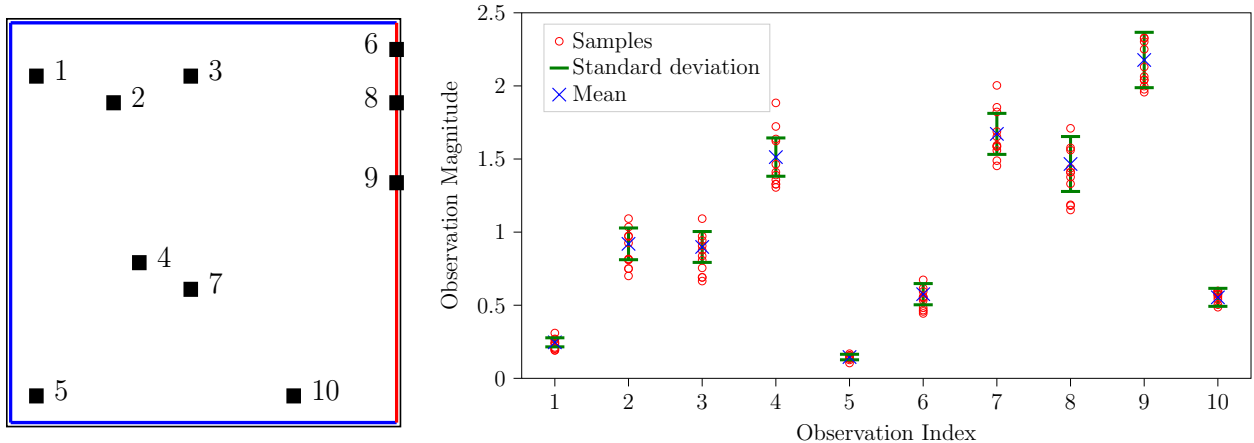

{\bf Generating train and test data sets.}
We start with drawing the parameter conductivity samples via a truncated Karhunen-Lo\'eve expansion
\begin{equation*}
    \eqnlab{heat_KL}
    u(x) = \sum_{i =1 }^q \sqrt{\lambda_i} \mb{\phi}_i(x) \ub_i, \quad x \in \LRs{0,1}^2,
\end{equation*}
where $\LRp{\lambda_i, \mb{\phi}_i}$ is the eigenpair of a two-point correlation function from \cite{constantine2016accelerating}, and $\ub = \LRc{\ub_i}_{i=1}^q \sim \mc{N}\LRp{0,\I}$ is a standard Gaussian random vector. We choose $q = 15$ eigenvectors corresponding to the first $15$ largest eigenvalues. For each sample $\ub$
we solve the heat equation for the corresponding temperature field $\ybfull$ by finite element method.
The observation samples $\yb^\text{clean}$ are constructed by extracting values of the temperature field $\ybfull$ at the  10 observable locations (see \cref{fig:2D_Heat_Model}), followed by the addition of Gaussian noise with the noise level of $\delta = 0.5\%$. Our training dataset consists of 100 independently drawn sample pairs. The middle and right figures in \cref{fig:2D_Heat_Model} show the conductivity coefficient fields $u$ and its corresponding temperature field $\ybfull$ for the first pair out of 100 training sample pairs. This particular pair serves as the training data for the single-sample training case for all approaches. For the inference (testing) step, we generate 500 independently drawn pairs $\LRp{\ub, \ybfull}$ following the same procedure discussed above.

{\bf Learned inverse and PtO/forward maps accuracy.}
We compare all approaches using the aforementioned two-phase sequential training protocol.
All methods are implemented under two scenarios: training with a single training sample and training with 100 training samples. For  \mcOPO{}, \mcOPOfull{}, \TNetAE{} and \TNetAEfull{} approaches, we perform data randomization for each epoch by adding random noise with magnitude $\epsilon = 10\%$ to the already-noise-corrupted observation samples as specified in \cref{eq:noise_training}. \cref{tab:2D_Heat_accuracy_table}  (with further details in \cref{fig:2D_Heat_accuracy_inverse} and \cref{fig:2D_Heat_accuracy_forward}, respectively) summarizes the average relative errors (with further details on average mean, standard deviation, and correlation between the true and predicted forward solutions, respectively) for 500 test samples calculated using \cref{eq:Err}, for both inverse and predicted observable solutions.
As can be seen, he model-constrained approaches are more accurate, and within the model-constrained approaches, 
 \TNetAE{} and \TNetAEfull{} are the most accurate ones: in fact one training sample is sufficient for these two methods.

Let us provide provide additional discussions on inverse learning (see \cref{tab:2D_Heat_accuracy_table} and \cref{fig:2D_Heat_accuracy_inverse}). \TNetAE{} and \TNetAEfull{} attain inverse solution accuracy of 45.23\%, comparable to the traditional Tikhonov regularization method with 44.99\% error. In contrast, all other methodologies (\purePOP{}, \pureOPO{}, \mcPOP{}, \mcOPO{}, and \mcOPOfull{}) fail to produce meaningful results. This performance disparity is due to the superior generalization capability of the \TNetAE{} and \TNetAEfull{} approaches, while other methods suffer from overfitting the provided training sample. 
As presented in \cref{sect:TNet_autoencoder}, \TNetAE{} and \TNetAEfull{} integrated with data randomization technique allows learning the inverse map as good as the Tikhonov regularization framework for linear inverse problems with only one arbitrary observation training sample. 
The data randomization technique serves two crucial functions: Exploring the unseen test observation sample space and exploiting the underlying physics via the model-constrained term. 

Next, we provide further details on learning PtO/forward maps (see \cref{tab:2D_Heat_accuracy_table} and \cref{fig:2D_Heat_accuracy_forward}). \TNetAE{} and \TNetAEfull{} can achieve highly accurate predictions of temperature solutions through the PtO/forward map (decoder), achieving relative errors of 1.57e-04 and 8.80e-04, respectively. This demonstrates their capability to learn accurately PtO/forward mappings from a single observation sample, again thanks to the data randomization. Since \TNetAE{} learns the observations directly, it is more accurate than \TNetAEfull{} which aims to learn the full solution state.
While our analysis in \cref{sect:mc_OPO} indicates that \mcOPO{} and \mcOPOfull{} can theoretically learn exact PtO/forward maps in linear problems, their actual performance (errors of 2.73e-02 and 4.21e-2, respectively) falls short compared to \TNetAE{} and \TNetAEfull{}. This underperformance can be attributed to two key factors: inaccurate inverse solutions obtained from the encoder and the nonlinear forward map.
On the other hand,  \purePOP{}, \pureOPO{}, and \mcPOP{} yield inaccurate PtO mappings, which is consistent with their purely data-driven architecture that does not encode physical constraints. The spatial distribution of prediction errors is further examined in \cref{fig:2D_Heat_accuracy_forward_full}, which depicts the mean and standard deviation of absolute pointwise errors for 500 test temperature field samples (unseen full state solutions). It can be seen, \TNetAEfull{} framework demonstrates consistently lower error statistics (two orders of magnitude smaller for one-sample case) compared to \mcOPOfull{}.
These results provide additional validation of the \TNetAEfull{} framework's efficacy in learning forward mappings (in tandem with learning the inverse solutions).

We have also seen that, for the larger data set of 100 samples, the accuracy of forward and inverse maps for all approaches is improved as expected. \TNetAE{} and \TNetAEfull{} maintain their superior performance for both PtO/forward and inverse solutions. We emphasize that the best inverse map obtained from \TNetAE{} and \TNetAEfull is just as good as the Tikhonov regularization method; thus, not much improvement is observed compared to the single-sample training case. In other words,  {\bf one training data is sufficient for \TNetAE{} and \TNetAEfull.} Coming in second are
\mcOPO{} and \mcOPOfull{}. Unlike the single-sample training scenario, these approaches now achieve reliable inverse maps (encoders), enabling their model-constrained decoders to achieve high accuracy in PtO/forward solutions.
Meanwhile, the purely data-driven approach \pureOPO{}, without additional information provided by the physics constraints, produces lower accuracy in inverse solutions compared to \mcOPO{}. This is not surprising, as analyzed in \cref{sect:naive_OPO}, the inaccurate inverse solutions from its encoder network inevitably lead to poor PtO solutions, resulting in its least accuracy. 
On the other hand, the \purePOP{} and \mcPOP{} approaches, which prioritize learning the PtO map (encoder), achieve better PtO solution accuracy than \pureOPO{}. 
Consequently, the inaccuracy of encoder outputs (due to the nonlinear PtO map) propagates through the decoder, resulting in the least accurate inverse solutions, even with the forward solver as the physics constraint.

\cref{fig:2D_Heat_accuracy_TNetAEfull_framework} shows 
a (random) representative inverse (conductivity field) and full state (temperature field) forward solutions obtained by \TNetAEfull{} trained with one training sample coupled with data randomization of noise level $\epsilon = 10\%$. It can be seen that the inverted conductivity field exhibits accuracy comparable to the Tikhonov regularization solution, and both closely approximate the true conductivity field distribution. Furthermore, the predicted temperature field demonstrates excellent agreement with the ground truth solution. These results underscore the effectiveness of combining model-constrained learning with data randomization techniques in the \TNetAEfull{} framework. We would like to point out that the significant difference in the test case in \cref{fig:2D_Heat_accuracy_TNetAEfull_framework} and the training sample in \cref{fig:2D_Heat_Model} demonstrates the framework's robust generalization capabilities to completely different test samples. This desirable characteristic enables the development of inverse and forward surrogate models using minimal training data\textemdash namely one single training sample without ground truth PoI\textemdash while maintaining reliable performance on unseen test cases. We would like to point out that such a capability comes with an offline cost: differentiable forward solver is required in this paper. One remedy is to train a surrogate (e.g. neural network) for the forward map or to use a differentiable reliable surrogate if available. 


{\bf \TNetAEfull{}  robustness to a wide range of noise levels.}
We further investigate the robustness of \TNetAEfull{} across varying noise levels using a single training sample. \cref{fig:Heat_noise_level} demonstrates that \TNetAEfull{} achieves relative errors comparable to the Tikhonov regularization framework across a broad noise spectrum (8\% to 20\%) for 500 test inverse solutions. Outside this ``optimal range", \TNetAEfull{} does not yield acceptable accuracy. On the one hand,  with low noise levels, the data randomization technique insufficiently explores the space of potentially unseen test samples, limiting generalization capabilities. On the other hand, excessive noise levels result in overwhelmingly corrupted samples that are statistically indistinguishable, and thus degrading the accuracy of the learned surrogate models. These observations are consistent with the theoretical prediction in \cref{sect:nonlinearProbabilistic}. We would like to point out that the minimum relative error achieved by \TNetAEfull{} matches that of the Tikhonov regularization framework, indicating that, as designed, \TNetAEfull{} successfully learns the Tikhonov regularization solver using only a single training sample without requiring ground truth PoI data.

{\bf \TNetAEfull{} robustness to arbitrary single-sample.}
In this section, we investigate the robustness of the \TNetAEfull{} framework to one arbitrarily chosen sample for training. To that end, ten independent training instances are conducted, each utilizing a single sample randomly selected from a pool of 100 training samples. The left figure in \cref{fig:2D_Heat_random_traning_samples} presents the spatial distribution of 10 observational points, including their location indices. The right figure of \cref{fig:2D_Heat_random_traning_samples} shows the statistical characteristics (mean and standard deviation) of 10000 distinct observation samples, and the observation magnitudes for the 10 randomly selected training cases at their respective locations.
Statistical analysis of the \TNetAEfull{} performance across these 10 distinct training instances yields an average relative error of 45.32\% (again similar to the Tikhonov regularization error) with a standard deviation of 0.32\% for inverse solutions. This small variance in relative error metrics demonstrates the \TNetAEfull{} robustness with respect to an arbitrary individual sample in the single-sample training scenario. In particular, the result shows that the prediction error is similar for any of these 10 individual random samples when used in the \TNetAEfull{} as the only training sample.






\subsection{Navier-Stokes equations}
\seclab{2D_NS}

The vorticity form of 2D Navier--Stokes equation for viscous and incompressible fluid \cite{FouierOP} is written as
\begin{equation*}
    \begin{aligned}
        \partial_t \omega(x,t) + v(x,t) \cdot \nabla \omega(x,t) & = \nu \Delta \omega(x,t) + f(x), & \quad x \in \LRp{0,1}^2, t \in (0, T], \\
        \nabla \cdot v(x,t)                                      & = 0,                             & \quad x \in \LRp{0,1}^2, t \in (0, T], \\
        \omega(x,0)                                              & = u(x),                   & \quad x \in \LRp{0,1}^2,
    \end{aligned}
\end{equation*} 
where $v \in \LRp{0,1}^2 \times (0, T]$ denotes the velocity field, $\omega = \nabla \times v$ represents the vorticity, and $u$ defines the initial vorticity which is the parameter of interest. The forcing function is specified as $f(x) = 0.1 \LRp{\sin \LRp{2 \pi \LRp{x_1 + x_2}} + \cos\LRp{2 \pi \LRp{x_1 + x_2}}}$, with viscosity coefficient $\nu = 10^{-3}$. The computational domain is discretized using a uniform $32 \times 32$ mesh in space, while the temporal domain $t \in (0, 10]$ is partitioned into 1000 uniform time steps with $\Delta t = 10^{-2}$. The inverse problem aims to reconstruct the initial vorticity field $u$ from vorticity measurements $\yb$ collected at 20 random spatial locations form the vorticity field $\omega$ at the final time $T = 10$.

\begin{figure}[htb!]
    \centering
    \begin{adjustbox}{center}
        \begin{tabular*}{\textwidth}{c@{\hskip -0.01cm} c@{\hskip -0.01cm} c@{\hskip -0.01cm} }
            \centering
            \raisebox{-0.5\height}{$\quad \quad \quad \quad \quad \quad $} &
            \raisebox{-0.5\height}{\includegraphics[width = 0.34\textwidth]{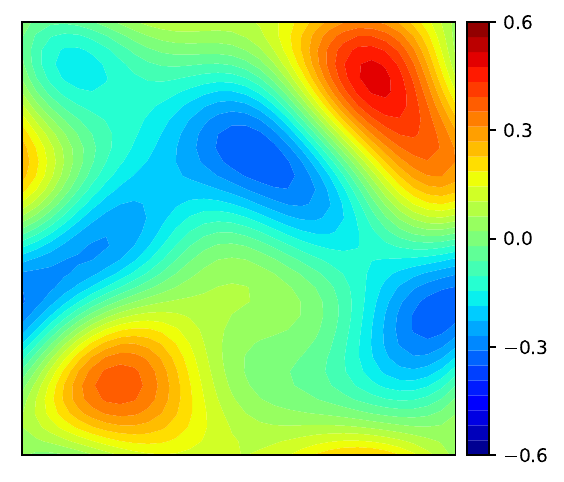}} &
            \raisebox{-0.5\height}{\includegraphics[width = 0.34\textwidth]{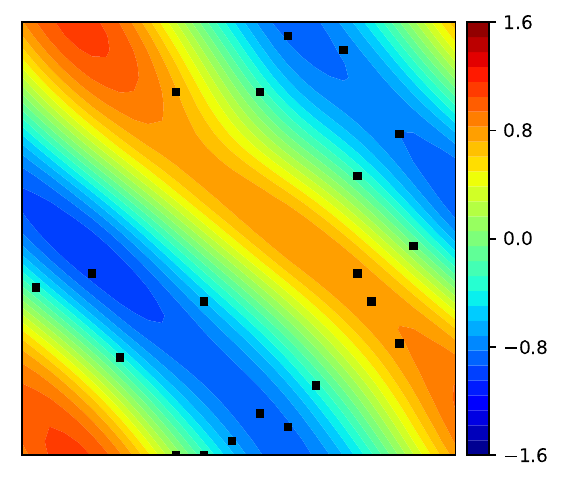}}
        \end{tabular*}
    \end{adjustbox}
    \caption{{\bf 2D Navier--Stokes equation.} {\bf Left:} A sample of the PoI $\ub$. {\bf Right:} A corresponding vorticity field $\ybfull$ at final time $T = 10$, observation $\yb$ are extracted at 20 random observed points. This pair of PoI and observation/vorticity field is used for training in one training sample case.}
    \figlab{2D_NS_Model}
\end{figure}

\begin{table}[htb!]
    \centering
    \caption{{\bf 2D Navier--Stokes equation.} Average relative error for inverse solutions and PtO/forward solutions obtained by all approaches trained with $\LRc{1,100}$ training samples. The model-constrained approaches are more accurate for both inverse (comparable to the Tikhonov\textemdash Tik\textemdash approach) and forward solution, and within the model-constrained approaches, \TNetAE{} and \TNetAEfull{} are the most accurate ones: in fact one training sample is sufficient for these two methods.}
    \begin{tabular}{c|rr|rr}
        \hline
                          & \multicolumn{2}{c|}{\textbf{1 training sample}} & \multicolumn{2}{c}{\textbf{100 training samples}}                                       \\
        \textbf{Approach} & \textbf{Inverse} (\%)                                & \textbf{Forward}                                  & \textbf{Inverse} (\%) & \textbf{Forward} \\
        \hline
        $\purePOP$        & 156.99                                          & 2.99 $\times 10^{-1}$                                          & 72.22            & 6.72 $\times 10^{-2}$         \\
        $\pureOPO$        & 103.94                                          & 5.60                                                           & 40.20            & 5.94 $\times 10^{-1}$         \\
        $\mcPOP$          & 161.48                                          & 2.99 $\times 10^{-1}$                                          & 76.33            & 6.72 $\times 10^{-2}$         \\
        $\mcOPO$          & 46.43                                           & 5.15 $\times 10^{-1}$                                          & 27.29            & 2.20 $\times 10^{-3}$         \\
        $\mcOPOfull$      & 46.43                                           & 3.79 $\times 10^{-1}$                                          & 27.29            & 2.12 $\times 10^{-3}$         \\
        $\TNetAE$         & 25.68                                           & 2.14 $\times 10^{-3}$                                          & 24.54            & 1.49 $\times 10^{-3}$         \\
        $\TNetAEfull$     & 25.68                                           & 2.10 $\times 10^{-3}$                                          & 24.54            & 1.45 $\times 10^{-3}$         \\
        Tik               & 22.71                                           &                                                                & 22.71            &                  \\
        \hline
    \end{tabular}
    \tablab{2D_NS_accuracy_table}
\end{table}

\begin{figure}[htbp]
    \centering
    \resizebox{.96\textwidth}{!}{
    \begin{tabular*}{\textwidth}{c@{\hskip -0.001cm} c@{\hskip -0.01cm} c@{\hskip -0.01cm} c@{\hskip -0.01cm} c@{\hskip -0.01cm}}
        \centering
        & \multicolumn{2}{c}{\textbf{1 training sample}}
        & \multicolumn{2}{c}{\textbf{100 training samples}}
        \\
        \rotatebox[origin=c]{90}{} &
        \raisebox{-0.5\height}{mean $ \quad $} &
        \raisebox{-0.5\height}{std $ \quad $} &
        \raisebox{-0.5\height}{mean $ \quad $} &
        \raisebox{-0.5\height}{std $ \quad $}
        \\
        \rotatebox[origin=c]{90}{$\purePOP$} &
        \raisebox{-0.5\height}{\includegraphics[width = 0.22\textwidth]{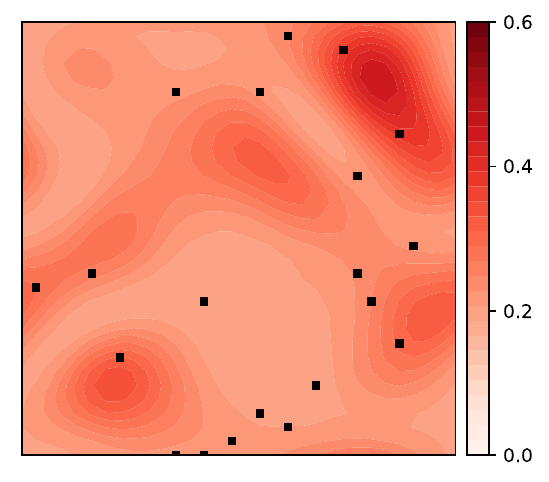}} &
        \raisebox{-0.5\height}{\includegraphics[width = 0.22\textwidth]{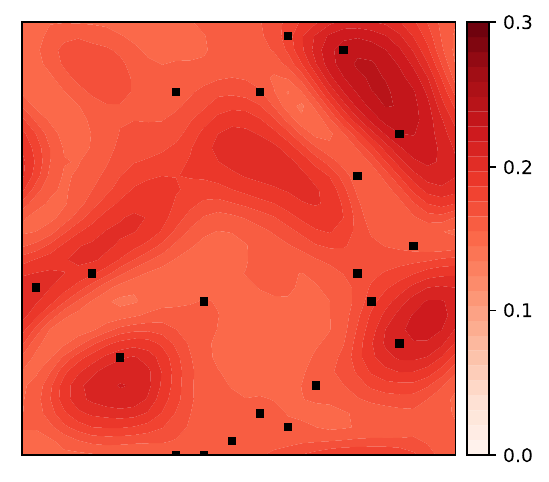}} &
        \raisebox{-0.5\height}{\includegraphics[width = 0.22\textwidth]{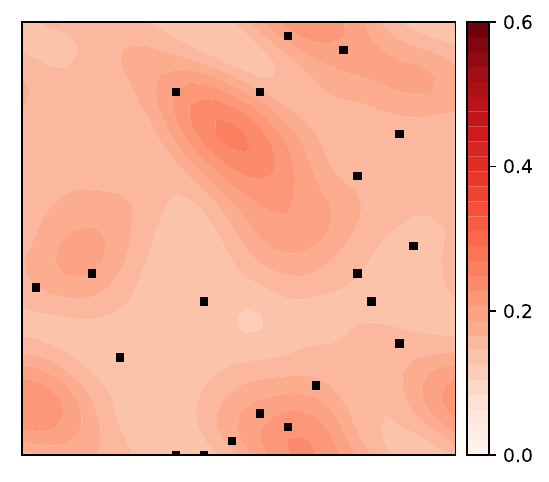}} &
        \raisebox{-0.5\height}{\includegraphics[width = 0.22\textwidth]{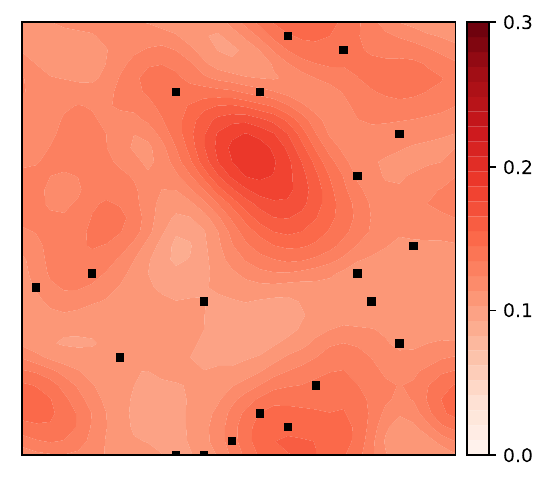}}
        \\
        \rotatebox[origin=c]{90}{$\pureOPO$} &
        \raisebox{-0.5\height}{\includegraphics[width = 0.22\textwidth]{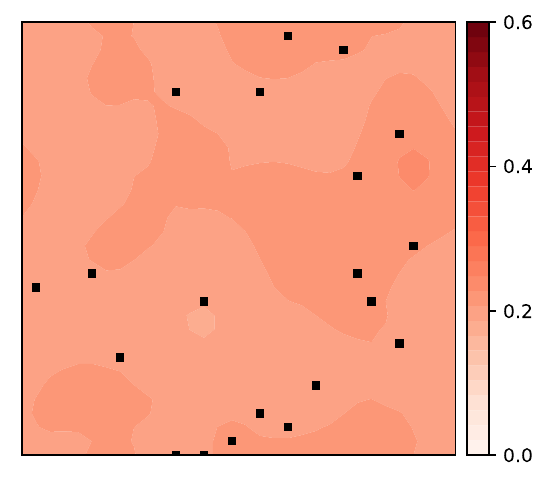}} &
        \raisebox{-0.5\height}{\includegraphics[width = 0.22\textwidth]{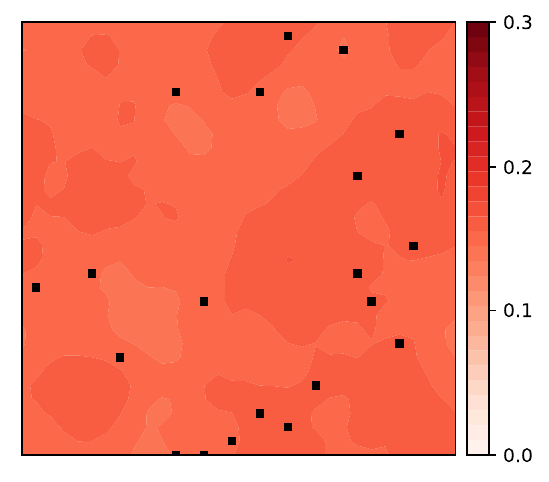}} &
        \raisebox{-0.5\height}{\includegraphics[width = 0.22\textwidth]{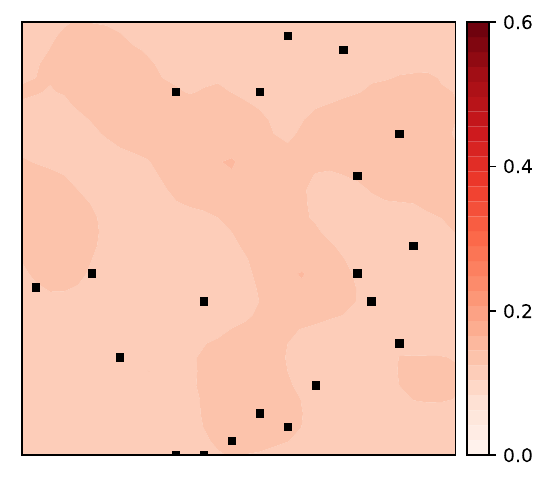}} &
        \raisebox{-0.5\height}{\includegraphics[width = 0.22\textwidth]{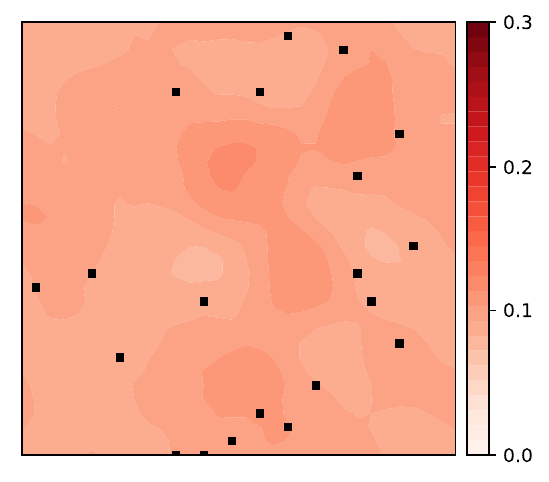}}
        \\
        \rotatebox[origin=c]{90}{$\mcPOP$} &
        \raisebox{-0.5\height}{\includegraphics[width = 0.22\textwidth]{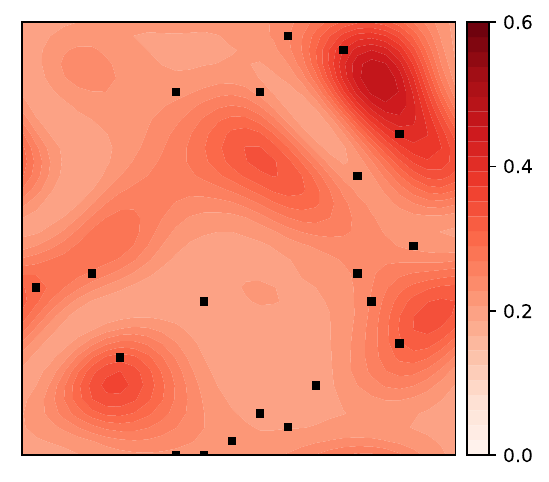}} &
        \raisebox{-0.5\height}{\includegraphics[width = 0.22\textwidth]{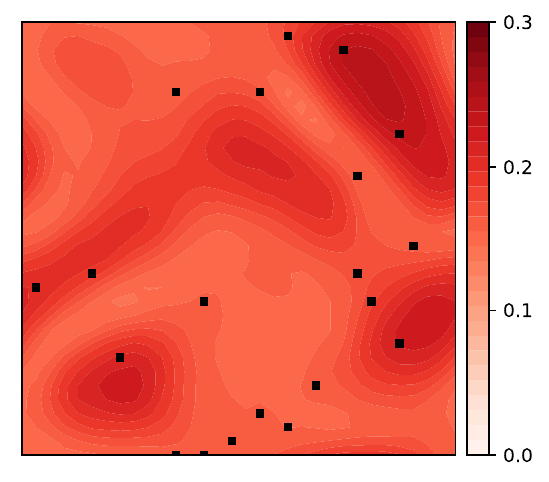}} &
        \raisebox{-0.5\height}{\includegraphics[width = 0.22\textwidth]{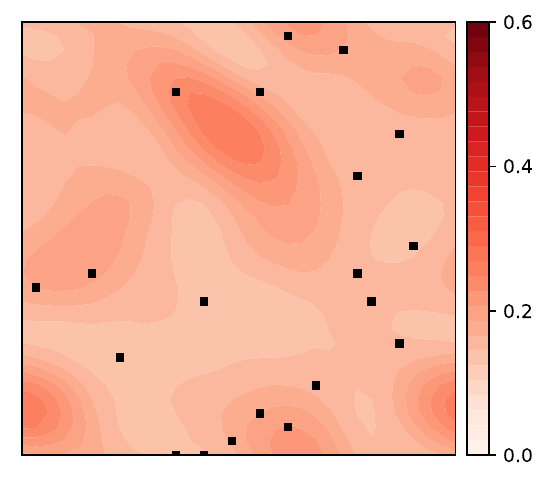}} &
        \raisebox{-0.5\height}{\includegraphics[width = 0.22\textwidth]{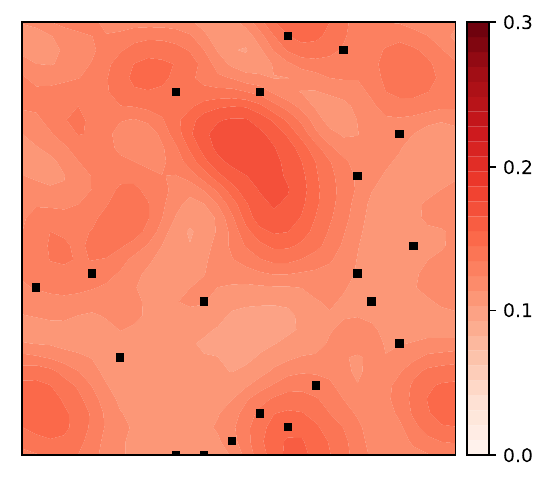}}
        \\
        \rotatebox[origin=c]{90}{$\begin{matrix} \mcOPO \\ \mcOPOfull \end{matrix}$} &
        \raisebox{-0.5\height}{\includegraphics[width = 0.22\textwidth]{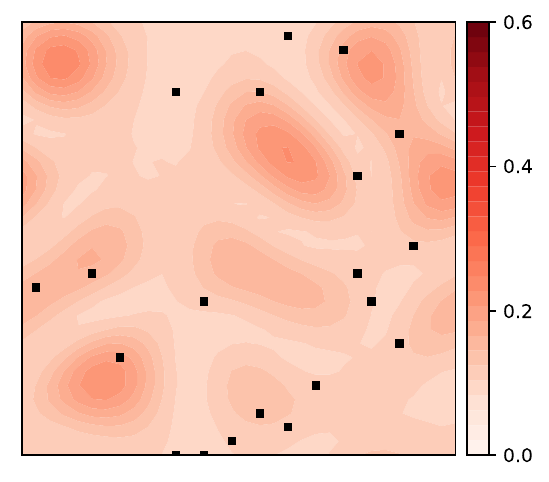}} &
        \raisebox{-0.5\height}{\includegraphics[width = 0.22\textwidth]{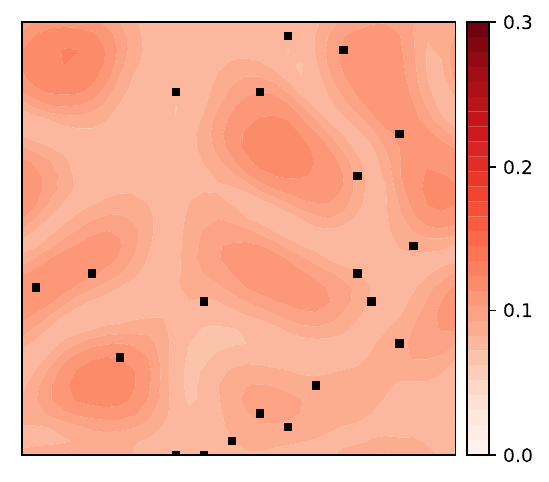}} &
        \raisebox{-0.5\height}{\includegraphics[width = 0.22\textwidth]{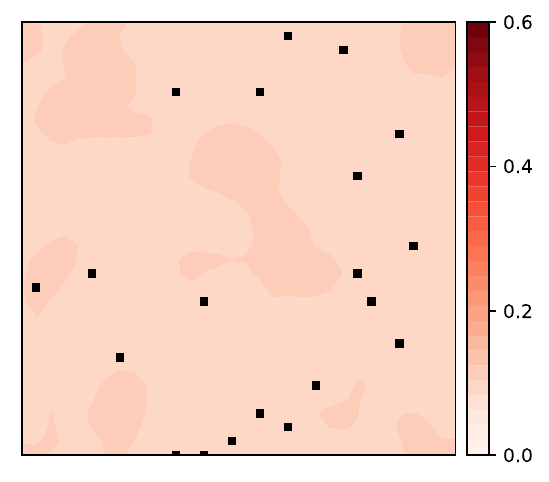}} &
        \raisebox{-0.5\height}{\includegraphics[width = 0.22\textwidth]{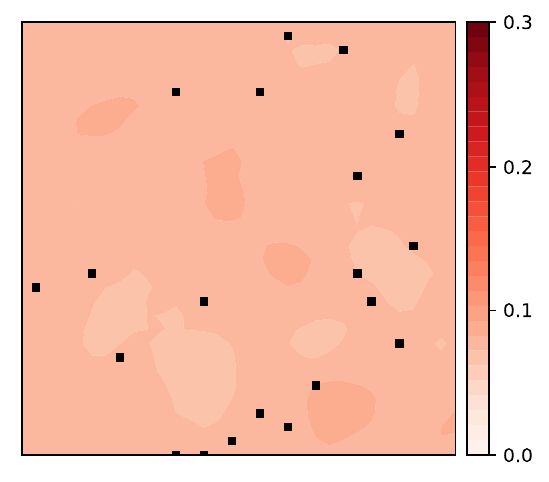}}
        \\
        \rotatebox[origin=c]{90}{$\begin{matrix} \TNetAE \\ \TNetAEfull \end{matrix}$} &
        \raisebox{-0.5\height}{\includegraphics[width = 0.22\textwidth]{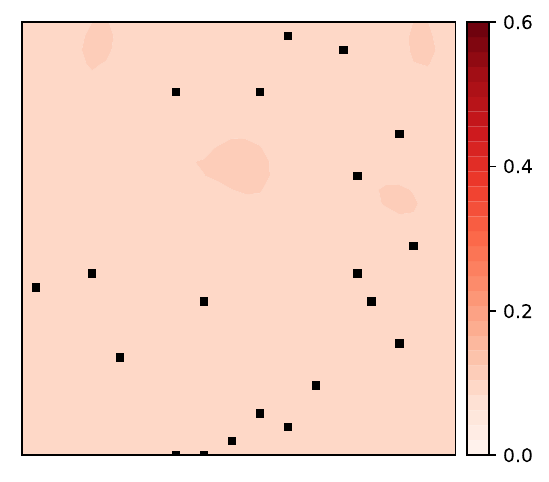}} &
        \raisebox{-0.5\height}{\includegraphics[width = 0.22\textwidth]{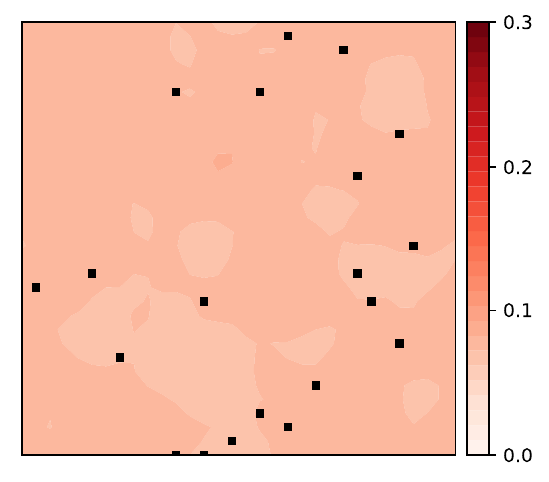}} &
        \raisebox{-0.5\height}{\includegraphics[width = 0.22\textwidth]{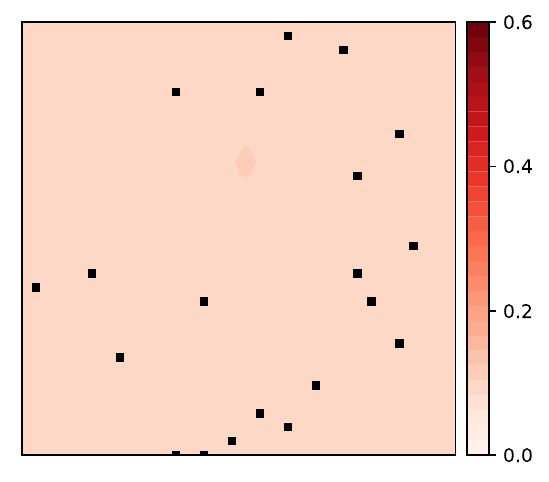}} &
        \raisebox{-0.5\height}{\includegraphics[width = 0.22\textwidth]{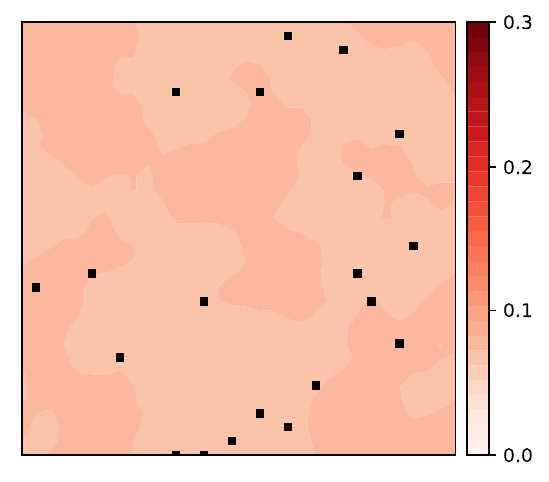}}
        \\
        \rotatebox[origin=c]{90}{Tik} &
        \raisebox{-0.5\height}{\includegraphics[width = 0.22\textwidth]{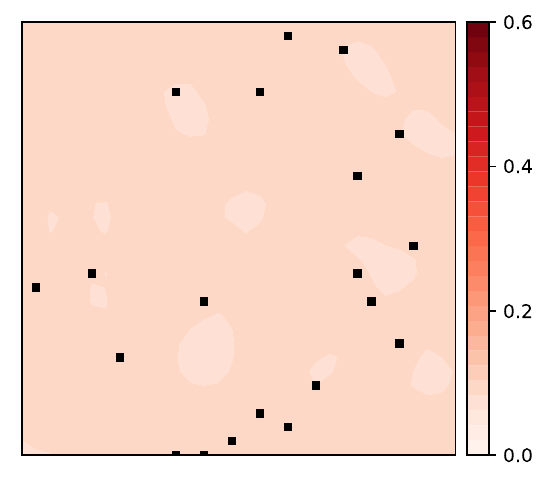}} &
        \raisebox{-0.5\height}{\includegraphics[width = 0.22\textwidth]{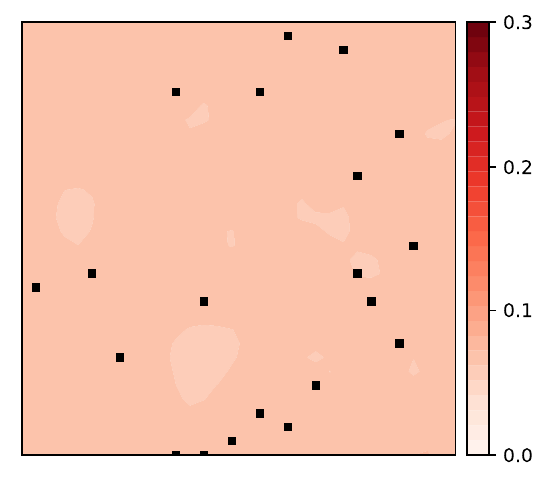}} & &
    \end{tabular*}
    }
    \caption{{\bf 2D Navier--Stokes equation.} Mean and standard deviation of absolute error of 500 test inverse solutions obtained from different approaches. Black points are observation locations.  Relatively to the Tikhonov approach (Tik), the model-constrained approaches are more accurate, and within the model-constrained approaches, \TNetAE{} and \TNetAEfull{} are the most accurate ones: in fact one training sample is sufficient for these two methods.}
    \figlab{2D_NS_accuracy_inverse}
\end{figure}

\begin{figure}[htb!]
    \centering
    \begin{tabular*}{\textwidth}{c@{\hskip -0.001cm} c@{\hskip -0.01cm} c@{\hskip -0.01cm} c@{\hskip -0.01cm} c@{\hskip -0.01cm} c@{\hskip -0.01cm} c@{\hskip -0.01cm}}
        \centering
        \rotatebox[origin=c]{90}{} &
        \raisebox{-0.5\height}{$\begin{matrix} \purePOP{}/\mcPOP{} \end{matrix}$} &
        \raisebox{-0.5\height}{\pureOPO{}} &
        \raisebox{-0.5\height}{\mcOPO{}} &
        \raisebox{-0.5\height}{\mcOPOfull{}} &
        \raisebox{-0.5\height}{\TNetAE{}} &
        \raisebox{-0.5\height}{\TNetAEfull{}}
        \\
        \rotatebox[origin=c]{90}{1 sample} &
        \raisebox{-0.5\height}{\includegraphics[width = 0.16\textwidth]{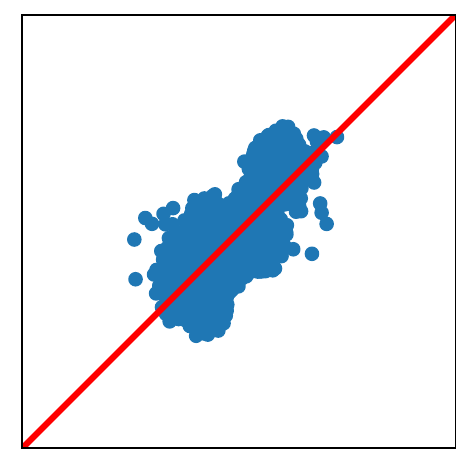}} &
        \raisebox{-0.5\height}{\includegraphics[width = 0.16\textwidth]{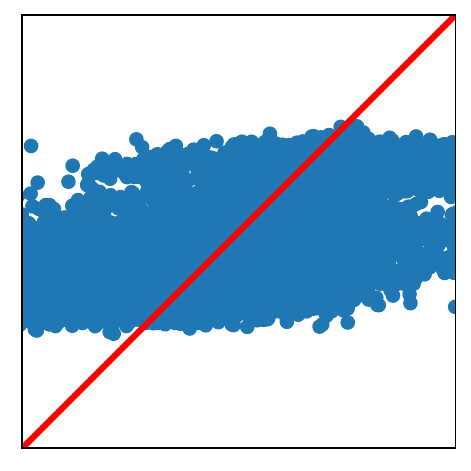}} &
        \raisebox{-0.5\height}{\includegraphics[width = 0.16\textwidth]{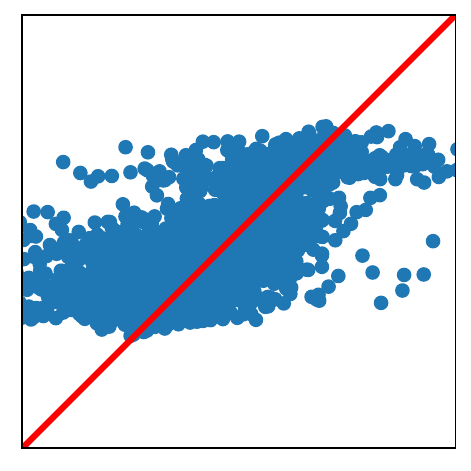}} &
        \raisebox{-0.5\height}{\includegraphics[width = 0.16\textwidth]{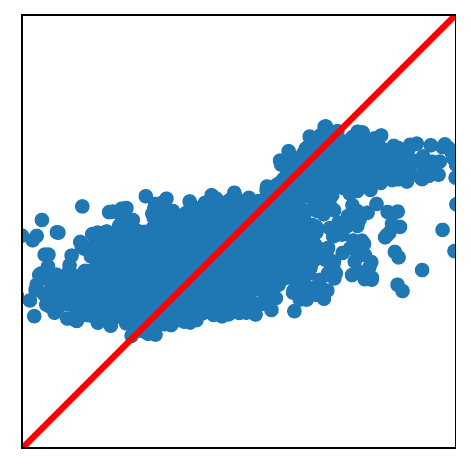}} &
        \raisebox{-0.5\height}{\includegraphics[width = 0.16\textwidth]{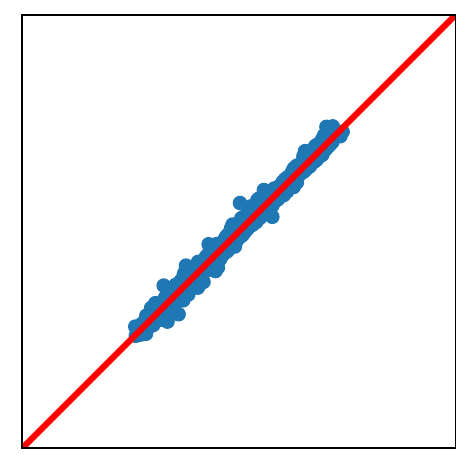}} &
        \raisebox{-0.5\height}{\includegraphics[width = 0.16\textwidth]{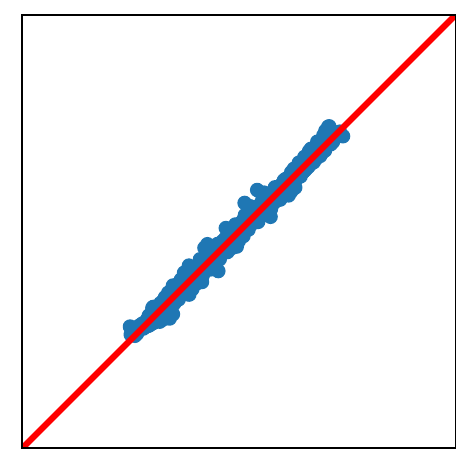}}
        \\
        \rotatebox[origin=c]{90}{100 samples} &
        \raisebox{-0.5\height}{\includegraphics[width = 0.16\textwidth]{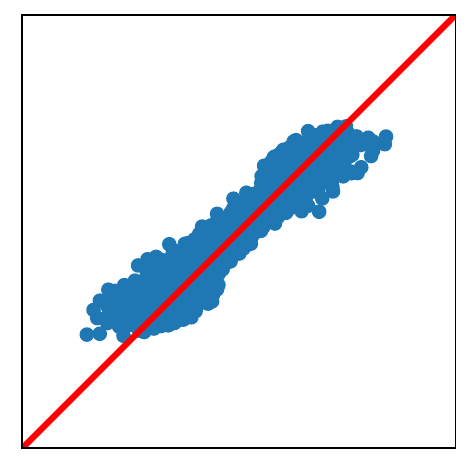}} &
        \raisebox{-0.5\height}{\includegraphics[width = 0.16\textwidth]{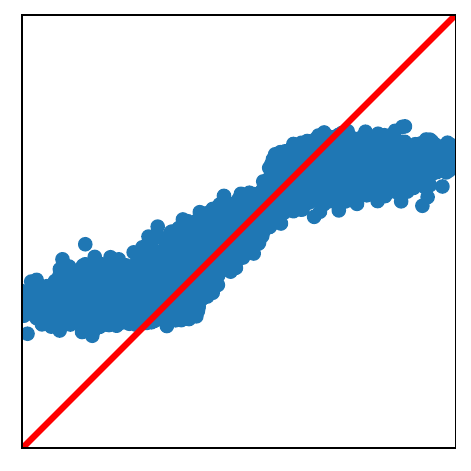}} &
        \raisebox{-0.5\height}{\includegraphics[width = 0.16\textwidth]{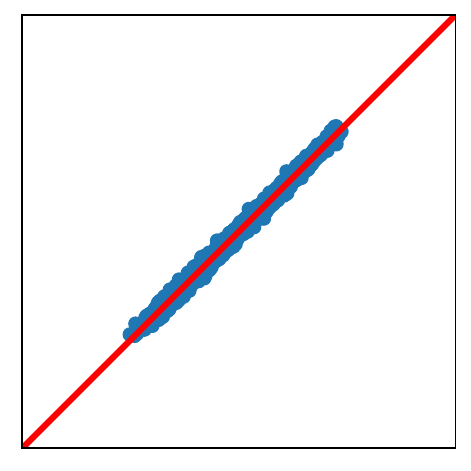}} &
        \raisebox{-0.5\height}{\includegraphics[width = 0.16\textwidth]{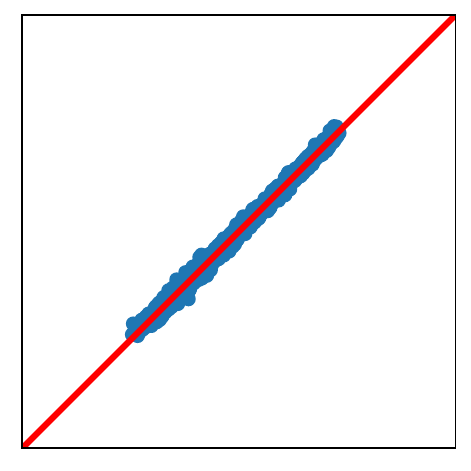}} &
        \raisebox{-0.5\height}{\includegraphics[width = 0.16\textwidth]{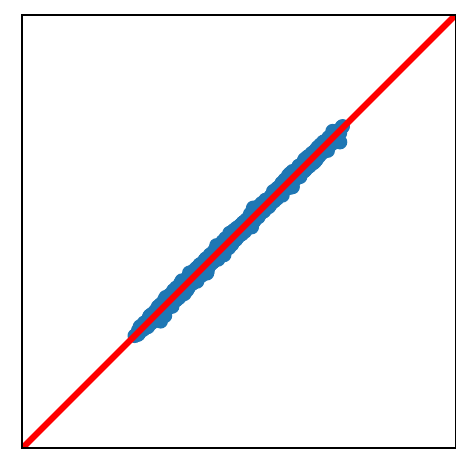}} &
        \raisebox{-0.5\height}{\includegraphics[width = 0.16\textwidth]{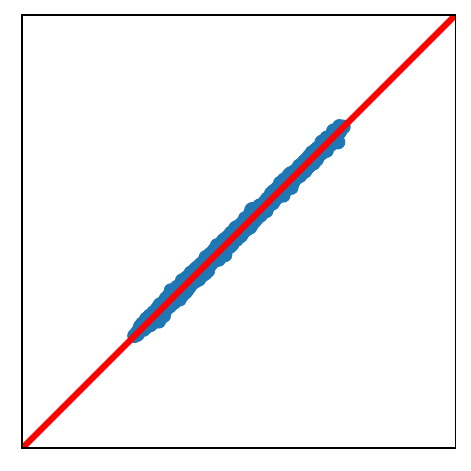}}
    \end{tabular*}
    \caption{{\bf 2D Navier--Stokes equation.} The comparison of the predicted observations on 500 test samples. In all plots plot, the x-axis is the magnitude of the true observation, and the y-axis is the magnitude of the predicted observation, both axes have a range of $\LRs{-3,3}$. The red line indicates the perfect matching between predictions and the ground truth observation data set. {\bf Top row:} Trained with $1$ training sample. {\bf Bottom row:} Trained with $100$ training samples.  As can be seen, model-constrained approaches are more accurate, and within the model-constrained approaches, \TNetAE{} and \TNetAEfull{} are the most accurate ones: in fact one training sample is sufficient for these two methods.}
    \figlab{2D_NS_accuracy_forward}
\end{figure}

\begin{figure}[htb!]
    \centering
    \begin{tabular*}{\textwidth}{c@{\hskip 0.001cm} c@{\hskip -0.01cm} c@{\hskip -0.01cm} c@{\hskip -0.01cm} c@{\hskip -0.01cm}}
        \centering
        & \multicolumn{2}{c}{\textbf{1 training sample}}
        & \multicolumn{2}{c}{\textbf{100 training samples}}
        \\
        &
        \raisebox{-0.5\height}{mean \quad} &
        \raisebox{-0.5\height}{std \quad} &
        \raisebox{-0.5\height}{mean \quad}&
        \raisebox{-0.5\height}{std \quad}
        \\
        \rotatebox[origin=c]{90}{$\mcOPOfull$} &
        \raisebox{-0.5\height}{\includegraphics[width = 0.25\textwidth]{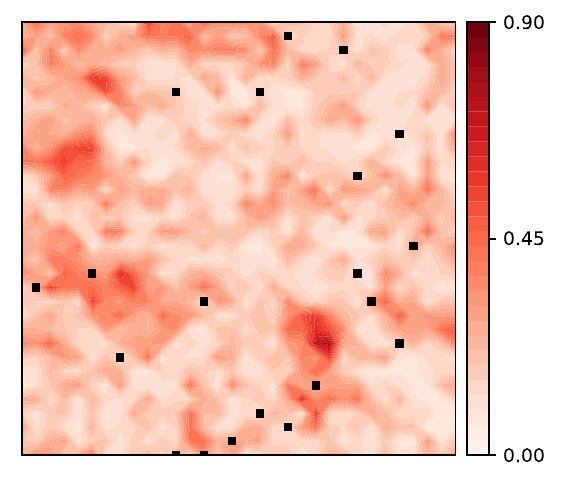}} &
        \raisebox{-0.5\height}{\includegraphics[width = 0.25\textwidth]{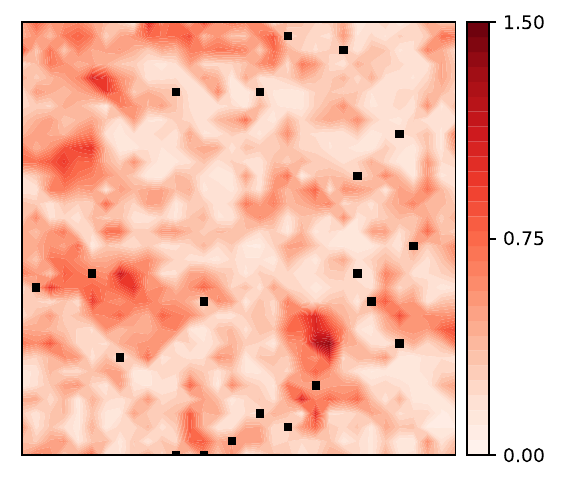}} &
        \raisebox{-0.5\height}{\includegraphics[width = 0.25\textwidth]{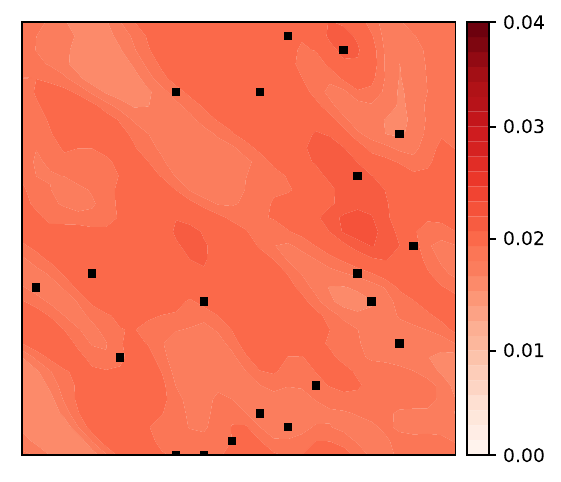}} &
        \raisebox{-0.5\height}{\includegraphics[width = 0.25\textwidth]{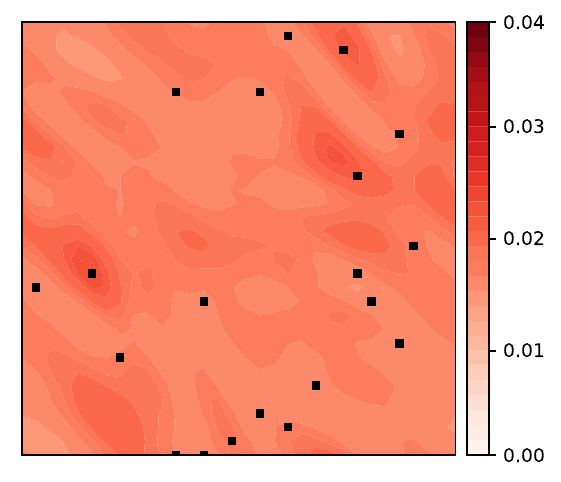}}
        \\
        \rotatebox[origin=c]{90}{$\TNetAEfull$} &
        \raisebox{-0.5\height}{\includegraphics[width = 0.25\textwidth]{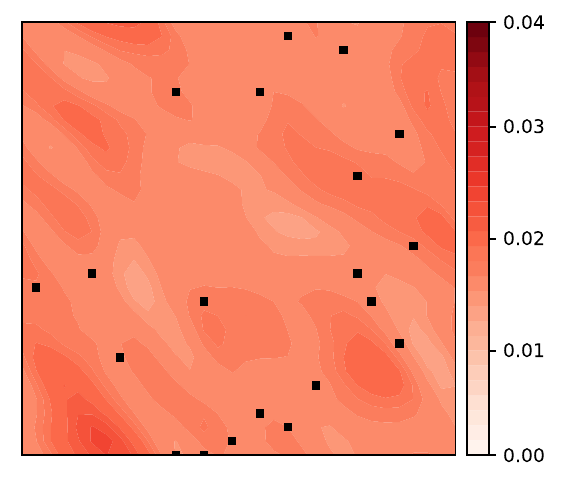}} &
        \raisebox{-0.5\height}{\includegraphics[width = 0.25\textwidth]{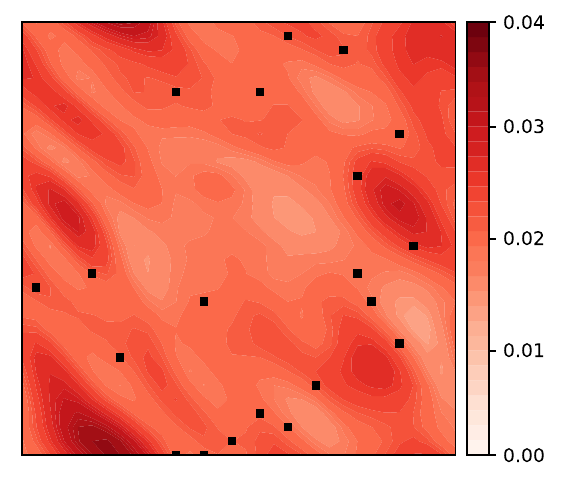}} &
        \raisebox{-0.5\height}{\includegraphics[width = 0.25\textwidth]{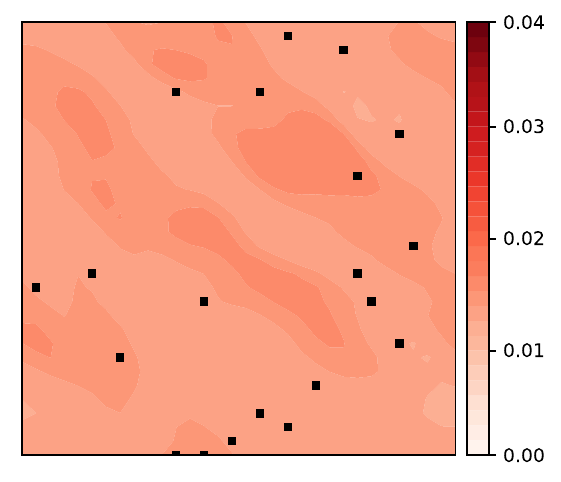}} &
        \raisebox{-0.5\height}{\includegraphics[width = 0.25\textwidth]{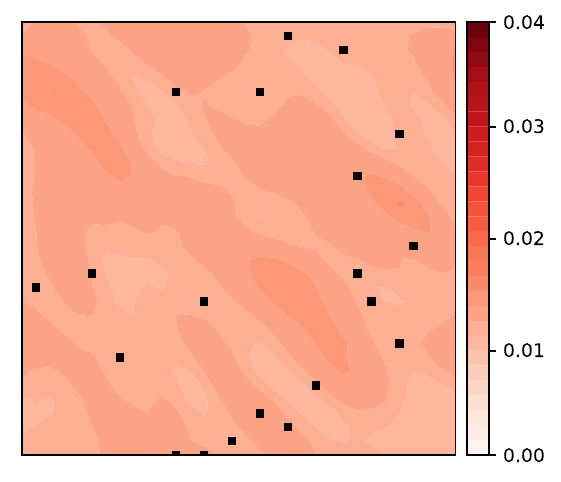}}
    \end{tabular*}
    \caption{{\bf 2D Navier--Stokes equation.} Mean and standard deviation of absolute pointwise error for 500 test vorticity solutions at $T = 10$ obtained from $\mcOPOfull$ and $ \TNetAEfull{}$. Black points are observational locations. $\TNetAEfull{}$ is more accurate, especially for the case with one training sample in which it achieves two orders of magnitude smaller error.}
    \figlab{2D_NS_accuracy_forward_full}
\end{figure}

{\bf Generating train and test data sets.}
To generate data pairs of $\LRp{u, \omega}$, we draw samples of $u(x)$ using the truncated Karhunen-Loève expansion
\begin{equation*}
    \eqnlab{Naiver_KL}
    u(x) = \sum_{i=1}^{24} \sqrt{\lambda_i} \, {\bf \phi}_i(x) \, z_i,
\end{equation*}
where $z_i \sim \mathcal{N} \LRp{0, 1}, i = 1, \hdots, 24$, and $\LRp{\lambda_i, {\bf \phi}_i}$ are eigenpairs obtained by the eigendecomposition of the covariance operator $ 7^{\frac{3}{2}} \LRp{-\Delta + 49 \textbf{I}}^{-2.5}$ with periodic boundary conditions. Next, we discretize the initial vorticity $u(x)$, denoted as $\ub$, and we solve the Navier-Stokes equation by the stream-function formulation with a pseudospectral method \cite{FouierOP} to obtain a discrete representation $\ybfull$ of $\omega\LRp{x,t}$ at time $t = 10$. The observation data $\yb$ consists of the vorticity field $\ybfull$ at $T = 10$ at 20 randomly distributed observational locations, with the subsequent addition of $\delta = 2\%$ Gaussian noise. Two distinct datasets are generated: a training set comprising 100 independent samples and a test set containing 500 samples. \cref{fig:2D_NS_Model} illustrates the first sample pair from the training dataset of 100 samples, which serves as the training sample for single-sample training scenarios for all approaches.

{\bf Learned inverse and PtO/forward maps accuracy.} Following the same procedure used for the 2D Poisson equation, encoder, and decoder networks are trained sequentially for two cases: using a single training sample pair (shown in \cref{fig:2D_NS_Model}) and using 100 training sample pairs. For \mcOPO{}, \mcOPOfull{}, \TNetAE{} and \TNetAEfull{} approaches, randomization are performed at each epoch with noise level $\epsilon = 25\%$.
\cref{tab:2D_NS_accuracy_table} presents the average relative errors of inverse and PtO/forward solutions computed from 500 test samples for all approaches. 

When trained on a single sample, \TNetAE{} and \TNetAEfull{} achieve the lowest average relative error of 25.68\% (closest to the gold-standard Tikhonov regularization solution with 22.71\% error) for inverse solutions among all approaches. Similarly, their PtO/forward solutions demonstrate superior accuracy with average relative errors of $2.14 \times 10^{-3}$ and $2.10 \times 10^{-3}$, respectively. This generalization accuracy is owing to the combination of data randomization and forward solver model-constrained terms. 
The \mcOPO{} and \mcOPOfull{} approaches come in second with regarding to the accuracy for inverse solutions, with a relative error of 46.43\%. This reduced accuracy (relatively to \TNetAE{} and \TNetAEfull{}) stems from a strong bias toward the single training sample, despite the forward solver constraint, as predicted in \cref{sect:mc_OPO}.
 \mcPOP{}, \pureOPO{}, and \purePOP{} exhibit significantly higher average relative errors of 161.48\%, 103.94\%, and 156.99\% respectively for inverse solutions. These poor results are expected for \pureOPO{} and \purePOP{} due to their purely data-driven nature, thus limiting generalization. For \mcPOP{}, the inaccurate encoder (learned PtO map) propagates errors to the decoder training, resulting in imprecise inverse surrogate models. For PtO/forward solutions, \purePOP{}, \pureOPO{}, \mcPOP{}, \mcOPO{}, and \mcOPOfull{} fail to produce accurate surrogate models, again due to overfitting to the single training sample despite forward solver regularization. 

When we increase the number of training samples to 100, and thus providing more information about the problem under consideration, significant accuracy improvements are observed for all approaches for both inverse and PtO/forward surrogate models.
\TNetAE{} and \TNetAEfull{} approaches maintain the best performance, achieving average relative errors of 24.54\% for inverse solutions compared to the Tikhonov (Tik) method with of 22.71\%. Their PtO/forward solutions exhibit good accuracy with average relative errors of $1.49 \times 10^{-3}$ and $1.45 \times 10^{-3}$, respectively. 
The average relative error of the inverse solution obtained by \mcOPO{} and \mcOPOfull{} is $27.29\%$, which is the second best among all approaches. The relative error of the PtO/forward solution obtained by \mcOPO{} and \mcOPOfull{} is $2.20 \times 10^{-3}$ and $2.12 \times 10^{-3}$, respectively, which is almost as good as \TNetAE{} and \TNetAEfull{}. This indicates the roles of model-constrained terms in reducing the overfitting effect when sufficient training data is provided.
In contrast, \purePOP{} and \mcPOP{} show substantially higher relative errors of 72.22\% and 76.33\%, respectively, for inverse solutions. These high errors stem from inaccuracies in their pre-trained PtO map (encoder), consistent with observations from the single-sample training scenario.
Meanwhile, \pureOPO{} framework shows improved inverse solution accuracy with a relative error of 40.20\%, yet remains less accurate than the model-constrained approaches (\mcOPO{}, \mcOPOfull{}, \TNetAE{}, and \TNetAEfull{}). This reduced performance is consistent with the error analysis for linear problems in \cref{sect:mc_autoencoder}. 
Moreover, \pureOPO{}'s PtO solution accuracy remains notably poor ($5.94 \times 10^{-1}$) despite the richer training dataset with 100 data pairs, reflecting the inherent PtO mapping errors analyzed in \cref{sect:naive_OPO}. 

We further elaborate on the accuracy of forward and inverse surrogate models obtained from all approaches. \cref{fig:2D_NS_accuracy_inverse} illustrates the spatial distribution of error statistics, presenting both mean and standard deviation of the absolute errors between predicted and ground truth inverse solutions for 500 test cases. \TNetAE{} and \TNetAEfull{} frameworks demonstrate superior performance, exhibiting the lowest mean and standard deviation of absolute errors across the domain for both single-sample and 100-sample training scenarios. 
The \mcOPO{} and \mcOPOfull{} approaches achieve comparable accuracy only when trained with 100 samples, while their single-sample training results show substantially high mean and standard deviation absolute error values. This performance difference underscores the enhanced generalization capabilities of the \TNetAE{} and \TNetAEfull{} frameworks compared to their model-constrained counterparts, \mcOPO{} and \mcOPOfull{}. 
In contrast, the \mcPOP{}, \purePOP{}, and \pureOPO{} approaches consistently demonstrate significantly higher mean and standard deviation of absolute errors across the domain in both training scenarios.

\cref{fig:2D_NS_accuracy_forward} presents a quantitative comparison between predicted observations and ground truth observations for different approaches. In the single-sample training scenario, only \TNetAE{} and \TNetAEfull{} demonstrate accurate predictions, while other approaches exhibit substantial deviations from ground truth values. With the 100-sample training dataset, \mcOPO{} and \mcOPOfull{} join \TNetAE{} and \TNetAEfull{} in achieving significant improvements in observation and vorticity field predictions. However, \purePOP{}, \pureOPO{}, and \mcPOP{} continue to produce inaccurate predictions even with 100 training data.
The capability of \TNetAEfull{} and \mcOPOfull{} to function as direct surrogate solvers for the Navier-Stokes equation deserves particular attention. \cref{fig:2D_NS_accuracy_forward_full} illustrates the spatial distribution of error statistics, showing the mean and standard deviation of absolute pointwise errors between predicted and true vorticity fields across 500 test samples. In the single-sample training scenario, \mcOPOfull{} exhibits high prediction errors, while \TNetAEfull{} maintains good accuracy in vorticity field predictions (in fact two orders of magnitude smaller). The transition to 100-sample training yields marked accuracy improvements for both frameworks in vorticity field predictions, demonstrating their potential as efficient surrogate forward solvers when provided with sufficient training data. 

\begin{figure}[htb!]
    \centering
    \begin{tabular*}{\textwidth}{c@{\hskip -0.01cm} c@{\hskip -0.01cm} c@{\hskip -0.01cm} c@{\hskip -0.01cm}}
        \centering
        \raisebox{-0.5\height}{$\ub_\text{Tik}$} &
        \raisebox{-0.5\height}{$\ub_{\TNetAEfull}$ } &
        \raisebox{-0.5\height}{$\ub_\text{True}$}&
        \raisebox{-0.5\height}{$\yb_{\TNetAEfull}$}
        \\
        \raisebox{-0.5\height}{\includegraphics[width = 0.25\textwidth]{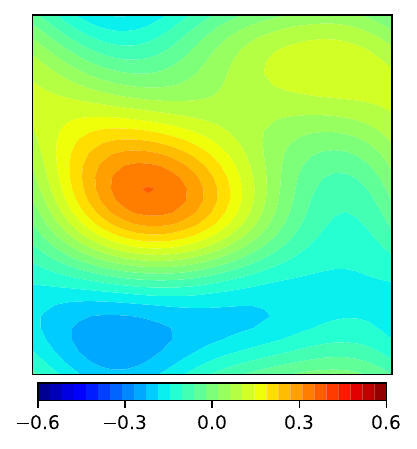}} &
        \raisebox{-0.5\height}{\includegraphics[width = 0.25\textwidth]{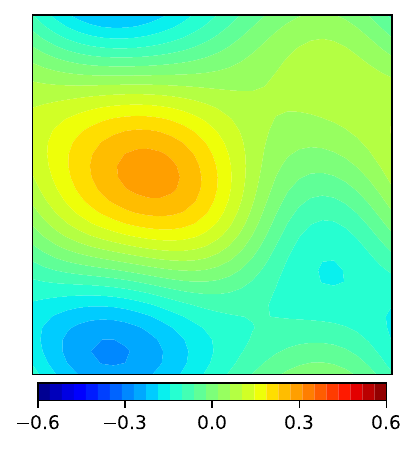}} &
        \raisebox{-0.5\height}{\includegraphics[width = 0.25\textwidth]{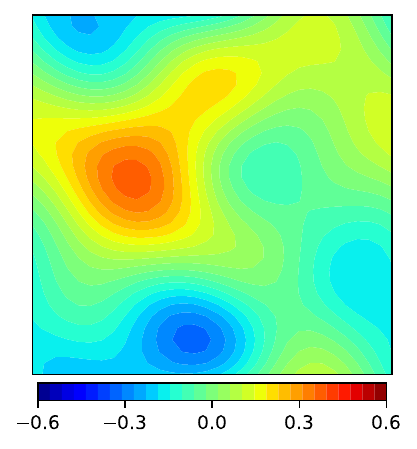}} &
        \raisebox{-0.5\height}{\includegraphics[width = 0.25\textwidth]{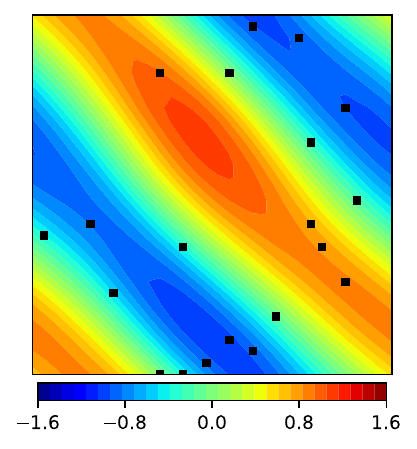}}
        \\
        \raisebox{-0.5\height}{$\snor{\ub_\text{Tik} - \ub_\text{True}}$} &
        \raisebox{-0.5\height}{$\snor{\ub_{\TNetAEfull} - \ub_\text{True}}$ } &
        \raisebox{-0.5\height}{$\yb_\text{True}$}&
        \raisebox{-0.5\height}{$\snor{\yb_{\TNetAEfull} - \yb_\text{True}}$}
        \\
        \raisebox{-0.5\height}{\includegraphics[width = 0.25\textwidth]{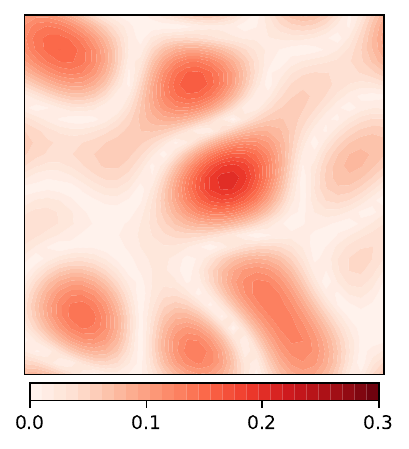}} &
        \raisebox{-0.5\height}{\includegraphics[width = 0.25\textwidth]{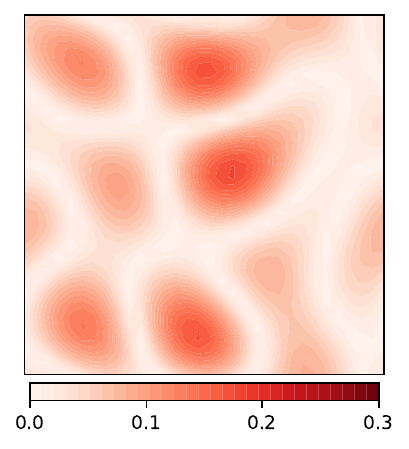}} &
        \raisebox{-0.5\height}{\includegraphics[width = 0.25\textwidth]{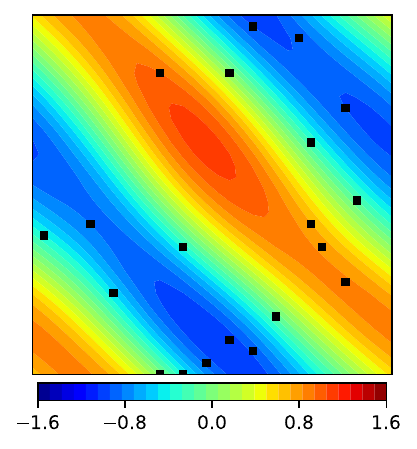}} &
        \raisebox{-0.5\height}{\includegraphics[width = 0.25\textwidth]{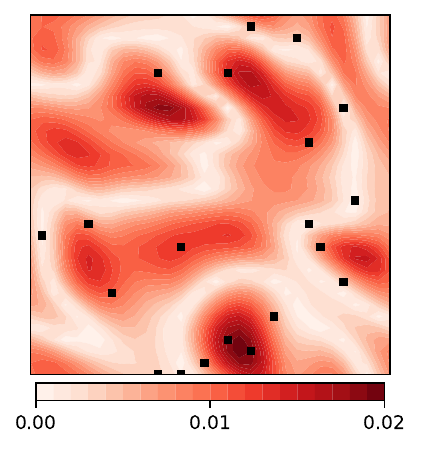}}
    \end{tabular*}
    \caption{{\bf 2D Navier--Stokes equation.}     A (random) representative case of inverse a and full forward solution at $T = 10$ obtained by $\TNetAEfull$ trained with 1 training sample coupled with data randomization of noise level $\sigma = 0.25$
    $\TNetAEfull$ inverse solution is comparable to the Tikhonov (Tik) inverse counterpart, and both are consistent with the ground truth (True). $\TNetAEfull$ full forward solution is almost identical (in fact within 2 digits of accuracy) to the underlying true solution.
    }
    \figlab{2D_NS_accuracy_TNetAEfull_framework}
\end{figure}


\cref{fig:2D_NS_accuracy_TNetAEfull_framework}
shows 
a (random) representative inverse (initial vorticity field) and full state (vorticity field) forward solutions obtained by \TNetAEfull{} at $T = 10$ trained with one training sample coupled with data randomization of noise level $\epsilon = 25\%$.
It can be seen that the inverted initial vorticity field exhibits accuracy comparable to the Tikhonov regularization solution, and both closely approximate the true initial vorticity field. Furthermore, the predicted vorticity field at the final time demonstrates excellent agreement with the ground truth solution. These results underscore the effectiveness of combining model-constrained learning with data randomization techniques in the \TNetAEfull{} framework.
It is important to note that the single training data pair, as shown in \cref{fig:2D_NS_Model}, is completely different from the shown test sample under consideration. This observation demonstrates a generalization capacity of \TNetAEfull{} framework to unseen test samples.

{\bf \TNetAEfull{} robustness to a wide range of noise levels.}
A survey of \TNetAEfull{} trained with one training sample over a wide range of noise levels is shown in \cref{fig:2D_NS_noise_level}. As can be seen, the solution accuracy is robust for a wide range of noise from $\epsilon \in \LRs{0.15, 0.5}$. 
Performance degradation is observed outside of this ``optimal" noise level range. At low noise levels, the data randomization process provides insufficient variation to effectively explore the space of unseen test samples, limiting the framework's ability to leverage forward solver constraints. On the other hand, excessive noise levels result in training data becoming statistically indistinguishable, degrading the framework's capacity to learn accurate inverse mappings. These observations are consistent with the theoretical prediction in \cref{sect:nonlinearProbabilistic}.

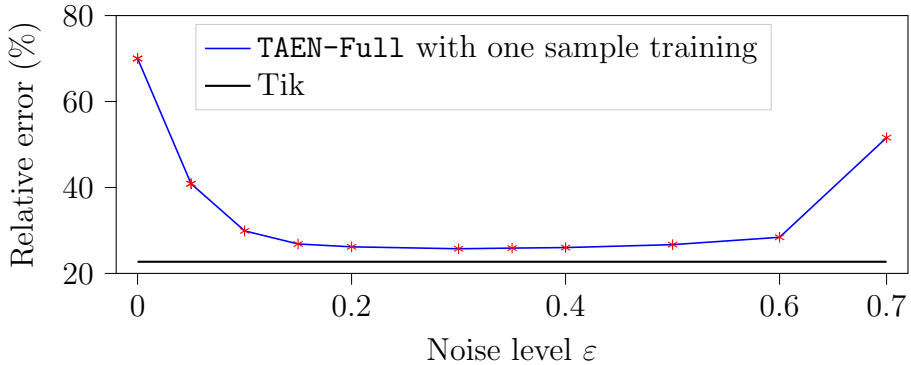
\begin{figure}[htb!]
    \centering
\begin{tikzpicture}[
    ]
    \definecolor{green01270}{RGB}{0,127,0}
    \definecolor{lightgray204}{RGB}{204,204,204}
    \definecolor{darkgray176}{RGB}{176,176,176}

    \begin{axis}[
            width=12cm,
            height=5cm,
            tick align=outside,
            tick pos=left,
            x grid style={darkgray176},
            xlabel={Noise level $\epsilon$},
            xmin=-0.02, xmax=0.72,
            xtick style={color=black},
            y grid style={darkgray176},
            ylabel={Relative error (\%)},
            ymin=20, ymax=80,
            ytick style={color=black},
            xtick={0,0.2,0.4,0.6, 0.7},
            ytick={20, 40, 60, 80},
            legend cell align={left},
            legend style={
                    fill opacity=0.8,
                    draw opacity=1,
                    text opacity=1,
                    at={(0.1,0.8)},
                    anchor=west,
                    draw=lightgray204
                },
        ]
        \addplot [semithick, blue]
        table {%
                0 70.0071957025789
                0 70.0071957025789
                0.05 40.8613967778297
                0.05 40.8613967778297
                0.1 29.9028815877735
                0.15 26.8504420257351
                0.2 26.1850302093624
                0.3 25.7359938940394
                0.35 25.8901403441274
                0.4 26.0126561743333
                0.5 26.6895520110969
                0.6 28.4069877663518
                0.7 51.5957436975767
            };
        \addlegendentry{$\TNetAEfull$ with one sample training }
        \addplot [thick, black]
        table {%
                0 22.71
                0.02 22.71
                0.05 22.71
                0.1 22.71
                0.15 22.71
                0.2 22.71
                0.25 22.71
                0.3 22.71
                0.35 22.71
                0.7 22.71
            };
        \addlegendentry{Tik}
        \addplot [draw=red, fill=red, mark=asterisk, only marks]
        table{%
                x  y
                0 70.0071957025789
                0 70.0071957025789
                0.05 40.8613967778297
                0.05 40.8613967778297
                0.1 29.9028815877735
                0.15 26.8504420257351
                0.2 26.1850302093624
                0.3 25.7359938940394
                0.35 25.8901403441274
                0.4 26.0126561743333
                0.5 26.6895520110969
                0.6 28.4069877663518
                0.7 51.5957436975767
            };
    \end{axis}

\end{tikzpicture}
    \caption{{\bf 2D Navier--Stokes equation.} Relative error of inverse solution over 500 test samples with different noise levels.}
    \figlab{2D_NS_noise_level}
\end{figure}

{\bf \TNetAEfull{} robustness to arbitrary single-sample.}
The robustness of \TNetAEfull{} to an arbitrary one-training sample is examined. To be more specific, we randomly pick 12 samples out of 100 training sample data sets. The indices of 20 random observation locations are presented in the left figure in \cref{fig:2D_NS_random_traning_samples}. Meanwhile, the mean and standard deviation of observation magnitudes of 10000 true observation samples at corresponding 20 random observation locations and the predicted observation magnitudes from 12 different observation samples
are shown in \cref{fig:2D_NS_random_traning_samples}. From 12 corresponding single-sample training cases, we obtain the mean and standard deviation of the relative error of the inverse solution is $25.88 \pm 0.19 \%$ (again close to the Tikhonov regularization error). 
This small variance in relative error metrics demonstrates the \TNetAEfull{} robustness with respect to an arbitrary individual sample in the single-sample training scenario. In particular, the result shows that the prediction error is similar for any of these 10 individual random samples when used in the \TNetAEfull{} as the only training sample.

\begin{figure}[htb!]
    \centering
    \begin{tabular*}{\textwidth}{c@{\hskip -0.01cm} c@{\hskip -0.01cm}  c@{\hskip -0.01cm}}
        \centering
        & & \\
        \raisebox{-0.5\height}{\resizebox{0.33\textwidth}{0.33\textwidth}{
\begin{tikzpicture}

\begin{axis}[
unit vector ratio*=1 1 1,
width=4.5cm,
height=4.5cm,
tick pos=left,
xmin=-0.01, xmax=1.01,
ymin=-0.01, ymax=1.01,
xtick=\empty, 
ytick=\empty  
]
\addplot [black, dotted]
table {%
0 0
0 1
};
\addplot [black, dotted]
table {%
0 0
1 0
};
\addplot [black, dotted]
table {%
0 1
1 1
};
\addplot [black, dotted]
table {%
1 0
1 1
};
\addplot [semithick, black, mark=square*, mark size=0.5, mark options={solid}, only marks]
table {%
0.483870967741935 0.032258064516129
0.612903225806452 0.0645161290322581
0.419354838709677 0
0.806451612903226 0.354838709677419
0.774193548387097 0.645161290322581
0.548387096774194 0.0967741935483871
0.419354838709677 0.354838709677419
0.774193548387097 0.419354838709677
0.67741935483871 0.161290322580645
0.903225806451613 0.483870967741935
0.161290322580645 0.419354838709677
0.032258064516129 0.387096774193548
0.354838709677419 0.838709677419355
0.225806451612903 0.225806451612903
0.548387096774194 0.838709677419355
0.870967741935484 0.258064516129032
0.741935483870968 0.935483870967742
0.354838709677419 0
0.870967741935484 0.741935483870968
0.612903225806452 0.967741935483871
};
\draw (axis cs:0.493870967741936,0.022258064516129) node[
  scale=0.4,
  anchor=base west,
  text=black,
  rotate=0.0
]{1};
\draw (axis cs:0.622903225806452,0.0545161290322581) node[
  scale=0.4,
  anchor=base west,
  text=black,
  rotate=0.0
]{2};
\draw (axis cs:0.390354838709677,+0.02) node[
  scale=0.4,
  anchor=base west,
  text=black,
  rotate=0.0
]{3};
\draw (axis cs:0.816451612903226,0.344838709677419) node[
  scale=0.4,
  anchor=base west,
  text=black,
  rotate=0.0
]{4};
\draw (axis cs:0.784193548387097,0.635161290322581) node[
  scale=0.4,
  anchor=base west,
  text=black,
  rotate=0.0
]{5};
\draw (axis cs:0.558387096774194,0.0867741935483871) node[
  scale=0.4,
  anchor=base west,
  text=black,
  rotate=0.0
]{6};
\draw (axis cs:0.429354838709677,0.344838709677419) node[
  scale=0.4,
  anchor=base west,
  text=black,
  rotate=0.0
]{7};
\draw (axis cs:0.784193548387097,0.409354838709677) node[
  scale=0.4,
  anchor=base west,
  text=black,
  rotate=0.0
]{8};
\draw (axis cs:0.68741935483871,0.151290322580645) node[
  scale=0.4,
  anchor=base west,
  text=black,
  rotate=0.0
]{9};
\draw (axis cs:0.913225806451613,0.473870967741935) node[
  scale=0.4,
  anchor=base west,
  text=black,
  rotate=0.0
]{10};
\draw (axis cs:0.171290322580645,0.409354838709677) node[
  scale=0.4,
  anchor=base west,
  text=black,
  rotate=0.0
]{11};
\draw (axis cs:0.042258064516129,0.377096774193548) node[
  scale=0.4,
  anchor=base west,
  text=black,
  rotate=0.0
]{12};
\draw (axis cs:0.364838709677419,0.828709677419355) node[
  scale=0.4,
  anchor=base west,
  text=black,
  rotate=0.0
]{13};
\draw (axis cs:0.235806451612903,0.215806451612903) node[
  scale=0.4,
  anchor=base west,
  text=black,
  rotate=0.0
]{14};
\draw (axis cs:0.558387096774194,0.828709677419355) node[
  scale=0.4,
  anchor=base west,
  text=black,
  rotate=0.0
]{15};
\draw (axis cs:0.880967741935484,0.248064516129032) node[
  scale=0.4,
  anchor=base west,
  text=black,
  rotate=0.0
]{16};
\draw (axis cs:0.751935483870968,0.925483870967742) node[
  scale=0.4,
  anchor=base west,
  text=black,
  rotate=0.0
]{17};
\draw (axis cs:0.314838709677419,+0.02) node[
  scale=0.4,
  anchor=base west,
  text=black,
  rotate=0.0
]{18};
\draw (axis cs:0.880967741935484,0.731935483870968) node[
  scale=0.4,
  anchor=base west,
  text=black,
  rotate=0.0
]{19};
\draw (axis cs:0.622903225806452,0.957741935483871) node[
  scale=0.4,
  anchor=base west,
  text=black,
  rotate=0.0
]{20};
\end{axis}

\end{tikzpicture}}}
        & \multicolumn{2}{c}{\raisebox{-0.5\height}{\resizebox{0.66\textwidth}{!}{\input{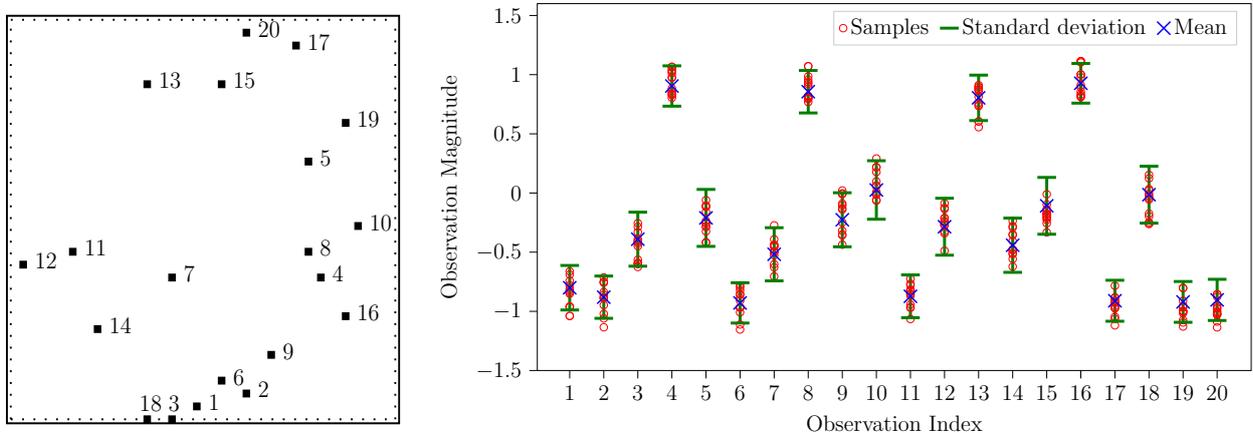}}}}
    \end{tabular*}
    \caption{\textbf{2D Navier--Stokes equation.} {\bf Left}: Index of $20$ observational locations. {\bf Right}: Mean and standard deviation of observation magnitudes of 10000 true observation samples at observational locations. The observation magnitudes of 12 different single-sample training cases.}
    \figlab{2D_NS_random_traning_samples}
\end{figure}

\begin{table}[htb!]
    \caption{The training cost (measured in hours) for the case of $n_t = 100$ randomized training samples for the heat and Navier--Stokes equations. The computational time (measured in seconds) for forward and inverse solutions  using  \TNetAEfull{} and numerical solvers, and the speed-up  of \TNetAEfull{} (the fourth row) relative to numerical solvers using NVIDIA A100 GPUs on Lonestar6 at the Texas Advanced Computing Center (TACC).}
    \tablab{comp_cost}
    \centering
    \begin{tabular}{l|l|c|c}
        \hline
        \multicolumn{2}{l|}{ $ $ } & {Heat equation} & {Navier--Stokes} \\
        \hline
        \multicolumn{2}{l|}{\makecell[l]{Training Encoder + Training Decoder \\ (hours)}} & 2 & 16 \\
        \hline
        \multirow{2}{*}{\makecell[l]{Test/Inference \\(second)}} 
        & Inverse (Encoder) & $2.74 \times 10^{-4}$ & $2.93 \times 10^{-4}$ \\
        & Forward (Decoder) & $2.86 \times 10^{-4}$ & $3.06 \times 10^{-4}$ \\
        \hline
        \multirow{2}{*}{\makecell[l]{Numerical solvers\\(second)}}
        & Inverse (Tikhonov) & $4.36 \times 10^{-2}$ & 7.26 \\
        & Forward  & $3.01 \times 10^{-2}$ & 0.38 \\
        \hline
        \multirow{2}{*}{Speed up} 
        & Inverse & 159 & \textbf{24,785} \\
        & Forward & 105 & \textbf{1,241} \\
        \hline
    \end{tabular}
\end{table}

\subsection{Training cost and speedup with deep learning solutions}
\seclab{computational_time}
The training costs for the case of $n_t = 100$ randomized training samples for heat equation and Navier--Stokes equations are presented in Table \ref{tab:comp_cost}. It can be observed that the heat equation requires a small amount of training time, about 2 hours, while the corresponding time for the Navier-Stokes equations is about 16 hours. It should be noted that executing the forward map and the backpropagation constitutes the majority of the training cost.
Table \ref{tab:comp_cost} also provides information on the computational cost of reconstructing PoIs given an unseen test observation sample and solving for PDE solutions given an unseen PoI sample. Specifically, for inverse solutions, we use the classical Tikhonov (TIK) regularization technique and our proposed deep learning approach \TNetAEfull{} using the encoder network. In contrast, we use numerical methods and \TNetAEfull{} decoder for predicting (forward) PDEs solutions. 
It can be seen that the training, and test computational time depends on the complexity of problems. We note that the complexity is estimated based on the number of time steps, the operations carried out per time step, and the mesh size. As can be seen, the more complicated the problem is, the more time Tikhonov and forward numerical solvers take to obtain the inverse and forward solutions, respectively. Unlike the Tikhonov approach, regardless of the complexity of the problems, the learned \TNetAEfull{} encoder and decoder maps using one hidden layer with 5000 neurons takes the same small amount of time: approximately $0.0003$ seconds. Note that the Tikhonov solver is implemented directly in \texttt{JAX} using the default BFGS algorithm with the gradient computed by the default Autograd functionality. Thus, the Tikhonov computation enjoys \texttt{JAX} optimized features including XLA (accelerated linear algebra), JIT (just-in-time compilation), and the nested primitive loop technique. In the meantime, the majority of computational time for numerical forward solvers stems from JIT compiling the numerical solvers. JIT functionality takes a significant amount of time to cache the modular function for the first, then all following calls of the same function are significantly faster. This is why, for simple problems such as the Heat equation, we observe a little difference between the forward solver and the Tikhonov solver which requires numerous forward solvers.
Even with such optimization, numerical forward and Tikhonov solvers are still orders of magnitude slower than \TNetAEfull{} neural networks.
In particular, for the Navier-Stokes equation, \TNetAEfull{} is $24,785$ times faster than Tikhonov and $1,241$ times faster than the numerical forward solver using the spectral method. We expect the computational gain to be more significant for larger-scale 3D time-dependent nonlinear forward problems. 
Clearly, once trained, obtaining \TNetAEfull{} solutions is simply a feed-forward neural network evaluation, which could be close to real-time or real-time depending on the network's depth and the width.

\section{Conclusion} 
\seclab{conclusions}

This paper presents a model-constrained Tikhonov autoencoder neural network framework, called \TNetAE{}, for solving forward and inverse problems simultaneously. A comprehensive error estimation analysis across pure data-driven, model-constrained, and Tikhonov model-constrained approaches is provided for linear problems. Moreover, among approaches, \TNetAE{} and its variant \TNetAEfull{} uniquely achieve both the exact Tikhonov solver and exact PtO map/forward map using an arbitrary observation sample for linear problems. This beauty stems from incorporating the data randomization strategy with a proper noise level within the model-constrained framework. Additionally, we establish theoretical support for sequential encoder-decoder training over simultaneous optimization. Numerical experiments on 2D heat equation and 2D Navier-Stokes equations validate these theoretical findings, demonstrating that \TNetAE{} and \TNetAEfull{} solutions achieve accuracy comparable to Tikhonov solutions for inverse problems and numerical forward solver solutions for forward problems while delivering computational speedups of several orders of magnitude, using only one arbitrary training sample. The numerical results also shown that \TNetAE{} and \TNetAEfull{} are robust, in the sense that their performance does not vary, for a wide range on noise level and the arbitrarily of the single training sample.
Ongoing work focuses on extending the \TNetAE{} framework to variable observational locations for inverse problems, where we anticipate developing encoders adaptable to diverse observation scenarios (varying measurement locations and quantities) while maintaining accurate PtO map/forward map representation through the decoder. 
Future work includes extensions to statistical inversions. While \TNetAE{}'s primary limitation, namely the requirement for a differentiable solver during training, can be addressed through differential numerical libraries or by substituting differential residual evaluation for the differentiable solver, these developments will be detailed in subsequent publications.

\section*{Acknowledgments}
This research is partially funded by the National Science Foundation awards NSF-OAC-2212442, NSF-2108320, NSF-1808576, and NSF-CAREER-1845799; by the Department of Energy award DE-SC0018147 and DE-SC0022211. The authors would like to thank Arjit Seth, Krishnanunni Chandradath Girija, and Wesley Lao for fruitful discussions. The authors also acknowledge the Texas Advanced Computing Center (TACC) at The University of Texas at Austin for providing HPC, visualization, database, or grid resources that have contributed to the research results reported within this paper. URL: http://www.tacc.utexas.edu

\section{Appendix}
\seclab{Appendix}
\subsection{Derivation of \texorpdfstring{$\W_e, \bb_e, \W_d, \bb_d $}{} for \pureOPO{}}
\seclab{derivation_pureOPO}

We train the encoder and decoder sequentially. First, we train the encoder, given training data samples. Then, we train the decoder, given the pretrained encoder. The optimal solutions for this training strategy are derived as follows. The loss function for the encoder is given as:

\begin{equation*}
    \mathcal{L}_e = \half \| \W_e Y + \bb_e \One^T - U \|_F^2
\end{equation*}

For simplicity, let's denote $Z = \W_e Y + \bb_e\One^T$. Then, loss function for the decoder given a pre-trained encoder is

\begin{equation*}
    \mathcal{L}_d = \half  \| \W_d Z + \bb_d \One^T - Y \|_F^2
\end{equation*}

To find the optimal values, we set all gradients to zero and solve the resulting system of equations:

\begin{itemize}
    \item Gradient with respect to $\bb_e$

          \begin{equation*}
              \eqnlab{grad_I_separate_bb_e}
              \begin{aligned}
                  \frac{\partial \mathcal{L}_e}{\partial \bb_e} & = (\W_e Y + \bb_e \One^T - U)\One = \left(\W_e Y + \bb_e \One^T\right) \One - U\One = 0
              \end{aligned}
          \end{equation*}

          Solving for $\bb_e$ we have:
          \begin{equation}
              \eqnlab{grad_I_separate_bb_e_2}
              \begin{aligned}
                  \bb_e = \frac{1}{n_t}(U\One - \W_e Y \One)
              \end{aligned}
          \end{equation}

    \item Gradient with respect to $\W_e$

          \begin{equation*}
              \eqnlab{grad_I_separate_W_e}
              \begin{aligned}
                  \frac{\partial \mathcal{L}_e}{\partial \W_e} & = (\W_e Y + \bb_e \One^T - U)Y^T = \left(\W_e Y + \bb_e \One^T \right) Y^T - U Y^T = 0
              \end{aligned}
          \end{equation*}

          Solving for $\W_e$ we have:
          \begin{equation}
              \eqnlab{grad_I_separate_W_e_2}
              \begin{aligned}
                  \W_e Y Y^T + \bb_e \One^T Y^T - UY^T                                                    & = 0                         \\
                  \W_e Y Y^T + \frac{1}{n_t}\left(U\One - \W_e Y \One \right) \One^T Y^T - UY^T            & = 0                         \\
                  \W_e Y Y^T - \W_e Y \frac{\One \One^T }{n_t} Y^T - UY^T + U \frac{\One \One^T }{n_t} Y^T & = 0                         \\
                  \W_e \bar{\Y} \bar{\Y}^T - \bar{\P} \bar{\Y}^T                                              & = 0                         \\
                  \W_e                                                                                    & = \bar{\P} \bar{\Y}^{\dagger}
              \end{aligned}
          \end{equation}

    \item Gradient with respect to $\bb_d$

          \begin{equation*}
              \eqnlab{grad_I_separate_bb_d}
              \begin{aligned}
                  \frac{\partial \mathcal{L}_d}{\partial \bb_d} & = (\W_d Z + \bb_d \One^T - Y)\One = \W_d Z \One + n_t \bb_d - Y\One = 0
              \end{aligned}
          \end{equation*}

          Solving for $\bb_d$ we have:
          \begin{equation}
              \eqnlab{grad_I_separate_bb_d_2}
              \begin{aligned}
                  \bb_d = \frac{1}{n_t}(Y\One - \W_d Z \One)
              \end{aligned}
          \end{equation}

    \item Gradient with respect to $\W_d$

          \begin{equation*}
              \eqnlab{grad_I_separate_W_d}
              \begin{aligned}
                  \frac{\partial \mathcal{L}_d}{\partial \W_d} & = (\W_d Z + \bb_d \One^T - Y)Z^T = \W_d Z Z^T + \bb_d \One^T Z^T - Y Z^T = 0
              \end{aligned}
          \end{equation*}

          Solving for $\W_d$ we have:
          \begin{equation}
              \eqnlab{grad_I_separate_W_d_2}
              \begin{aligned}
                  \W_d Z Z^T  +  \bb_d \One^T Z^T - Y Z^T = 0                                                \\
                  \W_d Z Z^T + \frac{1}{n_t}(Y\One - \W_d Z \One) \One^T Z^T  - Y Z^T = 0                     \\
                  \W_d Z Z^T - \W_d Z \frac{\One \One^T}{n_t} Z^T - Y Z^T + Y \frac{\One \One^T}{n_t} Z^T = 0 \\
                  \W_d \bar{Z} \bar{Z}^T - \bar{\Y} \bar{Z}^T = 0                                             \\
                  \W_d = \bar{\Y} \bar{Z}^{\dagger}
              \end{aligned}
          \end{equation}

\end{itemize}

Recall that ${Z} = \W_e Y + \bb_e \One^T $, then using the identities $\bar{\zb} = Z \frac{\One}{n_t}$ and $\bar{Z} = Z  - \bar{\zb} \One^T$, we have
\begin{equation}
    \eqnlab{bar_Z_I_separate}
    \begin{aligned}
        \bar{\zb} & = \left(\W_e Y + \bb_e \One^T\right)\One = \W_e {\Y} \One + \left( {\P} \One - \W_e {\Y} \One \right) = \bar{\ub} \\
        \bar{Z}   & = \W_e \bar{\Y} = \bar{\P} \bar{\Y}^{\dagger} \bar{\Y}
    \end{aligned}
\end{equation}

\begin{tcolorbox}[colback=black!5!white,colframe=black!75!black]
    {\bf Summary:} From \cref{eq:grad_I_separate_bb_e_2}, \cref{eq:grad_I_separate_W_e_2}, \cref{eq:grad_I_separate_bb_d_2}, \cref{eq:grad_I_separate_W_d_2}, and \cref{eq:bar_Z_I_separate}, the solutions for the encoder ($\W_e$, $\bb_e$) and decoder ($\W_d$, $\bb_d$) are:

    \begin{equation*}
        \eqnlab{W_b_separateI_all}
        \begin{aligned}
            \W_e & = \bar{\P} \bar{\Y}^{\dagger},                                                                       & \bb_e & = \frac{1}{n_t}\left(U\One - \W_e Y \One \right) = \bar{\ub} - \bar{\P} \bar{\Y}^{\dagger} \bar{\yb}                             \\
            \W_d & = \bar{\Y} \bar{Z}^{\dagger} = \bar{\Y} \left( \bar{\P} \bar{\Y}^{\dagger} \bar{\Y} \right)^{\dagger} , & \bb_d & = \frac{1}{n_t}(Y\One - \W_d Z \One) = \bar{\yb} - \bar{\Y} \left( \bar{\P} \bar{\Y}^{\dagger} \bar{\Y} \right)^{\dagger}\bar{\ub}
        \end{aligned}
    \end{equation*}
\end{tcolorbox}

\subsection{ Derivation of \texorpdfstring{$\W_e, \bb_e, \W_d, \bb_d $}{} for \purePOP}
\seclab{derivation_purePOP}

For the encoder loss, we have:

\[
    \mathcal{L}_e = \half \| \W_e U + \bb_e \One^T - Y \|_F^2
\]

For simplicity, let's denote $Z = \W_e U + \bb_e\One^T$. Then, the loss for decoder given pretrained encoder is:

\[
    \mathcal{L}_d = \half \| \W_d Z + \bb_d \One^T - U \|_F^2
\]

Recall that ${Z} = \W_e U + \bb_e \One^T $, then using the identities $\bar{\zb} = Z \frac{\One}{n_t}$ and $\bar{Z} = Z  - \bar{\zb} \One^T$, we have

\begin{equation*}
    \eqnlab{bar_Z}
    \begin{aligned}
        \bar{\zb} & = \bar{\ub}                                       \\
        \bar{Z}   & = \W_e \bar{\P} = \bar{\Y} \bar{\P}^{\dagger} \bar{\P} = \G \bar{\P} \bar{\P}^{\dagger} \bar{\P} = \G \bar{\P} = \bar{\Y}
    \end{aligned}
\end{equation*}

\begin{tcolorbox}[colback=black!5!white,colframe=black!75!black]
    {\bf Summary:} Following the same procedure as \cref{sect:derivation_pureOPO}, we can derive the optimal solutions for the encoder ($\W_e$, $\bb_e$) and decoder ($\W_d$, $\bb_d$) for the \purePOP{} model.
    \begin{equation*}
        \begin{aligned}
            \W_e & = \bar{\Y} \bar{\P}^{\dagger} = \G \bar{\P} \bar{\P}^{\dagger},                                                                       & \bb_e & = \frac{1}{n_t}\left(Y\One - \bar{\Y} \bar{\P}^{\dagger} U \One \right) = \bar{\yb} - \G \bar{\P} \bar{\P}^{\dagger} \bar{\ub} = \G  \LRp{\Ib - \bar{\P} \bar{\P}^{\dagger}} \bar{\ub}                              \\
            \W_d & = \bar{\P} \bar{Z}^{\dagger} = \bar{\P} \bar{\Y}^{\dagger} , & \bb_d & = \frac{1}{n_t}(U\One - \bar{\P} \bar{Z}^{\dagger} Z \One) = \bar{\ub} - \bar{\P} \bar{\Y}^{\dagger} \bar{\yb} \\
        \end{aligned}
    \end{equation*}
\end{tcolorbox}

\subsection{Derivation of \texorpdfstring{$\W_e, \bb_e, \W_d, \bb_d $}{} for \mcPOP}
\seclab{derivation_mcPOP}


For the encoder loss, we have:

\[
    \mathcal{L}_e = \halfv{1} \nor{\W_e \P + \bb_e \One^T - \Y}_{F}^2
\]

For simplicity, let's denote $Z = \W_e \P + \bb_e\One^T$. Then, the loss for decoder given pretrained encoder is:

\[
    \mathcal{L}_d = \halfv{1} \nor{\W_d Z + \bb_d \One^T - \P}_{F}^2 + \halfv{\lambda} \nor{\GB\LRp{\W_d Z + \bb_d \One^T - \Y}}_{F}^2
\]

To find the optimal values, we set all gradients to zero and solve the resulting system of equations:

\begin{itemize}
    \item Gradient with respect to $\bb_e$

          \begin{equation*}
              \begin{aligned}
                  \frac{\partial \mathcal{L}_e}{\partial \bb_e} & = (\W_e \P + \bb_e \One^T - \Y)\One = \W_e \P \One + n_t \bb_e - \Y\One = 0 \\
              \end{aligned}
          \end{equation*}
          Solving for $\bb_e$ we have:
          \begin{equation}
              \eqnlab{bb_e_II}
              \begin{aligned}
                  \bb_e = \frac{1}{n_t}(\Y\One - \W_e \P \One)
              \end{aligned}
          \end{equation}

    \item Gradient with respect to $\W_e$

          \begin{equation*}
              \begin{aligned}
                  \frac{\partial \mathcal{L}_e}{\partial \W_e} & = (\W_e \P + \bb_e \One^T - \Y)\P^T = \W_e \P \P^T + \bb_e \One^T \P^T - \Y \P^T = 0
              \end{aligned}
          \end{equation*}

          Solving for $\W_e$ we have:

          \begin{equation}
              \eqnlab{W_e_II}
              \begin{aligned}
                  \W_e \P \P^T + \bb_e \One^T \P^T - \Y \P^T = 0                                                     \\
                  \W_e \P \P^T + \frac{1}{n_t}(\Y\One - \W_e \P \One) \One^T \P^T - \Y \P^T = 0                       \\
                  \W_e \P \P^T - \W_e \P \frac{\One \One^T}{n_t} \P^T - \Y \P^T + \Y \frac{\One \One^T}{n_t} \P^T = 0 \\
                  \W_e \bar{\P} \bar{\P}^T - \bar{\Y} \bar{\P}^T = 0                                                 \\
                  \W_e = \bar{\Y} \bar{\P}^{\dagger}
              \end{aligned}
          \end{equation}

    \item Gradient with respect to $\bb_d$

          \begin{equation*}
              \begin{aligned}
                  \frac{\partial \mathcal{L}_d}{\partial \bb_d} & = (\W_d Z + \bb_d \One^T - U)\One + \lambda {\GB}^T({\GB}(\W_d Z + \bb_d \One^T) - Y)\One = 0 \\
              \end{aligned}
          \end{equation*}

          Solving for $\bb_d$ we have:

          \begin{equation}
              \eqnlab{bb_d_II}
              \begin{aligned}
                  \bb_d = \frac{1}{n_t}(U\One - \W_d Z \One) - \lambda {\GB}^T({\GB}(\W_d Z + \bb_d \One^T) - Y)\One
              \end{aligned}
          \end{equation}

    \item Gradient with respect to $\W_d$

          \begin{equation*}
              \begin{aligned}
                  \frac{\partial \mathcal{L}_d}{\partial \W_d} = (\W_d Z + \bb_d \One^T - U)Z^T + \lambda {\GB}^T({\GB}(\W_d Z + \bb_d \One^T) - Y)Z^T                                             & = 0 \\
                  \W_d ZZ^T + \bb_d \One^T Z^T - UZ^T + \lambda {\GB}^T{\GB}\W_d ZZ^T + \lambda {\GB}^T {\GB} \bb_d \One^T Z^T - \lambda {\GB}^TYZ^T                                            & = 0 \\
                  \left( \Ib +  \lambda {\GB}^T{\GB} \right) \W_d ZZ^T - \left( UZ^T + \lambda {\GB}^TYZ^T \right) + \left( \Ib + \lambda {\GB}^T {\GB} \right) \bb_d \One^T Z^T & = 0
              \end{aligned}
          \end{equation*}

          Solving for $\W_d$ we have:
          \begin{equation}
              \eqnlab{W_d_II}
              \begin{aligned}
                  \left( \Ib +  \lambda {\GB}^T{\GB} \right) \W_d ZZ^T              & = \left( UZ^T + \lambda {\GB}^TYZ^T \right) - \left( \Ib + \lambda {\GB}^T {\GB} \right) \bb_d \One^T Z^T                \\
                  \left( \Ib +  \lambda {\GB}^T{\GB} \right) \W_d ZZ^T              & = \left( UZ^T + \lambda {\GB}^TYZ^T \right)                                                                                     \\
                                                                                          & \quad \quad - \frac{1}{n_t}(U\One - \W_d Z\One + \lambda {\GB}^TY\One - \lambda {\GB}^T{\GB}\W_d Z\One) \One^T Z^T                \\
                  \left( \Ib +  \lambda {\GB}^T{\GB} \right) \W_d \bar{Z} \bar{Z}^T & = \bar{\P} \bar{Z}^T + \lambda {\GB}^T \bar{\Y} \bar{Z}^T                                                                         \\
                  \W_d                                                                     & = (\Ib + \lambda {\GB}^T{\GB})^{-1} \left( \bar{\P} \bar{Z}^{\dagger} + \lambda {\GB}^T \bar{\Y} \bar{Z}^{\dagger} \right)
              \end{aligned}
          \end{equation}
          
          Recall that $\bar{Z} = \W_e \bar{\P}$, then using the identities $\bar{\zb} = Z \frac{\One}{n_t}$ and $\bar{Z} = Z  - \bar{\zb} \One^T$, we have

          \begin{equation}
              \eqnlab{bar_Z_II}
              \begin{aligned}
                  \bar{\zb} & = \bar{\ub}                                       \\
                  \bar{Z}   & = \W_e \bar{\P} = \bar{\Y} \bar{\P}^{\dagger} \bar{\P} = \GB \bar{\P} \bar{\P}^{\dagger} \bar{\P} = \bar{\Y}
              \end{aligned}
          \end{equation}

          \begin{tcolorbox}[colback=black!5!white,colframe=black!75!black]
              {\bf Summary:} From \cref{eq:bb_e_II}, \cref{eq:W_e_II}, \cref{eq:bb_d_II}, \cref{eq:W_d_II}, and \cref{eq:bar_Z_II} we can derive the optimal solutions for the encoder ($\W_e$, $\bb_e$) and decoder ($\W_d$, $\bb_d$) for the \mcPOP{} model.
              \begin{equation*}
                  \begin{aligned}
                      \W_e   & = \bar{\Y} \bar{\P}^{\dagger} = \GB \bar{\P} \bar{\P}^\dagger,  \quad \quad \quad \quad \bb_e = \frac{1}{n_t}\left(Y\One - \bar{\Y} \bar{\P}^{\dagger} U \One \right) = \bar{\yb} - \GB \bar{\P} \bar{\P}^{\dagger} \bar{\ub},                                                                                              \\
                      \W_d   & = (\Ib + \lambda {\GB}^T{\GB})^{-1} \left( \bar{\P} \bar{\Y}^{\dagger} + \lambda {\GB}^T \bar{\Y} \bar{\Y}^{\dagger} \right),                                                                  \\
                      \bb_d & = (\Ib + \lambda {\GB}^T{\GB})^{-1} \left( \bar{\ub}  + \lambda {\GB}^T \bar{\yb}  - \left(\bar{\P} \bar{\Y}^{\dagger} + \lambda {\GB}^T \bar{\Y} \bar{\Y}^{\dagger} \right) \bar{\yb}   \right) \\
                  \end{aligned}
              \end{equation*}
          \end{tcolorbox}

\end{itemize}

\subsection{Derivation of \texorpdfstring{$\W_e, \bb_e, \W_d, \bb_d $}{} for \mcOPO}
\seclab{derivation_mcOPO}

For the encoder loss, we have:

\[
    \mathcal{L}_e = \halfv{1} \nor{\W_e \Y + \bb_e \One^T - \U}_{F}^2 + \halfv{\lambda}\nor{\GB \LRp{\W_e \Y + \bb_e \One^T} - \Y}_{F}^2
\]

For simplicity, let's denote $Z = \W_e \Y + \bb_e\One^T$. Then, the loss for decoder given pretrained encoder is:

\[
    \mathcal{L}_d = \halfv{1} \nor{\W_d Z + \bb_d \One^T - \GB Z}_{F}^2
\]

To find the optimal values, we set all gradients to zero and solve the resulting system of equations:

\begin{itemize}
    \item Gradient with respect to $\bb_e$

          \begin{equation*}
              \begin{aligned}
                  \frac{\partial \mathcal{L}_e}{\partial \bb_e} = (\W_e \Y + \bb_e\One^T - U)\One + \lambda {\GB}^T({\GB}\left(\W_e \Y + \bb_e\One^T\right) - Y)\One & = 0 \\
                  \W_e Y\One + \bb_e \One^T \One - U\One + \lambda {\GB}^T{\GB}\W_e Y\One + \lambda {\GB}^T{\GB}b_e \One^T \One - \lambda {\GB}^TY\One             & = 0 \\
                  \left( \Ib + \lambda {\GB}^T {\GB} \right) \bb_e n_t + \W_e \Y \One - U\One + \lambda {\GB}^T {\GB} \W_e \Y\One - \lambda {\GB}^T \Y \One & = 0 \\
              \end{aligned}
          \end{equation*}

          Solving for $\bb_e$ we have:

          \begin{equation*}
              \begin{aligned}
                  \bb_e & = \frac{1}{n_t} (\Ib + \lambda {\GB}^T{\GB})^{-1}(U\One - \W_e Y\One - \lambda {\GB}^T{\GB}\W_e Y\One + \lambda {\GB}^TY\One) \\
              \end{aligned}
          \end{equation*}

    \item Gradient with respect to $\W_e$
    
          \begin{equation*}
              \begin{aligned}
                  \frac{\partial \mathcal{L}_e}{\partial \W_e} = (\W_e \Y + \bb_e\One^T - U)\Y^T + \lambda {\GB}^T({\GB}\left(\W_e \Y + \bb_e\One^T\right) - Y)\Y^T                          & = 0 \\
                  \W_e \Y \Y^T + \bb_e \One^T \Y^T - \P \Y^T + \lambda {\GB}^T{\GB}\W_e \Y \Y^T + \lambda {\GB}^T{\GB} \bb_e \One^T \Y^T - \lambda {\GB}^T \Y\Y^T                         & = 0 \\
                  \left( \Ib + \lambda {\GB}^T {\GB} \right) \W_e \Y \Y^T + \left( \Ib + \lambda {\GB}^T {\GB} \right) \bb_e \One^T \Y^T - U\Y^T - \lambda {\GB}^T \Y \Y^T & = 0 \\
              \end{aligned}
          \end{equation*}

          Solving for $\W_e$ we have:

          \begin{equation}
              \eqnlab{W_e_III}
              \begin{aligned}
                  \left( \Ib + \lambda {\GB}^T {\GB} \right) \W_e \Y \Y^T & = UY^T + \lambda {\GB}^TYY^T                                                                                                  \\ & \quad - \frac{1}{n_t}(U\One - \W_e Y\One - \lambda {\GB}^T{\GB}\W_e Y\One + \lambda {\GB}^TY\One) \One^T Y^T  \\
                  (\Ib+ \lambda {\GB}^T{\GB}) \W_e \bar{\Y} \bar{\Y}^T      & = \bar{\P} \bar{\Y}^T + \lambda {\GB}^T \bar{\Y} \bar{\Y}^T                                                                       \\
                  \W_e                                                           & = (\Ib+ \lambda {\GB}^T{\GB})^{-1} \left(\bar{\P} \bar{\Y}^{\dagger} + \lambda {\GB}^T \bar{\Y} \bar{\Y}^{\dagger} \right) \\
              \end{aligned}
          \end{equation}

          Substituting $\W_e$ back into $\bb_e$ we have:
          \begin{equation}
              \eqnlab{bb_e_III}
              \begin{aligned}
                  \bb_e & = \frac{1}{n_t} (\Ib + \lambda {\GB}^T{\GB})^{-1}(U\One - \W_e Y\One - \lambda {\GB}^T{\GB}\W_e Y\One + \lambda {\GB}^TY\One)                                                   \\
                        & = \frac{1}{n_t} (\Ib + \lambda {\GB}^T{\GB})^{-1}(U\One - \left(\Ib + \lambda {\GB}^T{\GB} \right) \W_e Y\One + \lambda {\GB}^TY\One)                                   \\
                        & =  (\Ib + \lambda {\GB}^T{\GB})^{-1} (U\One - \left(\bar{\P} \bar{\Y}^{\dagger} + \lambda {\GB}^T \bar{\Y} \bar{\Y}^{\dagger} \right) Y\One + \lambda {\GB}^TY\One) \frac{1}{n_t} \\
                        & = (\Ib + \lambda {\GB}^T{\GB})^{-1} ( \bar{\ub} + \lambda {\GB}^T \bar{\yb} - \left(\bar{\P} \bar{\Y}^{\dagger} + \lambda {\GB}^T \bar{\Y} \bar{\Y}^{\dagger} \right) \bar{\yb} )
              \end{aligned}
          \end{equation}

    \item Gradient with respect to $\bb_d$

          \begin{equation*}
              \begin{aligned}
                  \frac{\partial \mathcal{L}_d}{\partial \bb_d} = (\W_d Z + \bb_d \One^T - {\GB} Z)\One = \W_d Z \One + \bb_d \One^T \One - {\GB} Z \One = 0
              \end{aligned}
          \end{equation*}

          Solving for $\bb_d$ we have:

          \begin{equation}
              \eqnlab{bb_d_III}
              \begin{aligned}
                  \bb_d = \frac{1}{n_t}({\GB} Z \One - \W_d Z \One) = {\GB} \bar{z} - \W_d \bar{z}
              \end{aligned}
          \end{equation}

    \item Gradient with respect to $\W_d$

          \begin{equation*}
              \begin{aligned}
                  \frac{\partial \mathcal{L}_d}{\partial \W_d} = (\W_d Z + \bb_d \One^T - {\GB} Z)Z^T = \W_d Z Z^T + \bb_d \One^T Z^T - {\GB} Z Z^T = 0
              \end{aligned}
          \end{equation*}

          Solving for $\W_d$ we have:

          \begin{equation}
              \eqnlab{W_d_III}
              \begin{aligned}
                  \W_d Z Z^T + \bb_d \One^T Z^T - {\GB} Z Z^T = 0  \\
                  \W_d \bar{Z} \bar{Z}^T = {\GB} \bar{Z} \bar{Z}^T \\
                  \W_d = {\GB} \bar{Z} \bar{Z}^{\dagger}
              \end{aligned}
          \end{equation}

    \item Recall that

          \begin{equation*}
              \begin{aligned}
                  Z & = \W_e Y + \bb_e \One^T                                                                                                                                                                                                                                                                                     \\
                    & = (\Ib+ \lambda {\GB}^T{\GB})^{-1} \left(\left(\bar{\P} \bar{\Y}^{\dagger} + \lambda {\GB}^T \bar{\Y} \bar{\Y}^{\dagger} \right) Y + \left( \bar{\ub} + \lambda {\GB}^T \bar{\yb} - \left(\bar{\P} \bar{\Y}^{\dagger} + \lambda {\GB}^T \bar{\Y} \bar{\Y}^{\dagger} \right) \bar{\yb} \right) \One^T\right) \\
                    & = (\Ib+ \lambda {\GB}^T{\GB})^{-1} \left( \bar{\P} \bar{\Y}^{\dagger} Y - \bar{\P} \bar{\Y}^{\dagger} \bar{\yb} \One^T + \lambda {\GB}^T \bar{\Y} \bar{\Y}^{\dagger} Y - \lambda {\GB}^T \bar{\Y} \bar{\Y}^{\dagger} \bar{\yb} \One^T + \left(\bar{\ub} + \lambda {\GB}^T \bar{\yb} \right) \One^T\right)   \\
                    & = (\Ib+ \lambda {\GB}^T{\GB})^{-1} \left( \bar{\P} \bar{\Y}^{\dagger} \bar{\Y} + \lambda {\GB}^T \bar{\Y} \bar{\Y}^{\dagger} \bar{\Y} + \left(\bar{\ub} + \lambda {\GB}^T \bar{\yb} \right)  \One^T  \right)                                                                                              \\
                    & = (\Ib+ \lambda {\GB}^T{\GB})^{-1} \left( \bar{\P} \bar{\Y}^{\dagger} \bar{\Y} + \lambda {\GB}^T \bar{\Y} \right) + (\Ib+ \lambda {\GB}^T{\GB})^{-1} \left( \left(\bar{\ub} + \lambda {\GB}^T \bar{\yb} \right)  \One^T  \right)                                                                 \\
              \end{aligned}
          \end{equation*}

          then using the identities $\bar{\zb} = Z \frac{\One}{n_t}$ and $\bar{Z} = Z  - \bar{\zb} \One^T$, we have

          \begin{equation}
              \eqnlab{bar_Z_III}
              \begin{aligned}
                  \bar{\zb} & = (\Ib+ \lambda {\GB}^T{\GB})^{-1} \left(\bar{\ub} + \lambda {\GB}^T \bar{\yb} \right)                        \\
                  \bar{Z}   & = (\Ib+ \lambda {\GB}^T{\GB})^{-1} \left( \bar{\P} \bar{\Y}^{\dagger} \bar{\Y} + \lambda {\GB}^T \bar{\Y} \right)
              \end{aligned}
          \end{equation}

\end{itemize}

\begin{tcolorbox}[colback=black!5!white,colframe=black!75!black]
    {\bf Summary:} From \cref{eq:bb_e_III}, \cref{eq:W_e_III}, \cref{eq:bb_d_III}, \cref{eq:W_d_III}, and \cref{eq:bar_Z_III} we can derive the optimal solutions for the encoder ($\W_e$, $\bb_e$) and decoder ($\W_d$, $\bb_d$) for the \mcOPO{} model.
    \begin{equation*}
        \begin{aligned}
            \W_e   & = (\Ib+ \lambda {\GB}^T{\GB})^{-1} \left(\bar{\P} \bar{\Y}^{\dagger} + \lambda {\GB}^T \bar{\Y} \bar{\Y}^{\dagger} \right),                                                      \\
            \bb_e & = (\Ib+ \lambda {\GB}^T{\GB})^{-1} ( \bar{\ub} + \lambda {\GB}^T \bar{\yb} - \left(\bar{\P} \bar{\Y}^{\dagger} + \lambda {\GB}^T \bar{\Y} \bar{\Y}^{\dagger} \right) \bar{\yb} ) \\
            \W_d   & = {\GB} \bar{Z} \bar{Z}^{\dagger}, \quad \quad \quad  \bb_d = {\GB} \left( \Ib -  \bar{Z} \bar{Z}^{\dagger} \right) \bar{\zb}
        \end{aligned}
    \end{equation*}
\end{tcolorbox}

\subsection{Derivation of \texorpdfstring{$\W_e, \bb_e, \W_d, \bb_d$}{} for \TNetAE}
\seclab{derivation_TNetAE}


The derivation is similar to the \mcOPO{} in \cref{sect:derivation_mcOPO}, the key different is $\P = \ub_0 \One^T \implies \bar{\ub} = \ub_0, \bar{\P} = 0$. Therefore, the optimal solutions for the encoder ($\W_e$, $\bb_e$) and decoder ($\W_d$, $\bb_d$) for the \TNetAE{} model are:

\begin{tcolorbox}[colback=black!5!white,colframe=black!75!black]
    {\bf Summary:} From \cref{eq:bb_e_III}, \cref{eq:W_e_III}, \cref{eq:bb_d_III}, \cref{eq:W_d_III}, and \cref{eq:bar_Z_III} we can derive the optimal solutions for the encoder ($\W_e$, $\bb_e$) and decoder ($\W_d$, $\bb_d$) for the \TNetAE{} model.
    \begin{equation*}
        \begin{aligned}
            \W_e   & = (\Ib+ \lambda {\GB}^T{\GB})^{-1} \left(\lambda {\GB}^T \bar{\Y} \bar{\Y}^{\dagger} \right),                                    \\
            \bb_e & = (\Ib+ \lambda {\GB}^T{\GB})^{-1} ( \ub_0 + \lambda {\GB}^T \bar{\yb} - \lambda {\GB}^T \bar{\Y} \bar{\Y}^{\dagger} \bar{\yb} ) \\
            \W_d   & = {\GB} \bar{Z} \bar{Z}^{\dagger}, \quad \quad \quad  \bb_d = {\GB} \left( \Ib -  \bar{Z} \bar{Z}^{\dagger} \right) \bar{\zb}
        \end{aligned}
    \end{equation*}
\end{tcolorbox}
where

\begin{equation*}
    \begin{aligned}
        \bar{\zb} & = (\Ib+ \lambda {\GB}^T{\GB})^{-1} \left(\ub_0 + \lambda {\GB}^T \bar{\yb} \right) \\
        \bar{Z}   & = (\Ib+ \lambda {\GB}^T{\GB})^{-1} \left(\lambda {\GB}^T \bar{\Y} \right)
    \end{aligned}
\end{equation*}

\subsection{ Derivation of \texorpdfstring{$\W_e, \bb_e, \W_d, \bb_d$}{} for \purePOP{} - simultaneous training}
\seclab{derivation_purePOP_together}

We consider the optimization of the \purePOP{} model with the encoder and decoder trained together. The combined loss function is given as:

\[
    \mathcal{L} = \half \| \W_e U + \bb_e \One^T - Y \|_F^2 + \halfv{\beta}  \| \W_d (\W_e U + \bb_e \One^T) + \bb_d \One^T - U \|_F^2
\]

For simplicity, let's denote $Z = \W_e U + \bb_e\One^T$. Then our loss function becomes:

\[
    \mathcal{L} = \half  \| Z - Y \|_F^2 + \halfv{\beta}  \| \W_d Z + \bb_d \One^T - U \|_F^2
\]

To find the optimal values, we set all gradients to zero and solve the resulting system of equations:

\begin{itemize}
    \item Gradient with respect to $\bb_d$

          \begin{equation*}
              \eqnlab{grad_VII_bb_d}
              \begin{aligned}
                  \frac{\partial \mathcal{L}}{\partial \bb_d} = (\W_d Z + \bb_d \One^T - U)\One  = \W_d Z \One + n_t \bb_d - U\One = 0
              \end{aligned}
          \end{equation*}
        Solving for $\bb_d$ we have:
        \begin{equation}
            \eqnlab{grad_VII_bb_d_2}
            \bb_d = \frac{1}{n_t}(U\One - \W_d Z \One)
        \end{equation}

    \item Gradient with respect to $\W_d$

          \begin{equation*}
              \eqnlab{grad_VII_W_d}
              \begin{aligned}
                  \frac{\partial \mathcal{L}}{\partial \W_d} = (\W_d Z + \bb_d \One^T - U)Z^T = \W_d Z Z^T + \bb_d \One^T Z^T - U Z^T = 0
              \end{aligned}
          \end{equation*}
        Solving for $\W_d$ we have:
        \begin{equation}
            \eqnlab{grad_VII_W_d_2}
            \begin{aligned}
                \W_d Z Z^T  +  \bb_d \One^T Z^T - U Z^T =                                                & 0                         \\
                \W_d Z Z^T - \W_d Z \frac{\One \One^T}{n_t} Z^T - U Z^T + U \frac{\One \One^T}{n_t} Z^T = & 0                         \\
                \W_d \bar{Z} \bar{Z}^T - \bar{\P} \bar{Z}^T =                                             & 0                         \\
                \W_d =                                                                                   & \bar{\P} \bar{Z}^{\dagger}
            \end{aligned}
        \end{equation}
        where $\bar{Z} = \W_e \bar{\P}$.
        From the condition for $\W_d$,  $\W_d \bar{Z} \bar{Z}^T - \bar{\P} \bar{Z}^T = \W_d \W_e\bar{\P} \bar{\P}^T \W_e^T - \bar{\P} \bar{\P}^T \W_e^T = 0$, then
        \begin{equation}
            \eqnlab{W_e_U_U_VII}
            \W_e \bar{\P} \bar{\P}^T = \W_e \bar{\P} \bar{\P}^T \W_e^T \W_d^T
        \end{equation}
        we have $\W_d \W_e \bar{\P} \bar{\P}^T \W_e^T \W_d^T$ is symmetric, and thus $\W_d \W_e \bar{\P} \bar{\P}^T $ is also symmetric. In other words, we have a relationship,
        \[ \W_d \W_e  \bar{\P} \bar{\P}^T = \bar{\P} \bar{\P}^T \W_e^T \W_d^T \]
        From \cref{eq:W_e_U_U_VII}, we have
        \begin{equation}
            \eqnlab{W_e_U_U_VII_2}
            \W_e \bar{\P} \bar{\P}^T = \W_e \W_d \W_e \bar{\P} \bar{\P}^T
        \end{equation}

    \item Gradient with respect to $\bb_e$

          \begin{equation*}
              \eqnlab{grad_VII_bb_e}
              \begin{aligned}
                  \frac{\partial \mathcal{L}}{\partial \bb_e} & = (Z - Y)\One + \beta \LRs{\W_d^T(\W_d Z + \bb_d \One^T - U)\One} \\
                                                              & = Z \One - Y\One + \beta \LRs{\W_d^T(\W_d Z + \bb_d \One^T - U)\One} = 0
              \end{aligned}
          \end{equation*}
        
          Solving for $\bb_e$ we have:
          \begin{equation}
            \eqnlab{grad_VII_bb_e_2}
            \begin{aligned}
                Z \One - Y\One + \beta \LRs{\W_d^T(\W_d Z + \bb_d \One^T - U)\One} & = 0 \\
                Z \One - Y\One + \beta \LRs{\left(\bar{\P} \bar{Z}^{\dagger}\right)^T \left(\bar{\P} \bar{Z}^{\dagger}\right) Z \One + n_t\left(\bar{\P} \bar{Z}^{\dagger}\right)^T \frac{1}{n_t}(U\One - \left(\bar{\P} \bar{Z}^{\dagger}\right) Z \One) - \left(\bar{\P} \bar{Z}^{\dagger}\right)^T U\One}                                                            & = 0 \\
                Z \One - Y\One + \beta \LRs{\left(\bar{\P} \bar{Z}^{\dagger}\right)^T \left(\bar{\P} \bar{Z}^{\dagger}\right) Z \One + \left(\bar{\P} \bar{Z}^{\dagger}\right)^T (U\One - \left(\bar{\P} \bar{Z}^{\dagger}\right) Z \One) - \left(\bar{\P} \bar{Z}^{\dagger}\right)^T U\One}                                                                            & = 0 \\
                Z \One - Y\One + \beta \LRs{\cancel{\left(\bar{\P} \bar{Z}^{\dagger}\right)^T \left(\bar{\P} \bar{Z}^{\dagger}\right) Z \One} + \cancel{\left(\bar{\P} \bar{Z}^{\dagger}\right)^T U\One} - \cancel{\left(\bar{\P} \bar{Z}^{\dagger}\right)^T \left(\bar{\P} \bar{Z}^{\dagger}\right) Z \One} - \cancel{\left(\bar{\P} \bar{Z}^{\dagger}\right)^T U\One}} & = 0 \\
                Z \One - Y\One                                                                                                                                                                                                                                                                                                                         & = 0 \\
                \W_e U \One + \bb_e \One^T \One - Y\One                                                                                                                                                                                                                                                                                                 & = 0 \\
                \bb_e = \frac{1}{n_t}\left(Y\One - \W_e U \One \right)
            \end{aligned}
        \end{equation}

    \item Gradient with respect to $\W_e$

          \begin{equation*}
              \eqnlab{grad_VII_W_e}
              \begin{aligned}
                  \frac{\partial \mathcal{L}}{\partial \W_e} & = (Z - Y)U^T + \beta \LRs{\W_d^T(\W_d Z + \bb_d \One^T - U)U^T} \\
                                                            & = Z U^T - Y U^T + \beta \LRs{\W_d^T(\W_d Z U^T + \bb_d \One^T U^T - U U^T)} = 0
              \end{aligned}
          \end{equation*}

        We first simplify the equation:
        
        \begin{equation*}
            \eqnlab{grad_VII_W_e_extra}
            \begin{aligned}
                Z U^T - Y U^T + \beta \LRs{\W_d^T(\W_d Z U^T + \bb_d \One^T U^T - U U^T)}                                                                  & = 0 \\
                \W_e \bar{\P} \bar{\P}^T - \bar{\Y} \bar{\P}^T + \beta \LRs{\W_d^T \W_d ZU^T + \W_d^T \frac{1}{n_t}(U\One - \W_d Z \One)\One^T U^T - \W_d^T U U^T} & = 0 \\
                \W_e \bar{\P} \bar{\P}^T - \bar{\Y} \bar{\P}^T + \beta \LRs{\W_d^T \W_d \bar{Z} \bar{\P}^T - \W_d^T \bar{\P} \bar{\P}^T}                            & = 0 \\
                \W_e \bar{\P} \bar{\P}^T - \bar{\Y} \bar{\P}^T + \beta \LRs{\W_d^T \W_d \W_e \bar{\P}\bar{\P}^T - \W_d^T \bar{\P} \bar{\P}^T}                         & = 0 \\
            \end{aligned}
        \end{equation*}

        The best we can solve for $\W_e$ is
        \begin{equation}
            \eqnlab{grad_VII_W_e_extra_2}
            \begin{aligned}
                \left( \Ib + \beta \W_d^T \W_d \right)  \W_e \bar{\P} \bar{\P}^T & = \bar{\Y} \bar{\P}^T + \beta \W_d^T \bar{\P} \bar{\P}^T                                                                                                    \\
                \W_e                                                          & = \left( \Ib + \beta \W_d^T \W_d \right)^{-1} \left( \bar{\Y} \bar{\P}^T + \beta \W_d^T \bar{\P} \bar{\P}^T \right) \left(\bar{\P} \bar{\P}^T\right)^{\dagger} \\
                                                                            & \text{since $A^T \left(A A^T \right)^{\dagger} = A^{\dagger}$, so}                                                                               \\
                \W_e                                                          & = \left( \Ib + \beta \W_d^T \W_d \right)^{-1} \left( \GB \bar{\P} \bar{\P}^{\dagger} + \beta \W_d^T \bar{\P} \bar{\P}^{\dagger} \right).
            \end{aligned}
        \end{equation}
\end{itemize}

\begin{tcolorbox}[colback=black!5!white,colframe=black!75!black]
    {\bf Summary:} From \cref{eq:grad_VII_bb_d_2}, \cref{eq:grad_VII_W_d_2}, \cref{eq:grad_VII_bb_e_2}, \cref{eq:grad_VII_W_e_extra_2}, and \cref{eq:W_e_U_U_VII_2}, we can derive the optimal solutions for the encoder ($\W_e$, $\bb_e$) and decoder ($\W_d$, $\bb_d$) for \pureOPO{} model as training autoencoder simultaneously.
    \begin{equation*}
        \begin{aligned}
            & \W_e   = \left( \Ib + \beta \W_d^T \W_d \right)^{-1} \left( \GB \bar{\P} \bar{\P}^{\dagger} + \beta \W_d^T \bar{\P} \bar{\P}^{\dagger} \right), && 
            & \bb_e = \bar{\yb} - \W_e \bar{\ub}, \\
            & \W_d   =  \bar{\P} \LRp{\W_e \bar{\P}}^\dagger, &&
            & \bb_d =  \bar{\ub} - \W_d \bar{\yb}, \\
            & \W_e \bar{\P} \bar{\P}^T = \W_e \W_d \W_e \bar{\P} \bar{\P}^T
        \end{aligned}
    \end{equation*}
\end{tcolorbox}

\subsection{ Derivation of \texorpdfstring{$\W_e, \bb_e, \W_d, \bb_d$}{} for \pureOPO{} - simultaneous training}
\seclab{derivation_pureOPO_together}

We consider the optimization of the \pureOPO{} model with the encoder and decoder trained together. The combined loss function is given as:

\[
    \mathcal{L} = \half \| \W_e Y + \bb_e \One^T - U \|_F^2 + \halfv{\beta}  \| \W_d (\W_e Y + \bb_e \One^T) + \bb_d \One^T - Y \|_F^2
\]

For simplicity, let's denote $Z = \W_e Y + \bb_e\One^T$. Then our loss function becomes:

\[
    \mathcal{L} = \half  \| Z - U \|_F^2 + \halfv{\beta}  \| \W_d Z + \bb_d \One^T - Y \|_F^2
\]

To find the optimal values, we set all gradients to zero and solve the resulting system of equations:

\begin{itemize}
    \item Gradient with respect to $\bb_d$

          \begin{equation*}
              \eqnlab{grad_VI_bb_d}
              \begin{aligned}
                  \frac{\partial \mathcal{L}}{\partial \bb_d} = (\W_d Z + \bb_d \One^T - Y)\One  = \W_d Z \One + n_t \bb_d - Y\One = 0
              \end{aligned}
          \end{equation*}
        Solving for $\bb_d$ we have:
        \begin{equation}
            \eqnlab{grad_VI_bb_d_2}
            \bb_d = \frac{1}{n_t}(Y\One - \W_d Z \One)
        \end{equation}

    \item Gradient with respect to $\W_d$

          \begin{equation*}
              \eqnlab{grad_VI_W_d}
              \begin{aligned}
                  \frac{\partial \mathcal{L}}{\partial \W_d} = (\W_d Z + \bb_d \One^T - Y)Z^T = \W_d Z Z^T + \bb_d \One^T Z^T - Y Z^T = 0
              \end{aligned}
          \end{equation*}
        Solving for $\W_d$ we have:
        \begin{equation}
            \eqnlab{grad_VI_W_d_2}
            \begin{aligned}
                \W_d Z Z^T  +  \bb_d \One^T Z^T - Y Z^T =                                                & 0                         \\
                \W_d Z Z^T - \W_d Z \frac{\One \One^T}{n_t} Z^T - Y Z^T + Y \frac{\One \One^T}{n_t} Z^T = & 0                         \\
                \W_d \bar{Z} \bar{Z}^T - \bar{\Y} \bar{Z}^T =                                             & 0                         \\
                \W_d =                                                                                   & \bar{\Y} \bar{Z}^{\dagger}
            \end{aligned}
        \end{equation}
        where $\bar{Z} = \W_e \bar{\Y}$.
        Form the condition for $\W_d$,  $\W_d \bar{Z} \bar{Z}^T - \bar{\Y} \bar{Z}^T = \W_d \W_e\bar{\Y} \bar{\Y}^T \W_e^T - \bar{\Y} \bar{\Y}^T \W_e^T = 0$, then
        \begin{equation}
            \eqnlab{W_e_Y_Y}
            \W_e \bar{\Y} \bar{\Y}^T = \W_e \bar{\Y} \bar{\Y}^T \W_e^T \W_d^T
        \end{equation}
        we have $\W_d \W_e \bar{\Y} \bar{\Y}^T \W_e^T \W_d^T$ is symmetric, and thus $\W_d \W_e \bar{\Y} \bar{\Y}^T $ is also symmetric. In other words, we have a relationship,
        \[ \W_d \W_e  \bar{\Y} \bar{\Y}^T = \bar{\Y} \bar{\Y}^T \W_e^T \W_d^T \]
        From \cref{eq:W_e_Y_Y}, we have
        \begin{equation}
            \eqnlab{W_e_Y_Y_2}
            \W_e \bar{\Y} \bar{\Y}^T = \W_e \W_d \W_e \bar{\Y} \bar{\Y}^T
        \end{equation}

    \item Gradient with respect to $\bb_e$

          \begin{equation*}
              \eqnlab{grad_VI_bb_e}
              \begin{aligned}
                  \frac{\partial \mathcal{L}}{\partial \bb_e} & = (Z - U)\One + \beta \W_d^T(\W_d Z + \bb_d \One^T - Y)\One        \\
                                                              & = Z \One - U\One + \beta \W_d^T(\W_d Z + \bb_d \One^T - Y)\One = 0
              \end{aligned}
          \end{equation*}

          Solving for $\bb_e$ we have:
          \begin{equation}
            \eqnlab{grad_VI_bb_e_2}
            \begin{aligned}
                Z \One - U\One + \beta \LRs{\W_d^T(\W_d Z + \bb_d \One^T - Y)\One}                                                                                                                                                                                                                                                                                   & = 0 \\
                Z \One - U\One + \beta \LRs{\left(\bar{\Y} \bar{Z}^{\dagger}\right)^T \left(\bar{\Y} \bar{Z}^{\dagger}\right) Z \One + n_t\left(\bar{\Y} \bar{Z}^{\dagger}\right)^T \frac{1}{n_t}(Y\One - \left(\bar{\Y} \bar{Z}^{\dagger}\right) Z \One) - \left(\bar{\Y} \bar{Z}^{\dagger}\right)^T Y\One}                                                            & = 0 \\
                Z \One - U\One + \beta \LRs{\left(\bar{\Y} \bar{Z}^{\dagger}\right)^T \left(\bar{\Y} \bar{Z}^{\dagger}\right) Z \One + \left(\bar{\Y} \bar{Z}^{\dagger}\right)^T (Y\One - \left(\bar{\Y} \bar{Z}^{\dagger}\right) Z \One) - \left(\bar{\Y} \bar{Z}^{\dagger}\right)^T Y\One}                                                                            & = 0 \\
                Z \One - U\One + \beta \LRs{\cancel{\left(\bar{\Y} \bar{Z}^{\dagger}\right)^T \left(\bar{\Y} \bar{Z}^{\dagger}\right) Z \One} + \cancel{\left(\bar{\Y} \bar{Z}^{\dagger}\right)^T Y\One} - \cancel{\left(\bar{\Y} \bar{Z}^{\dagger}\right)^T \left(\bar{\Y} \bar{Z}^{\dagger}\right) Z \One} - \cancel{\left(\bar{\Y} \bar{Z}^{\dagger}\right)^T Y\One}} & = 0 \\
                Z \One - U\One                                                                                                                                                                                                                                                                                                                         & = 0 \\
                \W_e Y \One + \bb_e \One^T \One - U\One                                                                                                                                                                                                                                                                                                 & = 0 \\
                \bb_e = \frac{1}{n_t}\left(U\One - \W_e Y \One \right)
            \end{aligned}
        \end{equation}

    \item Gradient with respect to $\W_e$

          \begin{equation*}
              \eqnlab{grad_I_W_e}
              \begin{aligned}
                  \frac{\partial \mathcal{L}}{\partial \W_e} & = (Z - U)Y^T + \beta \W_d^T(\W_d Z + \bb_d \One^T - Y)Y^T                 \\
                                                            & = Z Y^T - U Y^T + \beta \W_d^T(\W_d Z Y^T + \bb_d \One^T Y^T - Y Y^T) = 0
              \end{aligned}
          \end{equation*}

        We first simplify the equation:
        
        \begin{equation*}
            \eqnlab{grad_I_W_e_extra}
            \begin{aligned}
                Z Y^T - U Y^T + \beta \W_d^T(\W_d Z Y^T + \bb_d \One^T Y^T - Y Y^T)                                                                  & = 0 \\
                \W_e \bar{\Y} \bar{\Y}^T - \bar{\P} \bar{\Y}^T + \beta \LRs{\W_d^T \W_d ZY^T + \W_d^T \frac{1}{n_t}(Y\One - \W_d Z \One)\One^T Y^T - \W_d^T Y Y^T} & = 0 \\
                \W_e \bar{\Y} \bar{\Y}^T - \bar{\P} \bar{\Y}^T + \beta \LRs{\W_d^T \W_d \bar{Z} \bar{\Y}^T - \W_d^T \bar{\Y} \bar{\Y}^T}                            & = 0 \\
                \W_e \bar{\Y} \bar{\Y}^T - \bar{\P} \bar{\Y}^T + \beta \LRs{\W_d^T \W_d \W_e \bar{\Y}\bar{\Y}^T - \W_d^T \bar{\Y} \bar{\Y}^T}                         & = 0 \\
            \end{aligned}
        \end{equation*}

        The best we can solve for $\W_e$ is
        \begin{equation}
            \eqnlab{grad_I_W_e_extra_2}
            \begin{aligned}
                \left( \Ib + \W_d^T \beta \W_d \right)  \W_e \bar{\Y} \bar{\Y}^T & = \bar{\P} \bar{\Y}^T + \beta \W_d^T \bar{\Y} \bar{\Y}^T                                                                                                    \\
                \W_e                                                          & = \left( \Ib + \beta \W_d^T \W_d \right)^{-1} \left( \bar{\P} \bar{\Y}^T + \beta \W_d^T \bar{\Y} \bar{\Y}^T \right) \left(\bar{\Y} \bar{\Y}^T\right)^{\dagger} \\
                                                                            & \text{since $A^T \left(A A^T \right)^{\dagger} = A^{\dagger}$, so}                                                                               \\
                \W_e                                                          & = \left( \Ib + \beta \W_d^T \W_d \right)^{-1} \left( \bar{\P} \bar{\Y}^{\dagger} + \beta \W_d^T \bar{\Y} \bar{\Y}^{\dagger} \right).
            \end{aligned}
        \end{equation}
        
\end{itemize}

\begin{tcolorbox}[colback=black!5!white,colframe=black!75!black]
    {\bf Summary:} From \cref{eq:grad_VI_bb_d_2}, \cref{eq:grad_VI_W_d_2}, \cref{eq:grad_VI_bb_e_2}, \cref{eq:grad_I_W_e_extra_2}, and \cref{eq:W_e_Y_Y_2}, we can derive the optimal solutions for the encoder ($\W_e$, $\bb_e$) and decoder ($\W_d$, $\bb_d$) for \pureOPO{} model as training autoencoder simultaneously.
    \begin{equation*}
        \begin{aligned}
            & \W_e   = \left( \Ib + \beta \W_d^T \W_d \right)^{-1} \left( \bar{\P} \bar{\Y}^{\dagger} + \beta \W_d^T \bar{\Y} \bar{\Y}^{\dagger} \right), && 
            \bb_e & = \bar{\ub} - \W_e \bar{\yb}, \\
            & \W_d =  \bar{\Y} \LRp{\W_e \bar{\Y}}^\dagger, &&
            \bb_d & =  \bar{\yb} - \W_d \bar{\ub}, \\
            & \W_e \bar{\Y} \bar{\Y}^T = \W_e \W_d \W_e \bar{\Y} \bar{\Y}^T
        \end{aligned}
    \end{equation*}
\end{tcolorbox}

\bibliographystyle{plain}
\bibliography{references,referencesNew}

\end{document}